\documentclass[journal]{IEEEtran}

\usepackage{cite}
\usepackage{graphicx}
\usepackage{multicol}
\usepackage{subfig}
\usepackage{amsmath}
\usepackage{amsfonts}
\usepackage{romannum}
\usepackage{caption}
\usepackage{tabularx}
\usepackage{tabulary}
\usepackage{array}
\usepackage{tabu}

\usepackage{caption}
\ifCLASSINFOpdf
\else
\fi

\begin{document}

\title{Deeply Self-Supervised Contour Embedded\\ Neural Network Applied to Liver Segmentation}
\author{Minyoung Chung, Jingyu Lee, Minkyung Lee, Jeongjin Lee$^{\ast}$, and Yeong-Gil Shin%
\thanks{\textit{Asterisk indicates corresponding author.}}%
\thanks{M. Chung, J. Lee, M. Lee, and Y.-G. Shin are with the Department of Computer Science and Engineering, Seoul National University, Korea (e-mail: chungmy@snu.ac.kr).}%
\thanks{*J. Lee is with the Department of Computer Science and Engineering, Soong-sil University, Korea (e-mail: leejeongjin@ssu.ac.kr).}
\thanks{This work was partly supported by Institute for Information \& communications Technology Promotion(IITP) grant funded by the Korea government(MSIT) (No.2017-0-018715, Development of AR-based Surgery Toolkit and Applications). And this research was supported by the Basic Science Research Program through the National Research Foundation of Korea (NRF) funded by the Ministry of Science, ICT and Future Planning (No. 2017R1D1A1B03034484). And this work was partly supported by Institute for Information \& Communications Technology Promotion(IITP) grant funded by the Korean government(MSIP). [R0118-16-1003, Authoring Platform Technology for Next-Generation Plenoptic Contents].}
}


\maketitle

\begin{abstract}
\textit{Objective:} Herein, a neural network-based liver segmentation algorithm is proposed, and its performance was evaluated using abdominal computed tomography (CT) images.
\textit{Methods:} A fully convolutional network was developed to overcome the volumetric image segmentation problem. To guide a neural network to accurately delineate a target liver object, the network was deeply supervised by applying the adaptive self-supervision scheme to derive the essential contour, which acted as a complement with the global shape. The discriminative contour, shape, and deep features were internally merged for the segmentation results. 
\textit{Results and Conclusion:} 160 abdominal CT images were used for training and validation. The quantitative evaluation of the proposed network was performed through an eight-fold cross-validation. The result showed that the method, which uses the contour feature, segmented the liver more accurately than the state-of-the-art with a 2.13\% improvement in the dice score.
\textit{Significance:} In this study, a new framework was introduced to guide a neural network and learn complementary contour features. The proposed neural network demonstrates that the guided contour features can significantly improve the performance of the segmentation task.

\end{abstract}


\IEEEpeerreviewmaketitle

\section{Introduction}

\IEEEPARstart{L}{iver} segmentation plays a crucial role in liver structural analyses, volume measurements, and clinical operations (e.g., surgical planning). For clinical usage, the accurate segmentation of a liver is one of the key components of automated radiological diagnosis systems. The manual or semi-automatic segmentation of the liver is an impractical task because of its large shape variability and unclear boundaries. Unlike other organs, ambiguous boundaries with heart, stomach, pancreas, and fat make liver segmentation difficult. Thus, for a computer-aided diagnosis system, the fully automatic and accurate segmentation of the liver plays an important role in medical imaging.\par
Multiple methods have been proposed to segment a liver \cite{lim2006automatic, rusko2007fully, suzuki2010computer, lee2007efficient, zhang2010automatic, okada2007automated, ling2008hierarchical, heimann2009comparison, campadelli2009liver, van2007automatic}. The simplest and most intuitive approaches to perform liver segmentation are thresholding and region growing \cite{lim2006automatic, rusko2007fully}. Active contour model approaches \cite{suzuki2010computer, lee2007efficient} have also been reported, mainly using intensity distributions. However, such a local intensity-based approach easily fails owing to the great variability of shapes and intensity contrasts. Shape-prior-based methods such as active shape model, statistical shape model, and registration-based methods have been developed to overcome such difficulties \cite{ling2008hierarchical, zhang2010automatic, okada2007automated,  heimann2007statistical, kainmuller2007shape, wimmer2009generic, van2007automatic}. Shape-based methods are more successful than simple intensity-based methods owing to embedded shape priors. However, the shape-based methods suffer from limited prior information because of the difficulty of embedding all inter-patient organ shapes. Thus, the number of training statistical models directly affects the model matching performance.\par
In recent years, deep neural networks (DNNs) have widely been used for various imaging applications \cite{simonyan2014very, he2016deep, szegedy2017inception, long2015fully, noh2015learning, badrinarayanan2017segnet, fu2017stacked, dong2016image, jegou2017one}. For imaging applications, the convolutional neural network (CNN) is the most effective used network with respect to image classification \cite{simonyan2014very, he2016deep, szegedy2017inception}, segmentation \cite{long2015fully, noh2015learning, badrinarayanan2017segnet, fu2017stacked, jegou2017one}, and enhancement \cite{dong2016image, burger2012image}. Various active studies have successfully applied CNNs to medical image segmentation \cite{ronneberger2015u, cciccek20163d, chen2017voxresnet, milletari2016v, chen2017dcan, kamnitsas2017efficient, havaei2017brain, dou20173d, jegou2017one, chen2018deeplab, oktay2018anatomically, gibson2018automatic}. The U-net applies contracting and expanding paths together with skip connections, which successfully combines both low and high-level features \cite{ronneberger2015u}. However, the U-net is not suitable for volumetric image segmentation as it is a fully convolutional network (FCN) based on 2D images. A 2D network architecture cannot leverage complex 3D anatomical information. The 3D U-net has been used to overcome the limitation of the original U-net architecture to extract 3D contextual information via 3D convolutions with sparse annotations \cite{cciccek20163d}. However, the 3D U-net presents limitations of slice-based annotations. In \cite{milletari2016v}, a full 3D-CNN-based U-net-like architecture was reported to segment volumetric medical images using a dice coefficient loss metric and overcome the class imbalance issue. The deep contour-aware network \cite{chen2017dcan} has been developed to depict clear contours with a multi-task framework. The VoxResNet has performed brain tissue segmentation using a voxelwise residual network \cite{chen2017voxresnet}. A residual learning mechanism has been used to classify each voxel \cite{he2016deep}. Subsequently, an auto-context algorithm \cite{tu2010auto} has been employed to further refine the voxelwise prediction results. Deeply supervised networks \cite{lee2015deeply} have been developed to hierarchically supervise multiple layers and segment medical images \cite{dou20173d}. Deep supervision has allowed effective fast learning and regularization of the network. A fully connected conditional random field model has been applied as a post-processing step to refine the segmentation results \cite{dou20173d}. In \cite{oktay2018anatomically}, the incorporation of global shape information with neural networks was presented. A convolutional autoencoder network was constructed to learn anatomical shape variations from training images \cite{oktay2018anatomically}.\par

\begin{figure}[t]
    \centering
    \includegraphics[width=\linewidth]{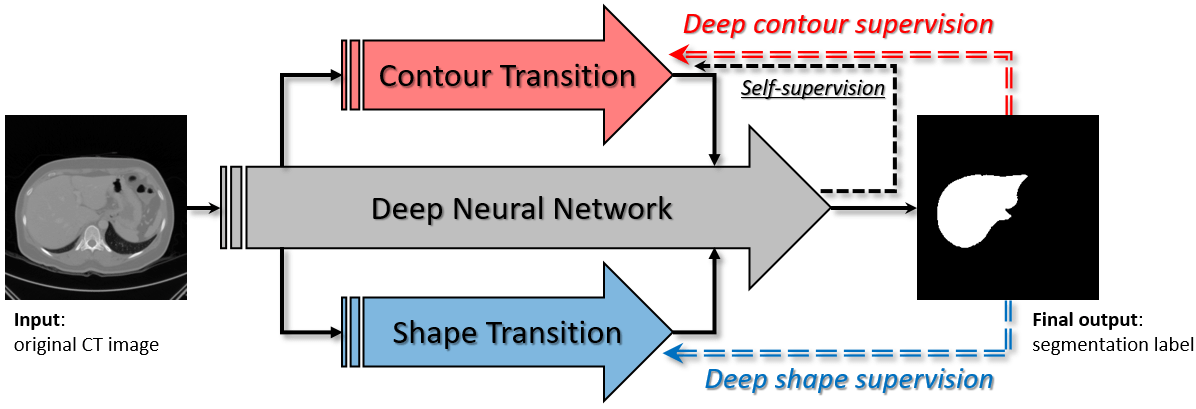}
    \caption{Proposed network architecture. The contour and shape features are embedded in the base DNN by applying deep supervisions to the two separate transition layers. The final prediction of the network is used to modify the ground-truth contour image to guide the network to learn effective contour features (i.e., contour self-supervision).}
    \label{fig:brief}
\end{figure}

Herein, we propose a deeply self-supervising CNN with adaptive contour features. Instead of learning explicit ground-truth contour features such as in reference \cite{chen2017dcan}, we guide a neural network to learn complementary contour region that can aid the accurate delineation of the target liver object. The main objective for learning partially significant contour is that, unlike other segmentation problems (e.g., glands), the contour of a liver is difficult to obtain accurately, even with DNNs, because of its ambiguous boundaries. Learned partial contours are later fused with a global shape prediction to derive the final segmentation (Fig. \ref{fig:brief}). As shown in Fig. \ref{fig:brief}, the network can be interpreted as a contour embedded shape estimation that uses three discriminative features: shape, contour, and deep features. Similar to the method presented in reference \cite{gibson2018automatic}, the proposed base network architecture was designed as a densely connected V-net structure \cite{milletari2016v}. The number of parameters and layers are effectively reduced using a densely connected network architecture \cite{huang2017densely} and separable convolutions while preserving the network capability. Finally, the learned DNN was used for automatic segmentation of the liver from CT images.\par
The remainder of this article is organized as follows. In Section {\Romannum{2}}, several CNN models that are closely related to the proposed method are reviewed. The proposed method is described in Section {\Romannum{3}}. The experimental results, discussion, and conclusion are presented in Sections {\Romannum{4}}, {\Romannum{5}}, and {\Romannum{6}}, respectively.

\section{RELATED WORK}
In this section, the CNN mechanism is reviewed and three major related works that contribute to key steps of our method are described: the V-net \cite{milletari2016v}, deeply supervising networks (DSNs) \cite{lee2015deeply, dou20173d}, and densely connected convolutional networks (DenseNets) \cite{huang2017densely}.

\subsection{V-net}
The V-net is a volumetric FCN used for medical image segmentation \cite{milletari2016v}. The U-net architecture \cite{ronneberger2015u} was extended to a volumetric convolution (i.e., 3D convolution), and U-net-like downward and upward transitions (i.e., convolutional reduction and de-convolutional expanding of feature dimensions \cite{milletari2016v}) were adopted together with multiple skip connections via an element-wise summation scheme. The dice loss was first presented to overcome the class imbalance issue.

\subsection{DSN}
A DSN was proposed to supervise a network to a deep level \cite{lee2015deeply}. Accordingly, a loss function penetrates through multiple layers in a DNN. The deeply supervising scheme makes intermediate features highly discriminative so that the final classifier can easily be a more accurate discriminative classifier for the output. Another aspect of the DSN is that training difficulties due to exploding and vanishing gradient issues can be alleviated by direct and deep gradient flows. In \cite{dou20173d}, a 3D deep supervision mechanism was adapted to volumetric medical image segmentation. The explicit supervision was exploited to hidden layers, and the auxiliary losses were integrated to the final loss with the last output layer to back-propagate gradients.

\begin{figure*}[h!bt]
    \centering
    \includegraphics[width=\linewidth]{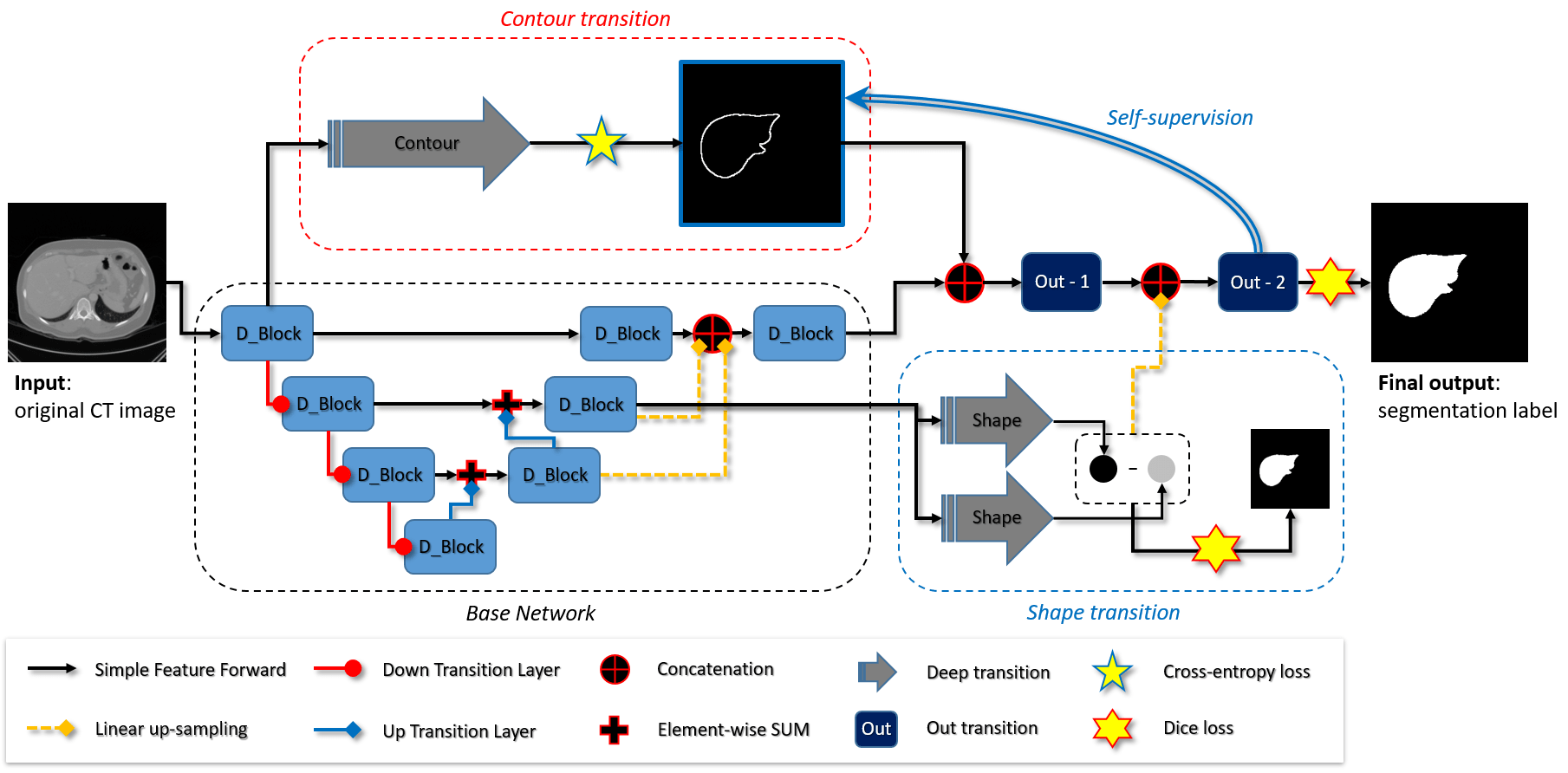}
    \caption{Proposed volumetric network architecture. Stacked densely connected blocks (i.e., D\_Block) form a base architecture with multiple skip connections. The red (i.e., circled arrows) and blue arrows (i.e., squared arrows from the D\_Blocks) indicate down- and up-transition layers, respectively. The orange lines (i.e., dotted squared arrows) indicate the up-sampling layers with a linear interpolation scheme. The red and blue dotted boxes represent the contour and shape transitions, respectively. The two transitions are deeply supervised by the contour and ground-truth images. The final output prediction is achieved by successive out-transition layers that combine the deep features. All the images are displayed as 2D for simplicity.}
    \label{fig:network}
\end{figure*}

\subsection{DenseNet}
A DenseNet \cite{huang2017densely} connects each layer to every other layer in a feed-forward manner. The main advantage of the presented architecture is that the gradient directly flows to deep layers, accelerating the learning procedure. The feature reuse also strongly contributes to a substantial reduction of the number of parameters. This structure can be viewed as an implicit deep supervision network similar to the explicit version \cite{lee2015deeply}. The \(l^{th}\) layer obtains the concatenation of all outputs of the preceding layers as follows \cite{huang2017densely}:
\newline
\begin{equation}
    x_l=H_l(x_0, x_1, ..., x_{l-1}),
\label{eq:densenet}
\end{equation}\newline
where \(x_l\) is the output of the \(l^{th}\) layer, \([x_0, x_1, ..., x_{l-1}]\) is the concatenation of the feature-maps produced in the previous layers, and \(H_l\) is a non-linear transformation at the \(l^{th}\) layer (e.g., composition of the convolution and non-linear activation function). The feature-reusing scheme of the DenseNet, which causes the reduction of the parameters, is an effective feature for the 3D volumetric neural network as the volume data lack GPU memory for DNNs.

\section{METHODOLOGY}

The base architecture of the network is composed of several contracting, expanding paths, and skip connections, similar to the V-net \cite{milletari2016v}. The key feature of the proposed network is that two different deep-supervisions are embedded in the network: contour and shape transition layers (i.e., the red and blue dotted boxes in Fig. \ref{fig:network}). Deeply supervised contour and shape features are sequentially concatenated for the final segmentation result. There are three different non-linear modules in the proposed model: a D\_Block (Fig. \ref{fig:denseblock}) and deep and out-transition layers (Fig. \ref{fig:transition}). Each module comprises a convolution, batch normalization \cite{ioffe2015batch}, rectified linear unit (ReLU) non-linearity \cite{nair2010rectified}, and skip connections. The details of the architecture and deep supervisions are described in the following subsections.



\subsection{Base Network Architecture}
\begin{figure}[t]
    \centering
    \includegraphics[width=\linewidth]{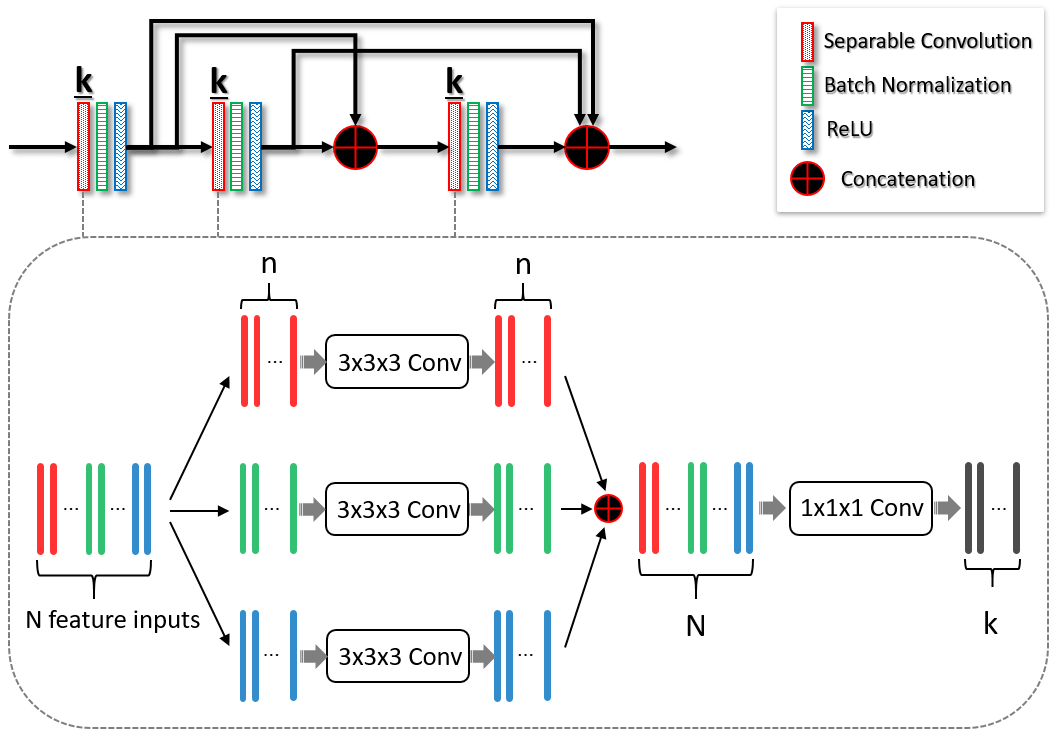}
    \caption{Densely connected block component (i.e., D\_Block). The number of feature inputs for each separable convolution is $N$ and the features are separated by groups containing $n$ features. \(k\) is the number of features produced by a $1\times1\times1$ convolution applied to concatenated features. The number of total features of a single D\_Block becomes \(3k\). Separable convolutions are applied to all D\_Block units.}
    \label{fig:denseblock}
\end{figure}

As shown in Fig. \ref{fig:network}, the D\_Block is the base non-linear module of the network. The D\_Block is composed of non-linear transformation series: a convolution, batch normalization, and ReLU non-linear activation function (Fig. \ref{fig:denseblock}). These transformations are densely connected for feature reuse. Unlike the previous research reported in reference \cite{huang2017densely}, depth-wise separable convolutions \cite{chollet2017xception} are introduced in the densely connected block instead of bottleneck layers \cite{szegedy2016rethinking} or compression layers \cite{huang2017densely} for a more efficient use of the parameters.\par

The base network uses a D\_Block as a non-linear module and performs several contracting (i.e., down-transition), expanding (i.e., up-transition) paths, and concatenating skip connections. For the down-transition layers (i.e., down-sampling feature dimensions; circled red lines in Fig. \ref{fig:network}), the feature map is down-sampled by a factor of 2 for each dimension via convolutions with stride 2. The number of features of the input is preserved. For the up-transition layers (i.e., up-sampling feature dimensions; squared blue lines in Fig. \ref{fig:network}), de-convolution (i.e., transposed convolution) is used, restoring the number of features as that of the skip connected upper layer for feature summation. Each up-transitioned layer is summed with previous feature outputs (i.e., element-wise summation in Fig. \ref{fig:network}) and passes through a D\_Block unit. The feature outputs of the lower layers are up-scaled (i.e., orange lines in Fig. \ref{fig:network}) and concatenated for further propagation of the layers. At the final stage, the contour and shape features are sequentially concatenated to the out-transition layers (Fig. \ref{fig:network}).\par

The final prediction of the network is achieved by integrating the three major features: 1) deep features from the base network (i.e., stack of D\_Block), 2) contour features from the contour transition branch (i.e., the red-dotted box in Fig. \ref{fig:network}), and 3) shape features from the shape transition branch (i.e., the blue-dotted box in Fig. \ref{fig:network}). The two deep transition layers are deeply supervised for each feature extraction.

\subsection{Deeply Supervised Transition Layers}
\begin{figure}[t]
    \centering
    \subfloat[Deep transition layers (i.e., contour and shape).]{\includegraphics[width=3.5in]{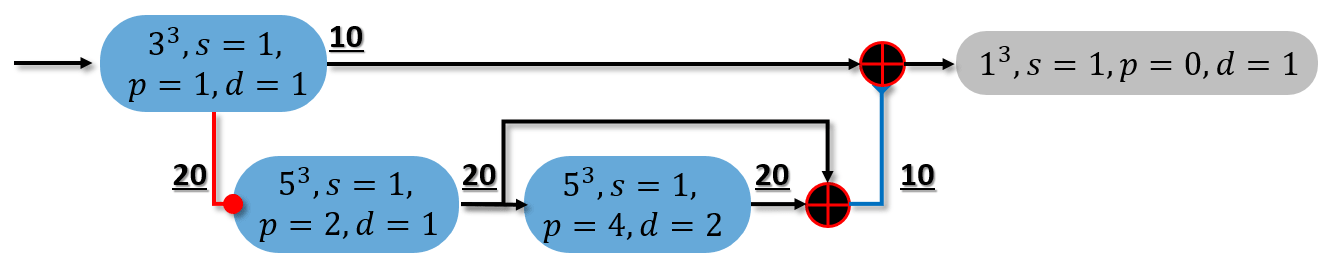}%
    \label{fig:transition_deep}}
    \vfil
    \subfloat[Out-transition layers.]{\includegraphics[width=3.5in]{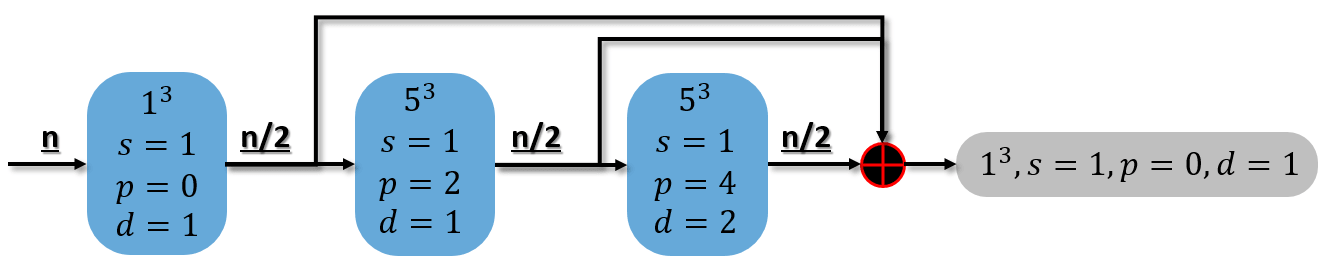}%
    \label{fig:transition_out}}
    \caption{Transition layers for (a) contour, shape, and (b) out-transition layers. The blue boxes indicate a series of convolutions, batch normalization, and non-linear activation. The gray boxes indicate a single convolution layer. Each kernel size, stride ($s$), padding ($p$), and dilation value ($d$) is specified.}
    \label{fig:transition}
\end{figure}

The transition layers (Fig. \ref{fig:transition}) are also composed of non-linear transformation series such as the D\_Block. In the transition layers, however, separable convolutions are not used. The deep transition layers (Fig. \ref{fig:transition_deep}) perform down- and up-transitions (i.e., the red- and blue-circled arrows in Fig. \ref{fig:transition_deep}) as in the base network. By contracting and expanding paths, the deep transition layer can extract more multi-scaled features (i.e., higher receptive field) with respect to the contour and shape features. The out-transition layers simply forward the feature maps with dense connections followed by a $1\times1\times1$ convolution (Fig. \ref{fig:transition_out}). There are two out-transition layers in the network for integrating features at the final stage.\par

As shown in Fig. \ref{fig:network}, we applied two different deep supervision mechanisms in the proposed model: shape and contour transitions. The shape supervision is applied to the output feature map of the two shape transition layers (i.e., the blue-dotted box in Fig. \ref{fig:network}). Two identical transitions were applied separately to learn the complementary residuals. The final shape estimation was performed by a simple subtraction between the two feature maps. Using this method, a compact shape estimation architecture that constitutes two complementary feature extractors was successfully designed and could be used to aid the prediction. The effectiveness of the residual connection is evaluated in Section \Romannum{4}.\par

For deep supervision of the contour (i.e., the red-dotted box in Fig. \ref{fig:network}), the ground-truth contour image \(\Gamma_c\), was dynamically modified for every iteration (paired blue arrow in Fig. \ref{fig:network}):
\newline
\begin{equation}
    \Tilde{\Gamma_c}=\Gamma_c \otimes (\bold{y}_p),
\label{eq:closs1}
\end{equation}
\newline
where \(\otimes\) is an element-wise multiplication operator and \(\bold{y}_p\) is a binary image with respect to the threshold value, \(p\):
\newline
\begin{equation}
    \bold{y}_p(x)=
    \begin{cases}
    1, & \text{if}\ \bold{y}(x)<p \\
    0, & \text{otherwise},
    \end{cases}
\label{eq:closs2}
\end{equation}
\newline
where \(\bold{y}\) is the output probability prediction of the proposed network for a given iteration. That is, the ground-truth contours (i.e., foreground voxels in \(\Gamma_c\)) were automatically erased if our network successfully delineated the corresponding labels at the output. This adaptive self-supervision procedure aids the contour transition layer to effectively delineate the misclassified contour region with respect to low-level features (e.g., edge). The discriminative feature of the contour transition was later combined with the shape prediction for the final liver object delineation.

\subsection{Overall Loss Function}
The vectors \(\bold{x}=\{x_i\in\textit{R}, i\in\mathbb{R}^3\}\) and \(\bold{y}=\{y_i\in\{0,1\}, i\in\mathbb{R}^3\}\) represent the input image and ground-truth label, respectively. The task of the given learning system is to model a conditional probability distribution, \(P(\bold{y}|\bold{x})\). To effectively model the probability distribution, the proposed network model was trained to map the segmentation function \(\phi(\bold{x}):\bold{x}\longrightarrow\{0,1\}\) by minimizing the following loss function:
\newline
\begin{equation}
\begin{split}
    \textit{L}_p(\bold{x}, \bold{y};W)=&\mathcal{D}(F_o, \bold{y})+\alpha\mathcal{D}(F^0_s-F^1_s, \bold{y})+\\
    &\beta\chi(F_c, \Tilde{\Gamma_c})+\gamma\|W\|_2^2,
\label{eq:loss_total}
\end{split}
\end{equation}
\newline
where $F_o$, $F^i_s$, and $F_c$ indicate the output features of out, shapes, and contour transitions, respectively. $\Gamma$ is the binary ground-truth label and $W$ is the set of parameters of the network. \(\mathcal{D}\) indicates the dice loss \cite{milletari2016v}, and \(\chi\) indicates softmax-cross-entropy loss,
\newline
\begin{equation}
    \chi(\bold{x}, \bold{y})=\sum_i(-w_i log(\frac{exp(\bold{x}[y_i])}{\sum_j{exp(\bold{x}[j])}})).
\label{eq:loss_xentropy}
\end{equation}
\newline
\(\alpha, \beta\), and \(\gamma\) in (\ref{eq:loss_total}) are weighting parameters. The \(w_i\) parameter in (\ref{eq:loss_xentropy}) is a class balancing weight. The output of the network is obtained by applying softmax to the final output feature maps.\par

\subsection{Data Preparation and Augmentation}
In total, 160 subjects were acquired: 90 subjects from a publicly available dataset\footnote{DOI:http://doi.org/10.5281/zenodo.1169361} in \cite{gibson2018automatic}, 20 subjects from the MICCAI-Sliver07 dataset \cite{heimann2009comparison}, 20 subjects from 3Dircadb\footnote{https://www.ircad.fr/research/3dircadb}, and an additional 30 annotated subjects with the help of clinical experts in the field. In the dataset, the slice thickness ranged from 0.5 to 5.0mm, and the pixel sizes ranged from 0.6 to 1.0mm.\par
For the training dataset, all abdominal computed tomography images were resampled by $128\times128\times64$. The image was pre-processed using fixed windowing values: level = 10 and width = 700 (i.e., clipped the intensity values under $-340$ and over $360$). After re-scaling, the input images were normalized into the range [0-1] for each voxel. On-the-fly random affine deformations were subsequently applied to that dataset for each iteration with 80\% probability. Finally, the cutout image augmentation \cite{devries2017improved} was performed with 80\% probability. The position of the cutout mask was not constrained with respect to the boundaries. A randomly sized zero mask was applied in the range \(L/5\leq l \leq L/4\), where \(l\) and \(L\) are the lengths of the mask and the image in each dimension, respectively. To the best of our knowledge, this is the first study applying a cutout \cite{devries2017improved} augmentation to an image segmentation problem. The effect of the cutout augmentations is presented in Section \Romannum{4}.

\subsection{Learning the Network}
'Xavier' initialization \cite{glorot2010understanding} is used for initializing all the weights of the proposed network. While training the network, the loss parameters were fixed to \(\alpha=1\), \(\beta=1\), and \(\gamma=0.1\) in (\ref{eq:loss_total}). The parameter \(p\) was set to 1 until 100 epochs, and decayed by multiplying 0.9 for every 10 epochs until 0.5 (i.e., the minimum value of \(p\)). For the dense block unit, \(k=16\) and \(n=4\) were used as parameters for the D\_Block. The Adam optimizer was used with a batch size of 4 and learning rate 0.001. The learning rate was decayed by multiplying 0.1 for every 50 epochs. The network was trained for 300 epochs using an Intel i7-7700K desktop system with a 4.2 GHz processor, 32 GB memory, and Nvidia Titan Xp GPU machine. It took 10 h to complete all the training procedures.
\par

\begin{figure}[t]
    \centering
    \subfloat[eight-fold (i.e., 140/20) cross-validation.]{\includegraphics[width=\linewidth]{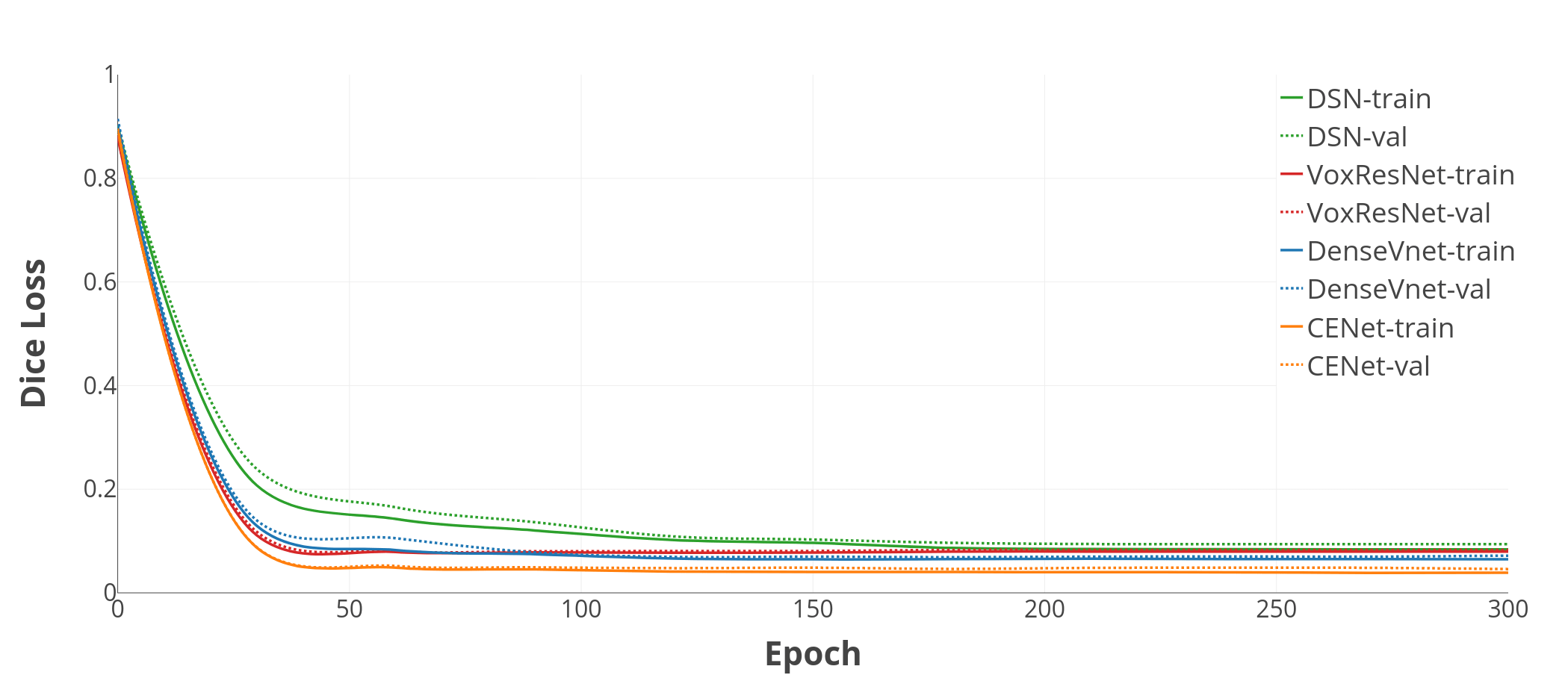}%
    \label{fig:learningcurve_8fold}}
    \vfil
    \centering
    \subfloat[10/150 cross-validation.]{\includegraphics[width=\linewidth]{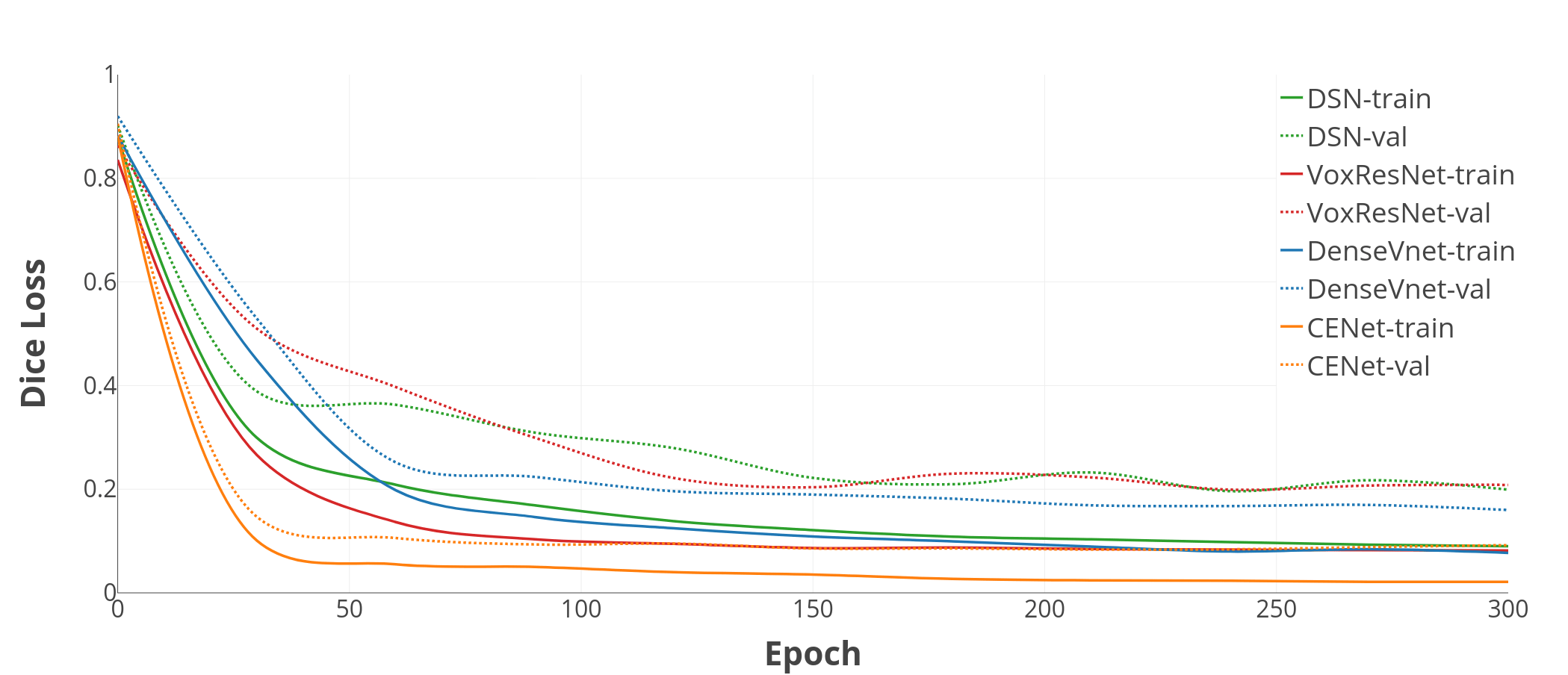}%
    \label{fig:learningcurve_sfold_wcutout}}
    \vfil
    \centering
    \subfloat[10/150 cross-validation without cutout augmentations.]{\includegraphics[width=\linewidth]{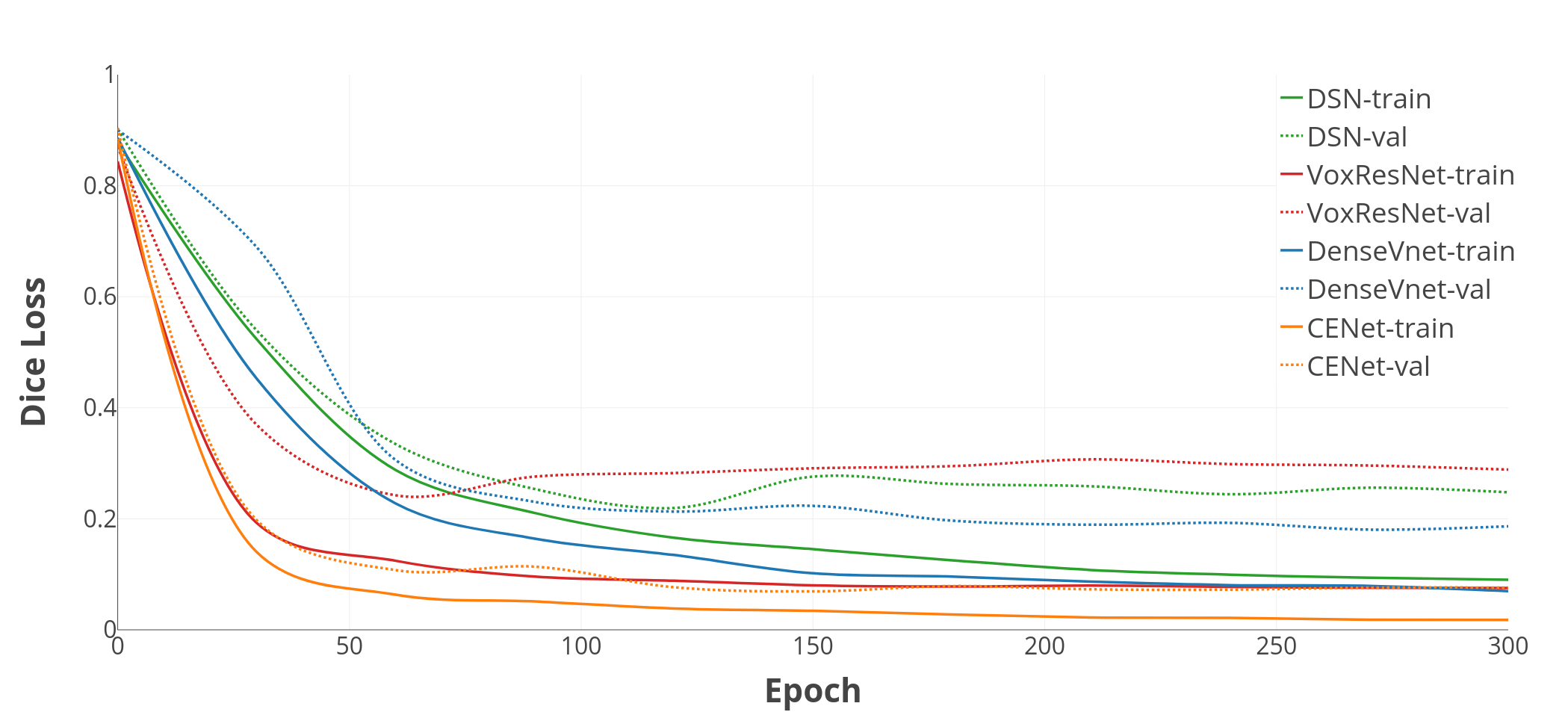}%
    \label{fig:learningcurve_sfold_wocutout}}
    \vfil
    \caption{Learning curves of the DSN \cite{dou20173d}, VoxResNet \cite{chen2017voxresnet}, DenseVNet \cite{gibson2018automatic}, and CENet with multiple cross-validations: (a) 140 images were used for training and 20 images were used for validation (i.e., eight-fold cross-validation). (b) and (c) 10 images were used for training and 150 images were used for validation. (c) shows the learning curve without cutout augmentation.}
    \label{fig:learningcurve}
\end{figure}

\begin{figure*}[tb]
    \centering
        \begin{minipage}[b]{3.5in}
            \captionsetup[subfigure]{labelformat=parens,labelsep=space,font=small}
            \centering
                \vfil
                \includegraphics[width=1.11in]{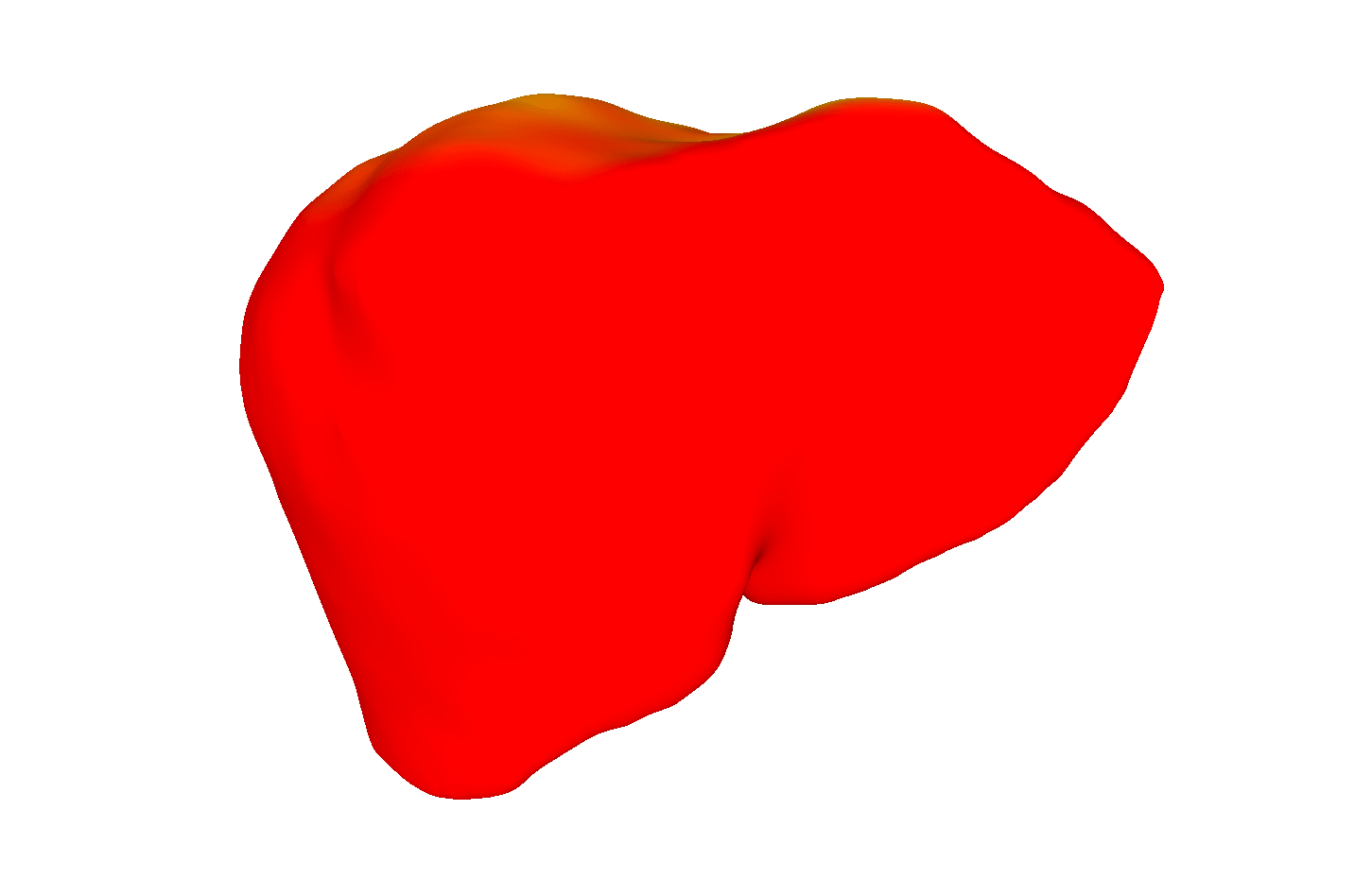}
                \includegraphics[width=1.11in]{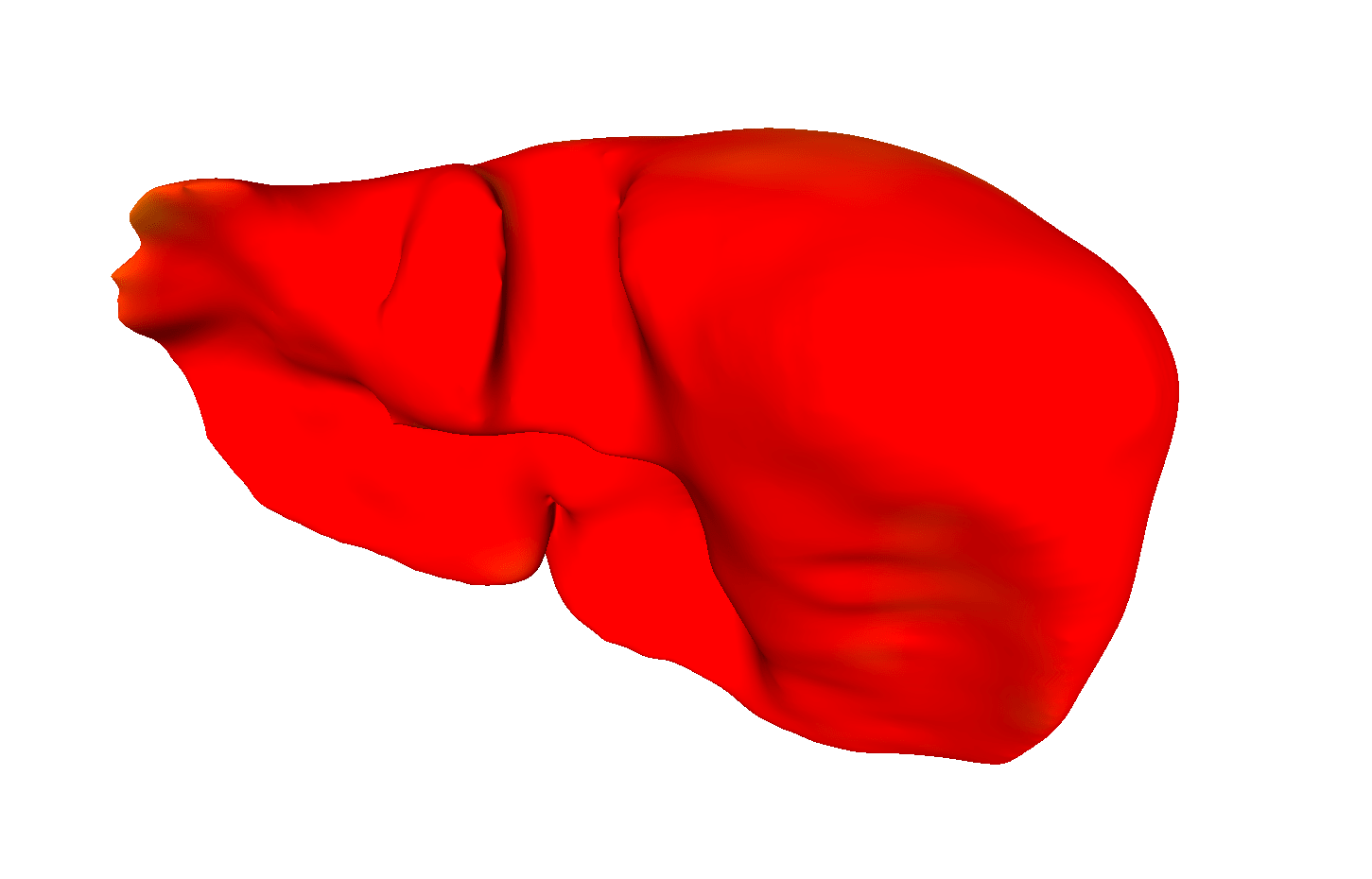}
                \includegraphics[width=1.11in]{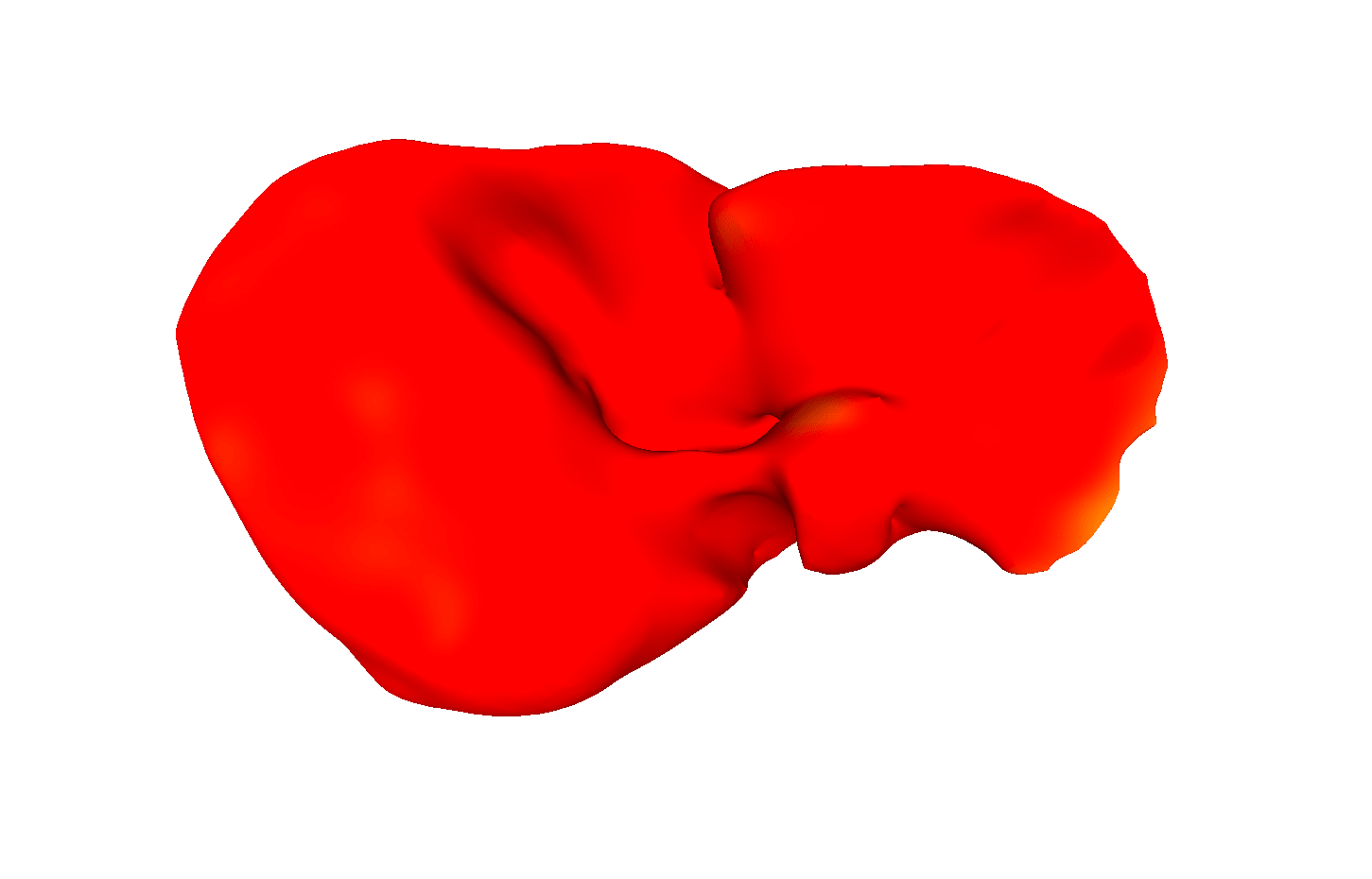}
                \vfil
                \includegraphics[width=1.11in]{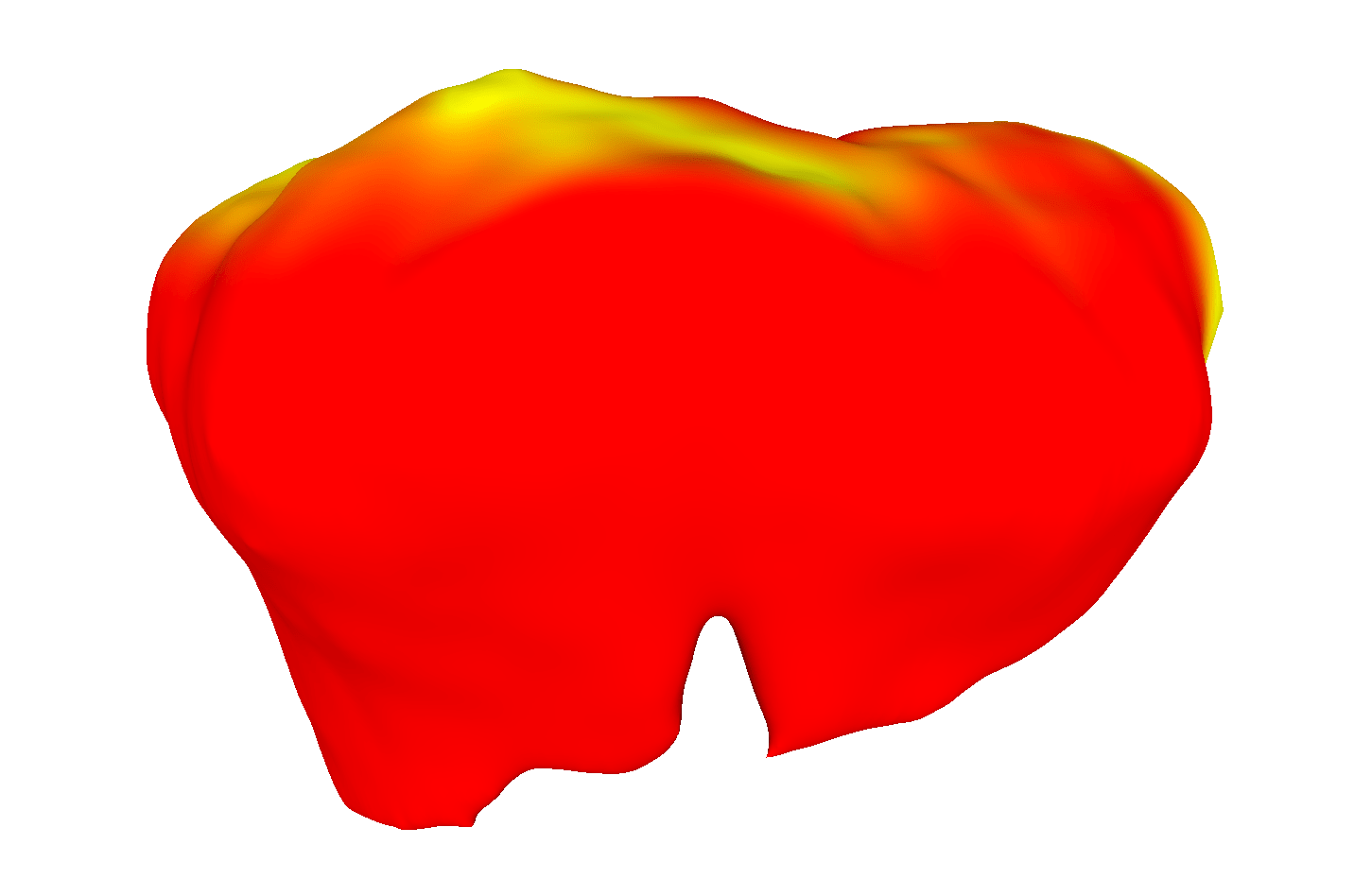}
                \includegraphics[width=1.11in]{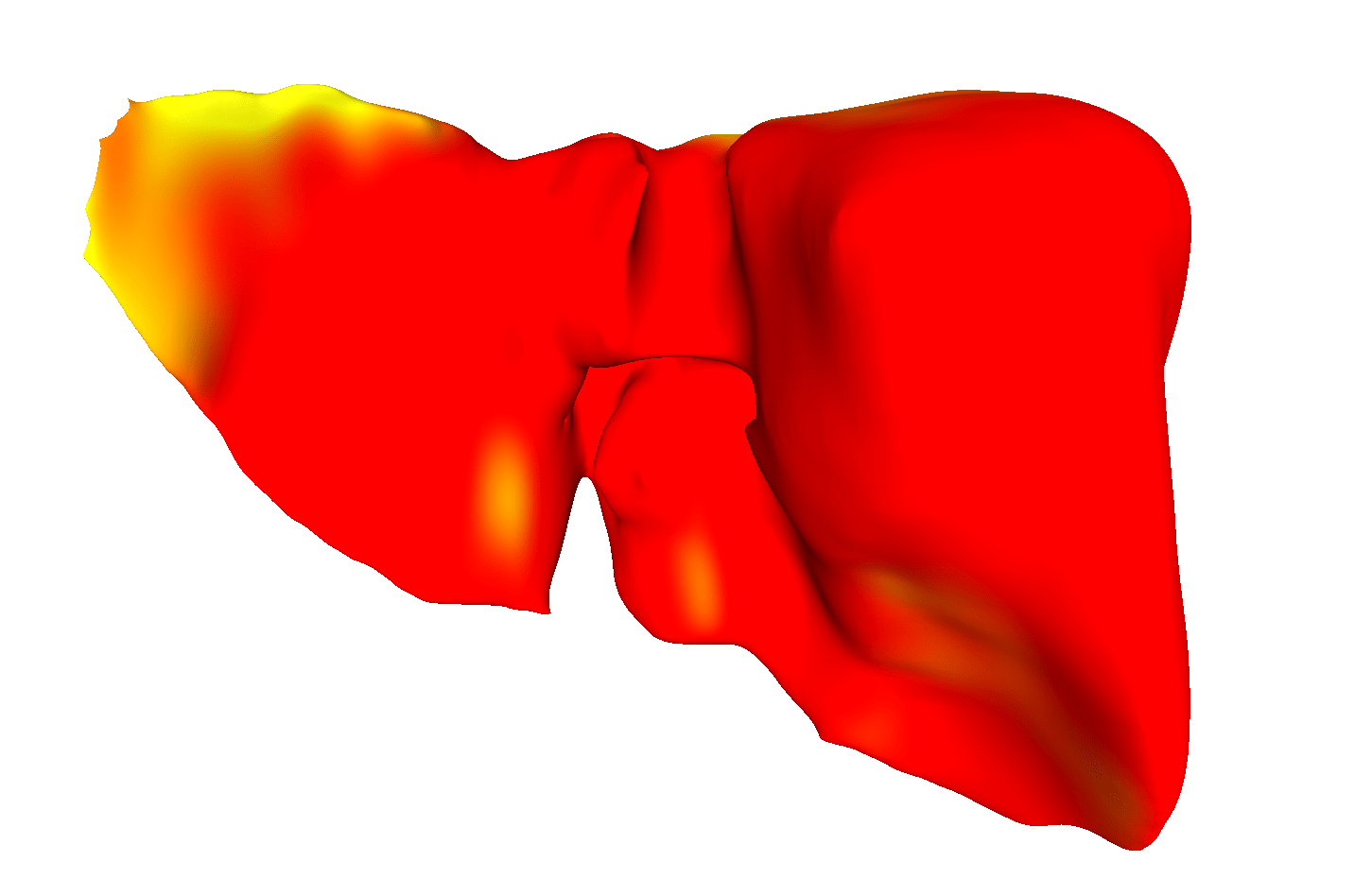}
                \includegraphics[width=1.11in]{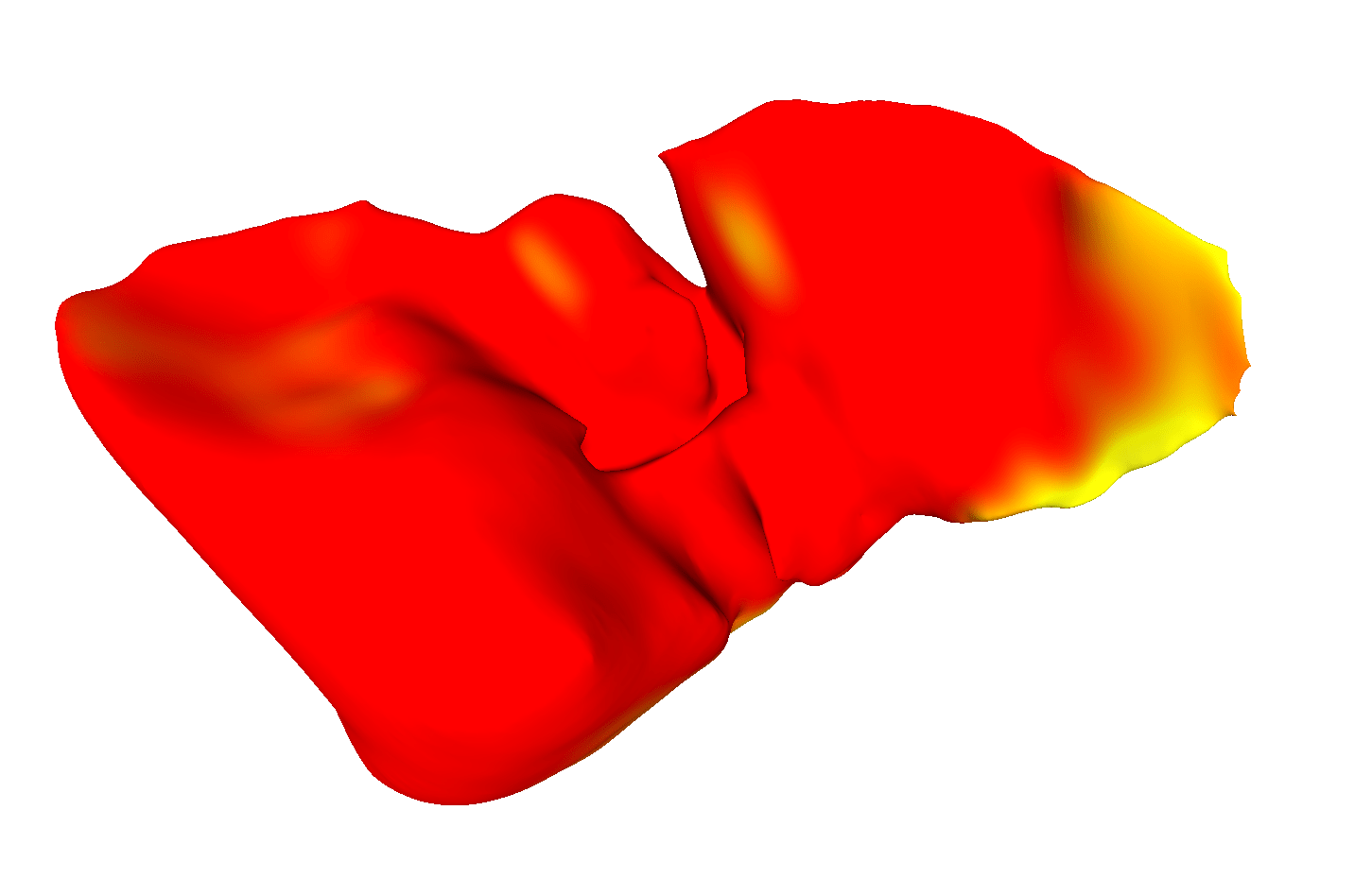}
                \vfil
                \includegraphics[width=1.11in]{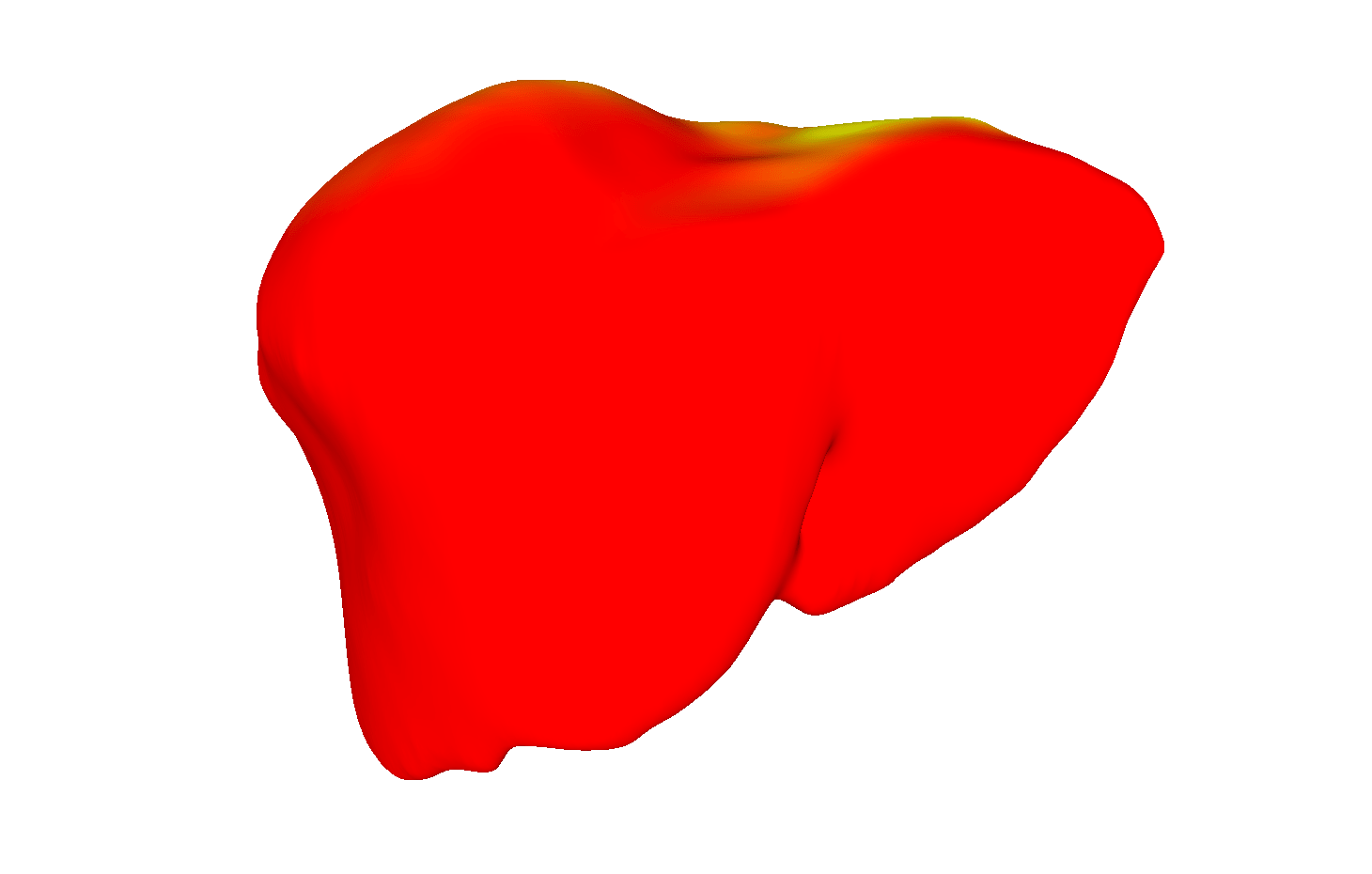}
                \includegraphics[width=1.11in]{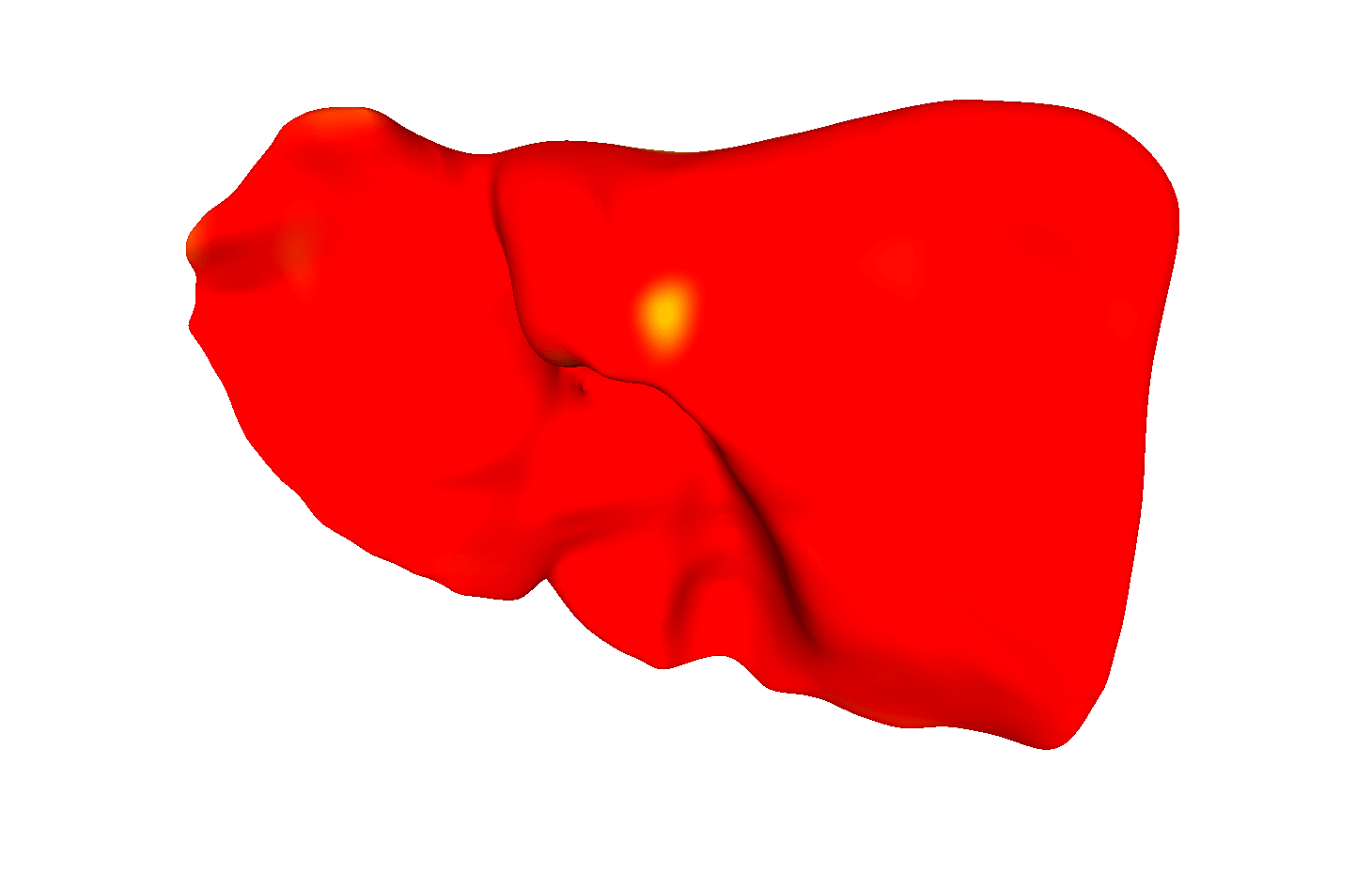}
                \includegraphics[width=1.11in]{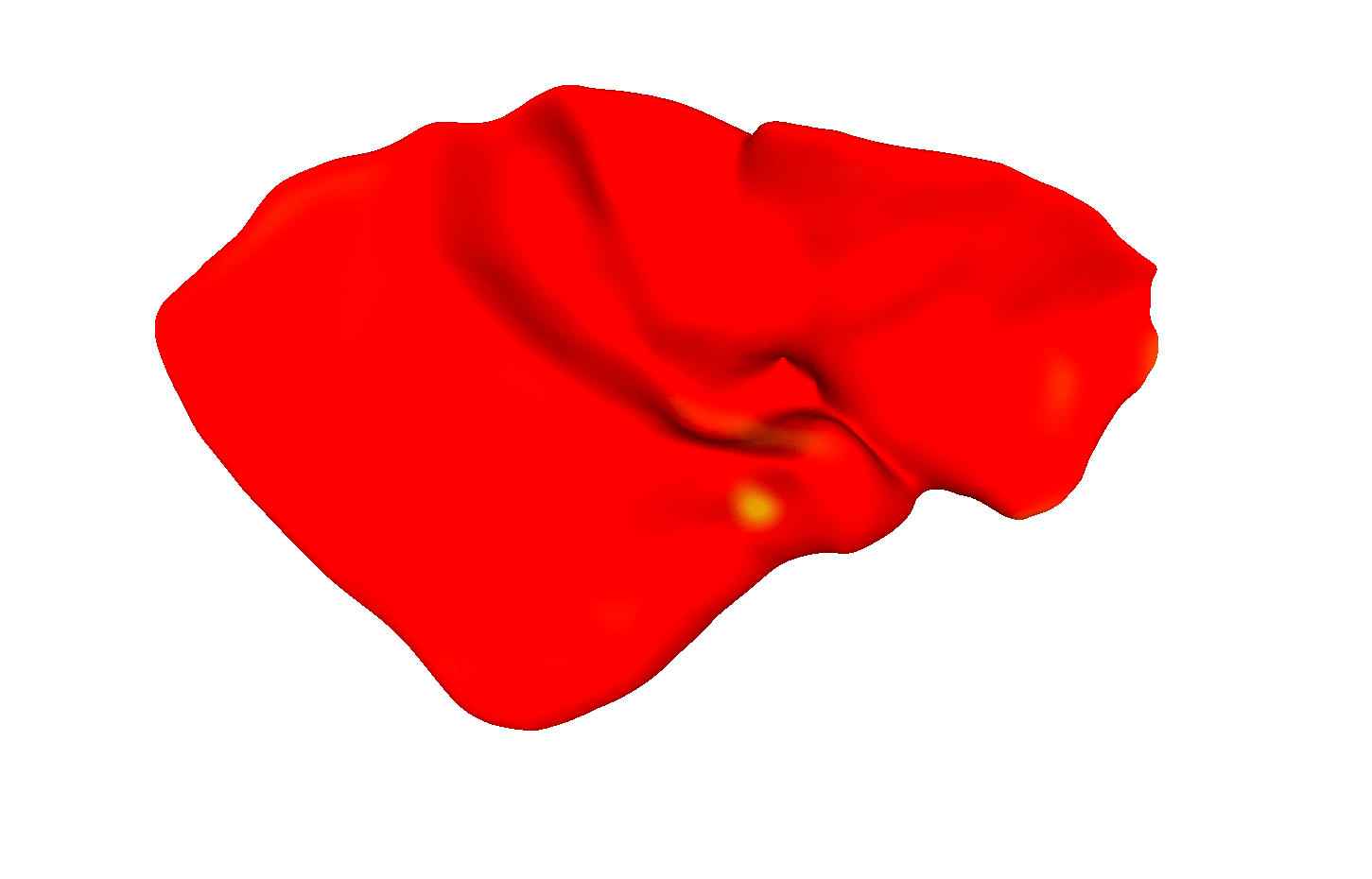}
                \vfil
                \includegraphics[width=1.11in]{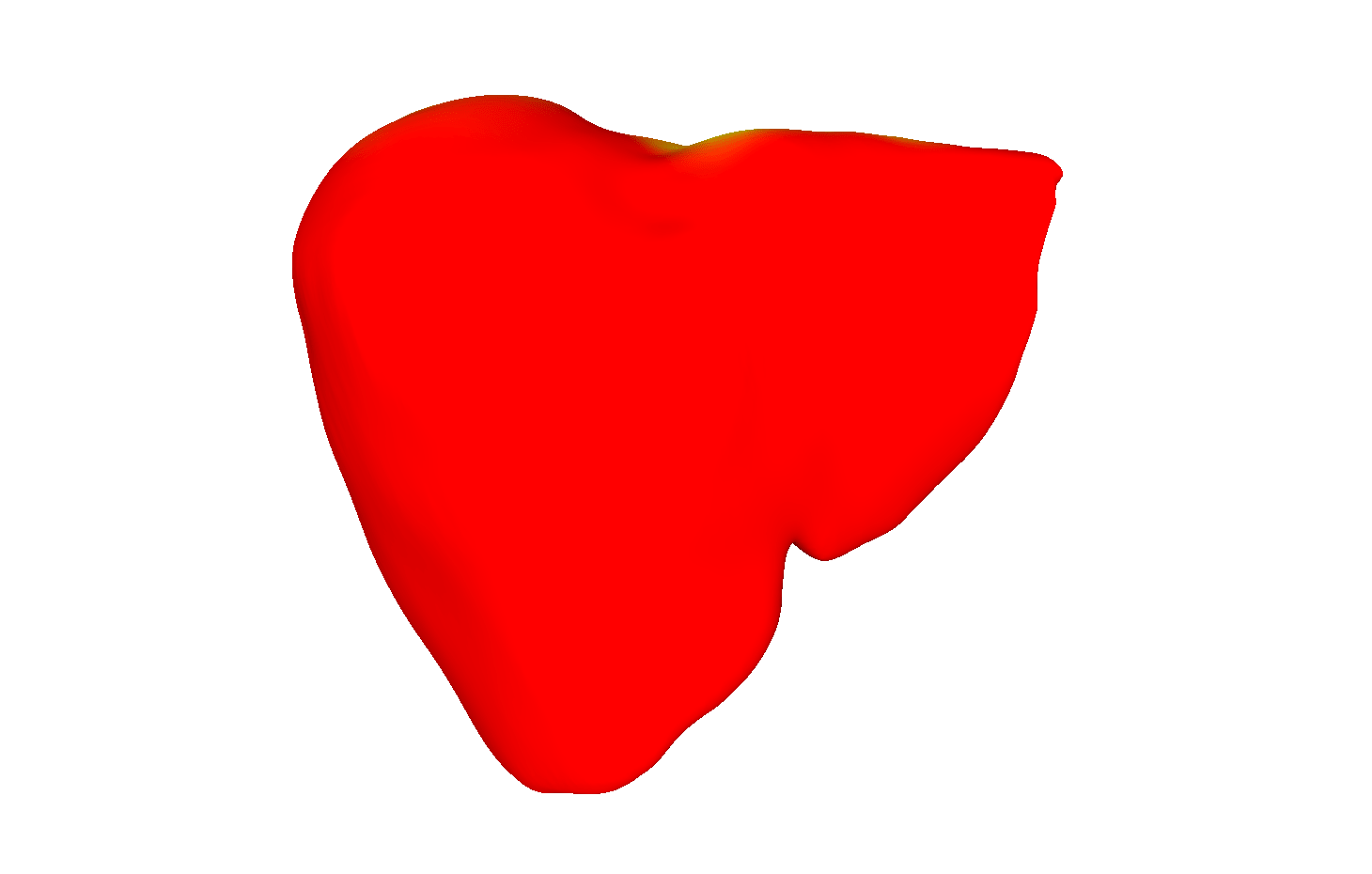}
                \includegraphics[width=1.11in]{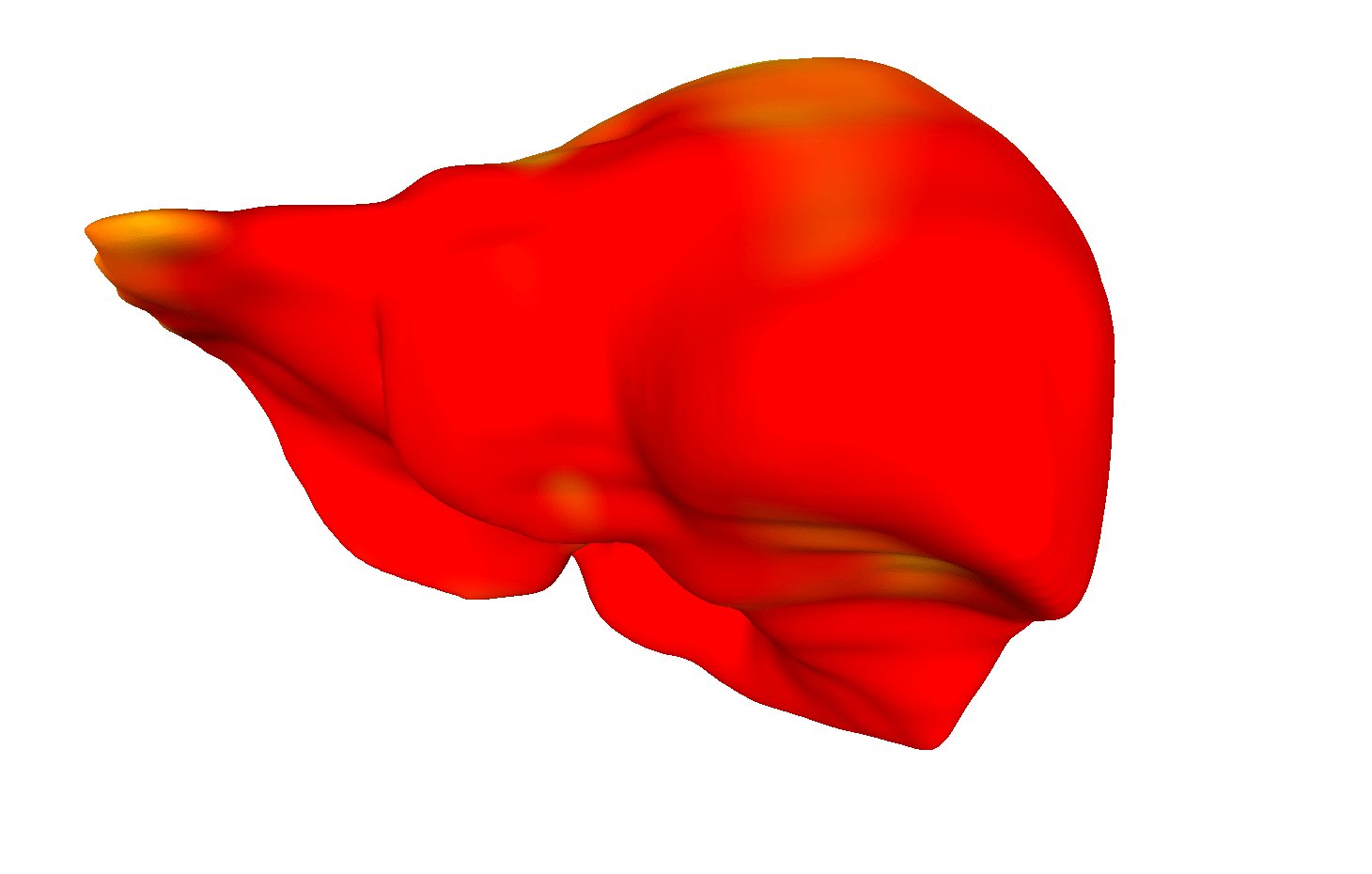}
                \includegraphics[width=1.11in]{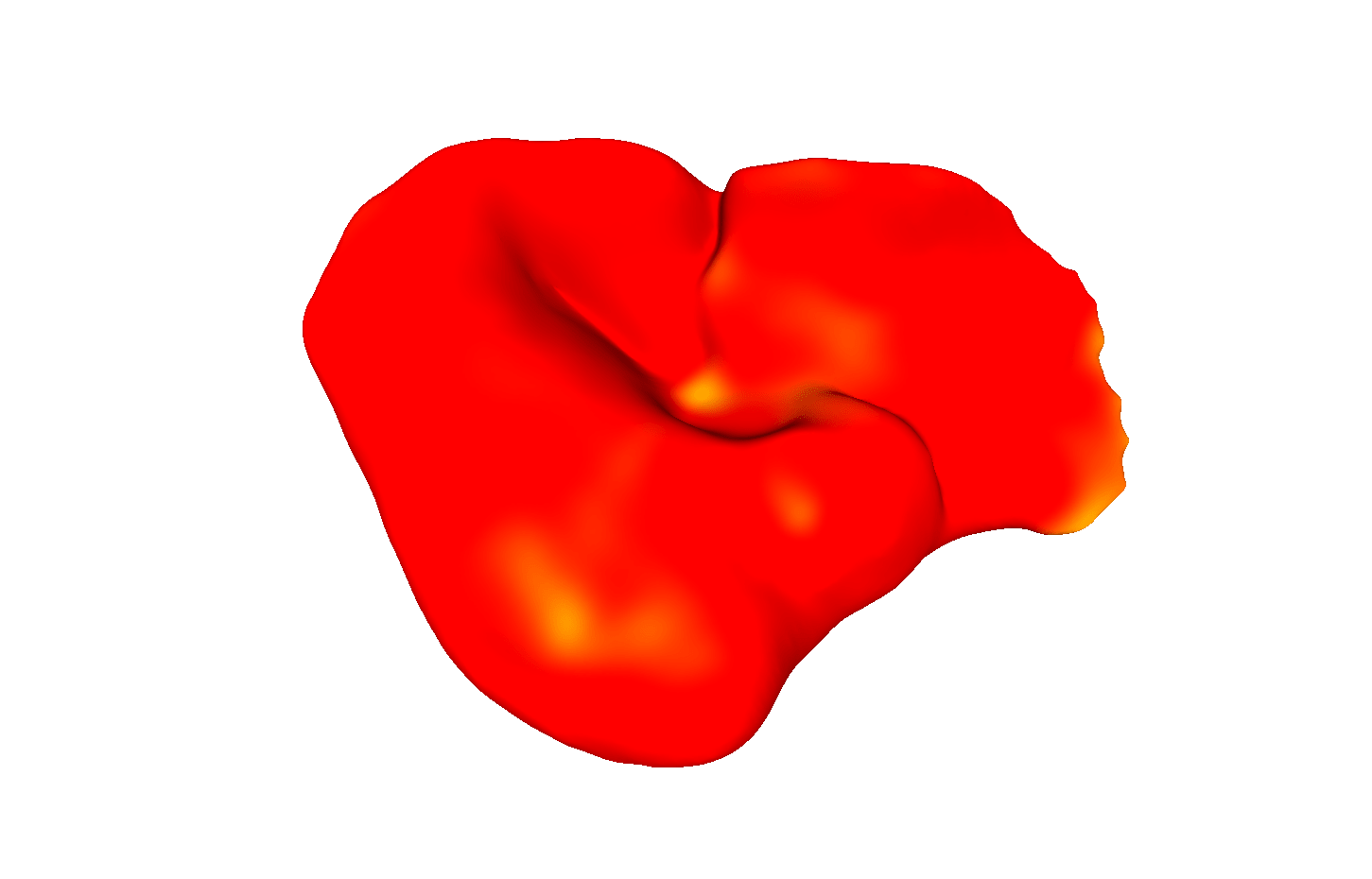}
                \vfil
                \includegraphics[width=1.11in]{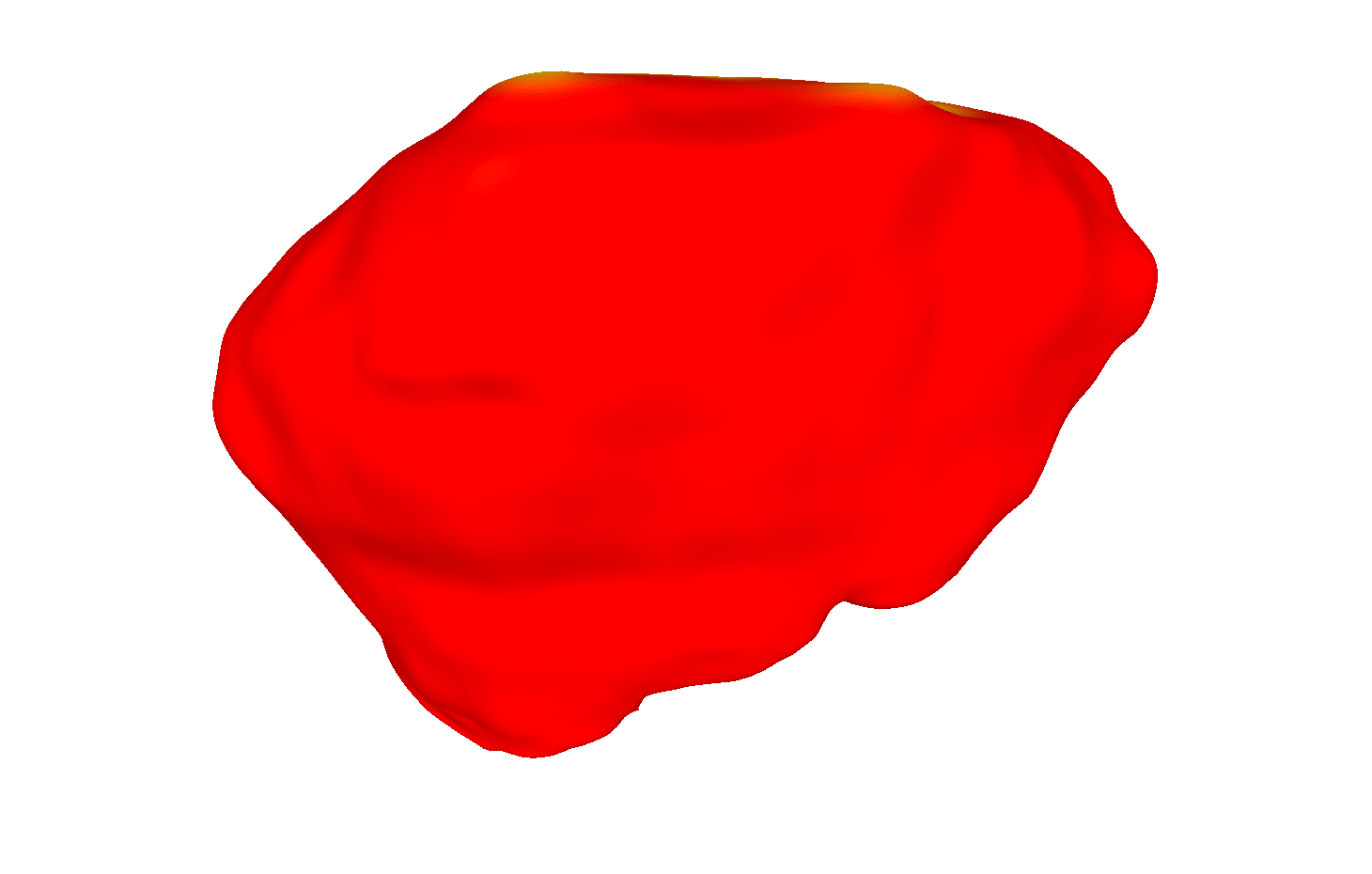}
                \includegraphics[width=1.11in]{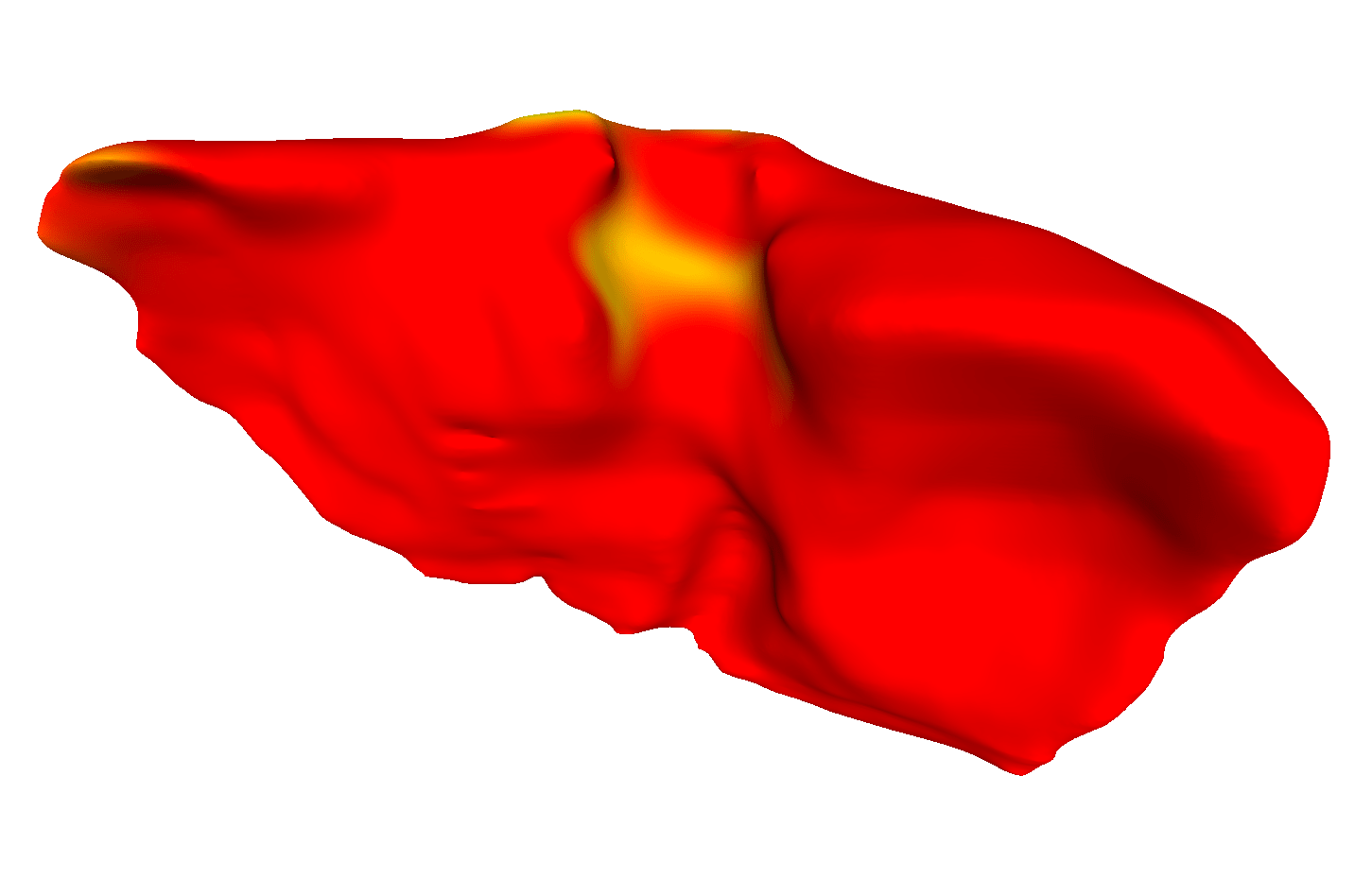}
                \includegraphics[width=1.11in]{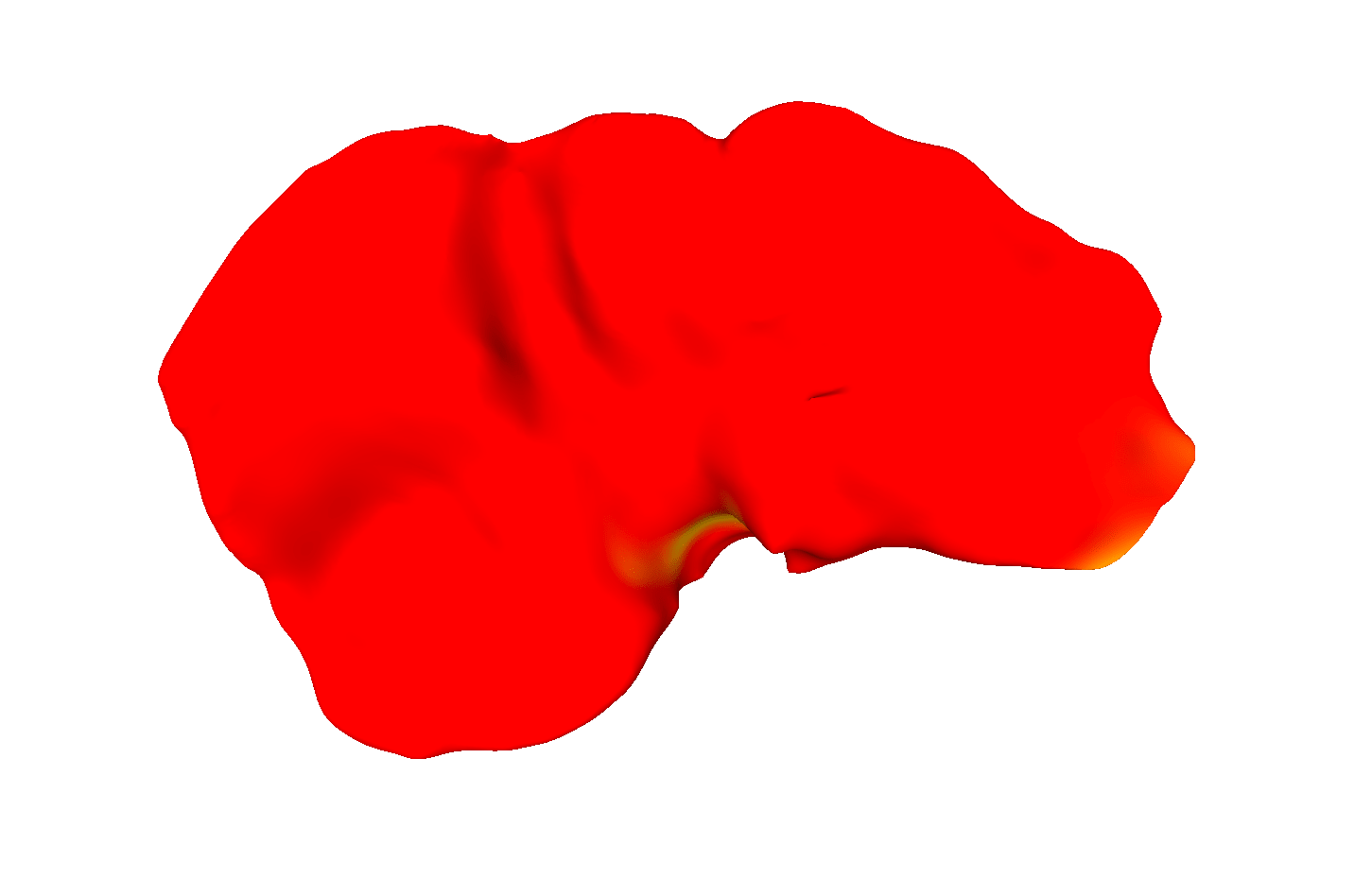}
            \captionof{subfigure}{Fully-supervised contour feature maps.}
            
        \end{minipage}
        \hfil
        \begin{minipage}[b]{3.5in}
            \centering
                \vfil
                \includegraphics[width=1.11in]{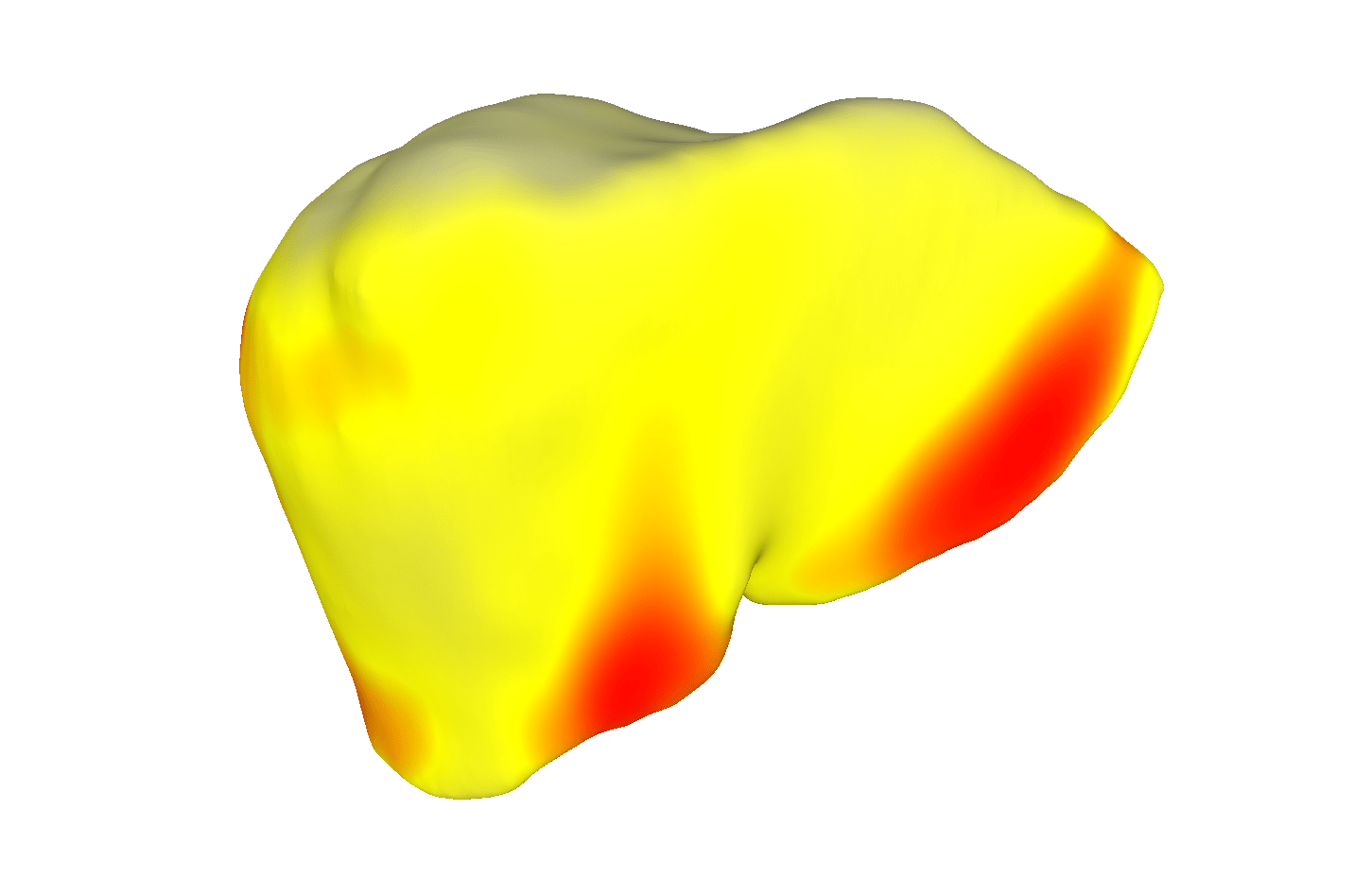}
                \includegraphics[width=1.11in]{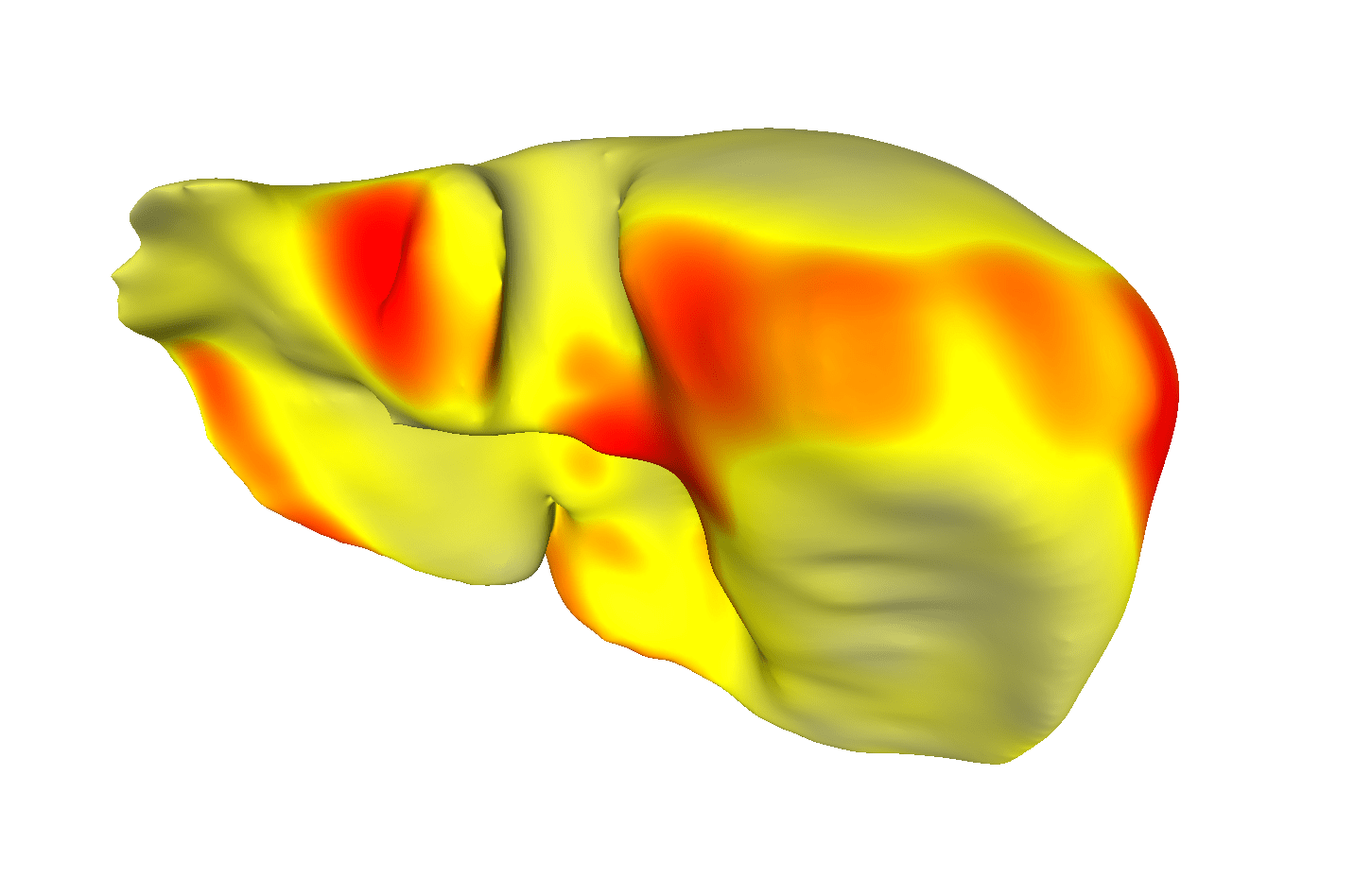}
                \includegraphics[width=1.11in]{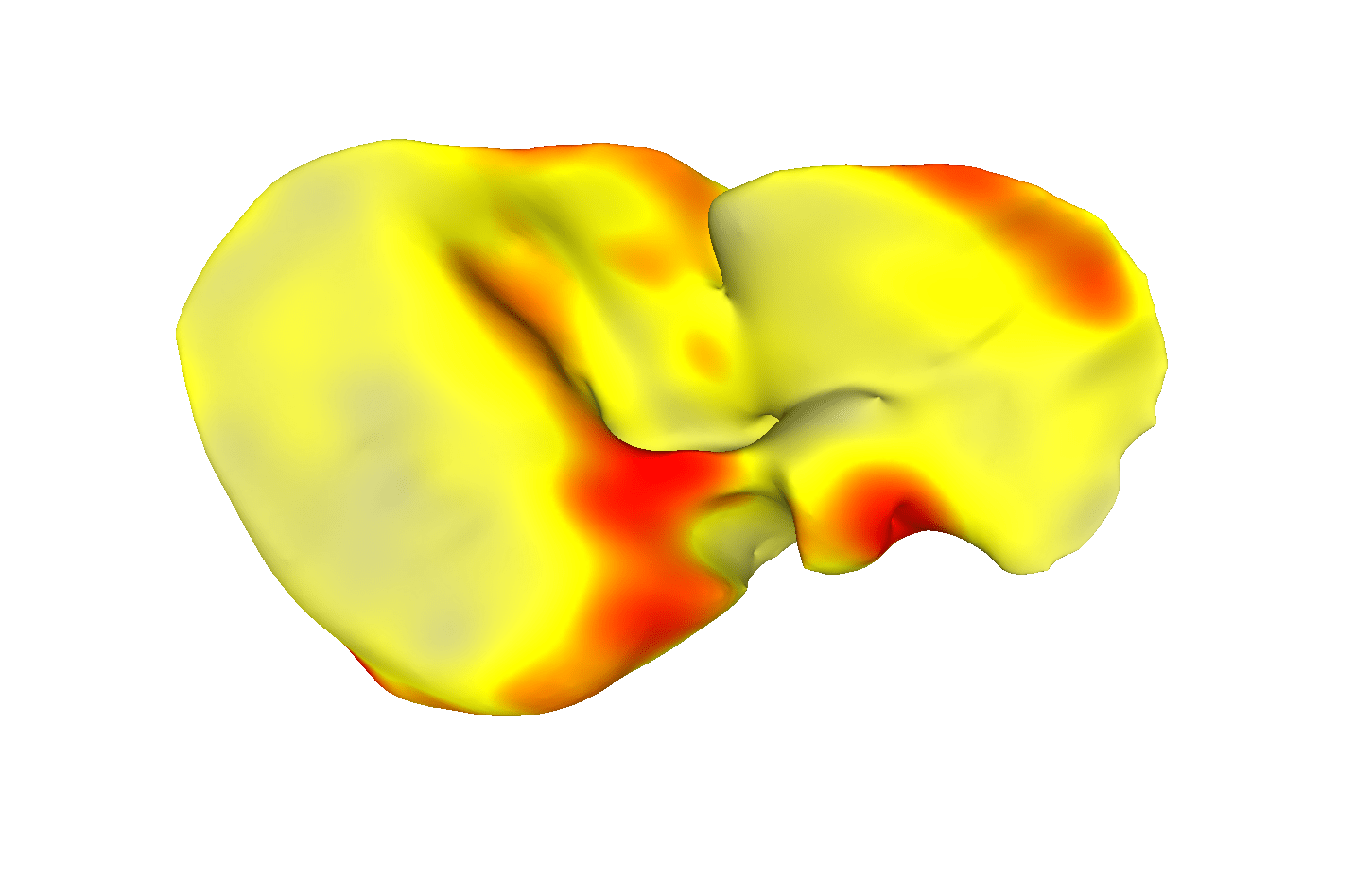}
                \vfil
                \includegraphics[width=1.11in]{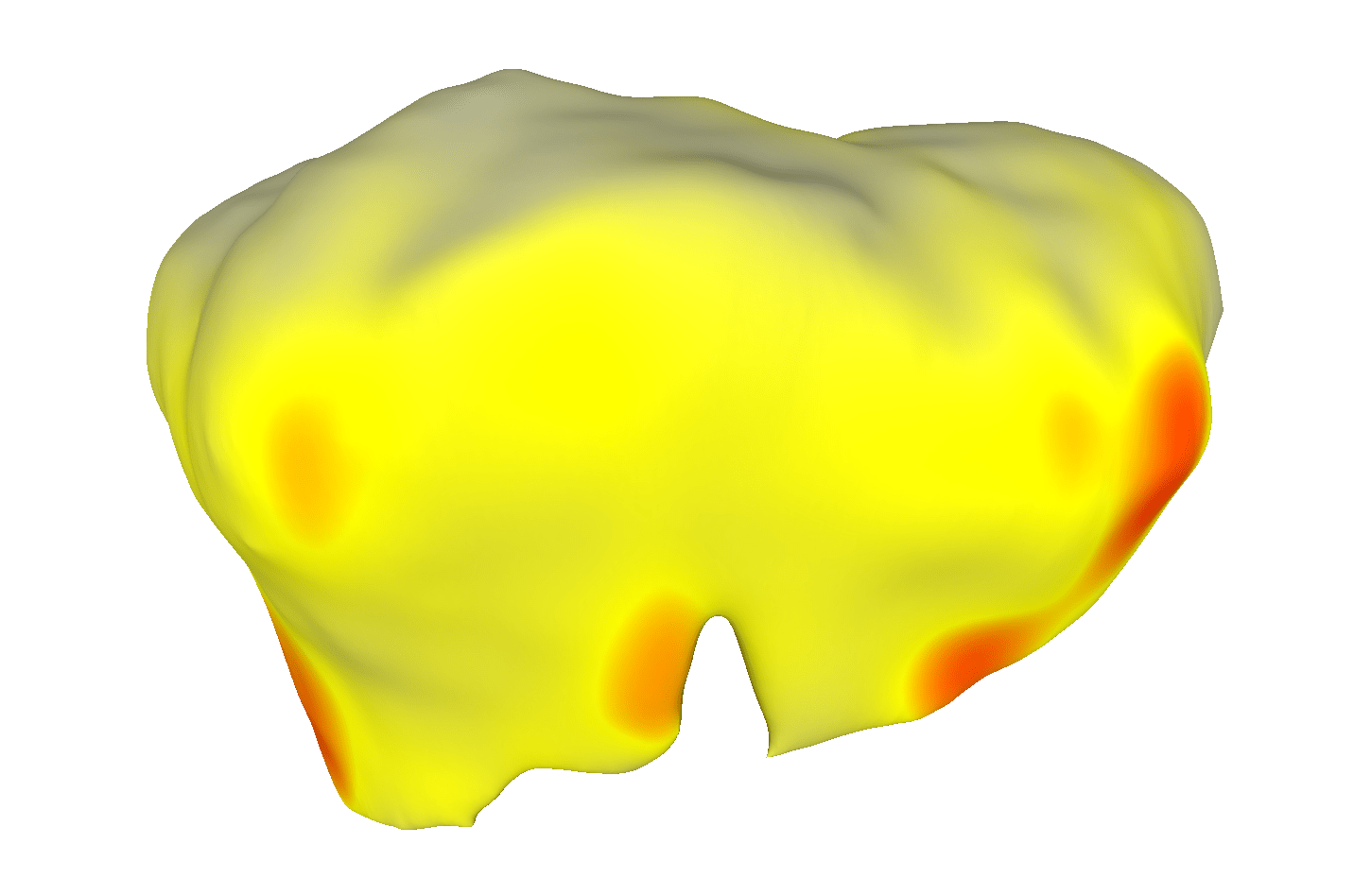}
                \includegraphics[width=1.11in]{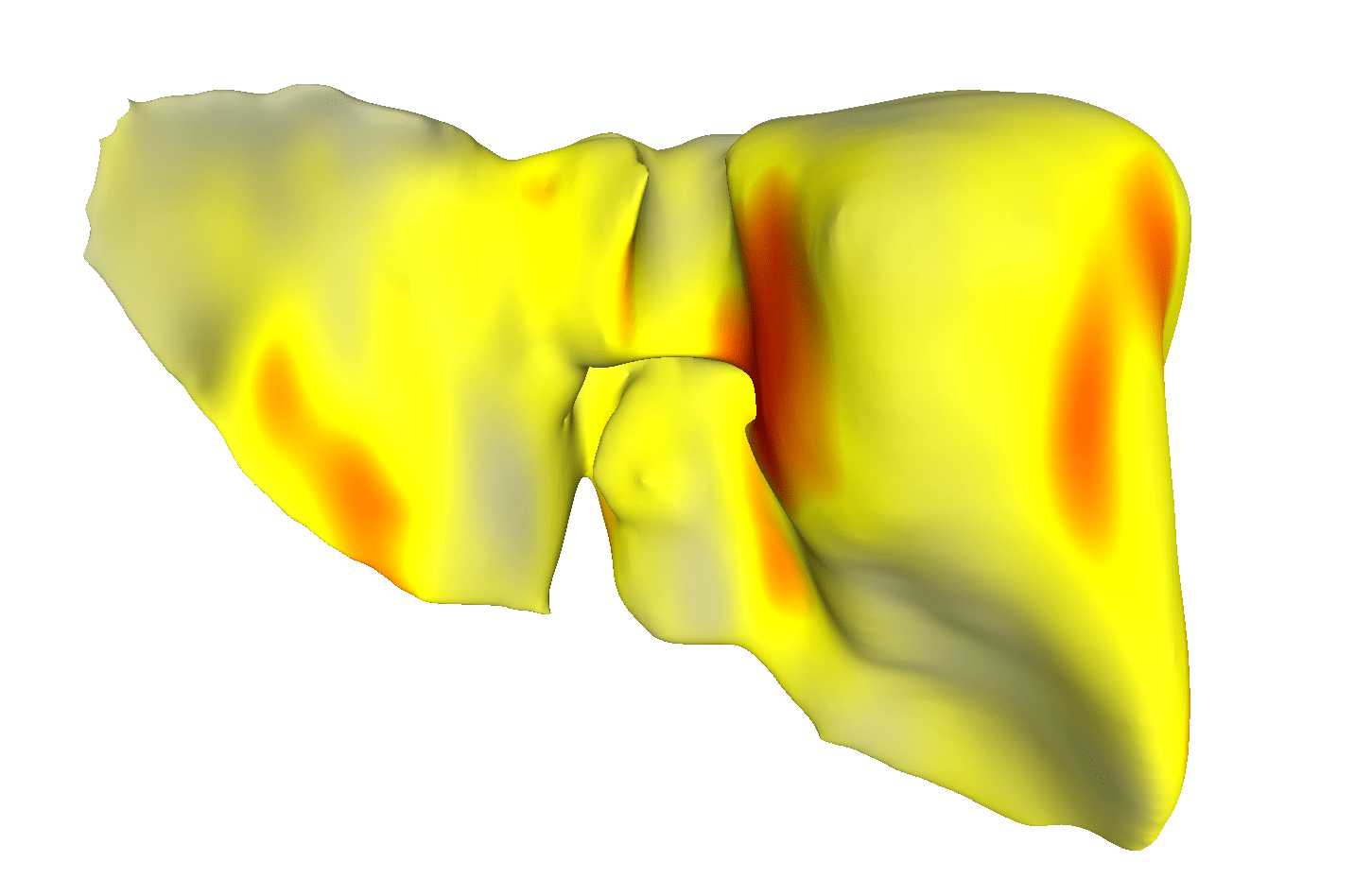}
                \includegraphics[width=1.11in]{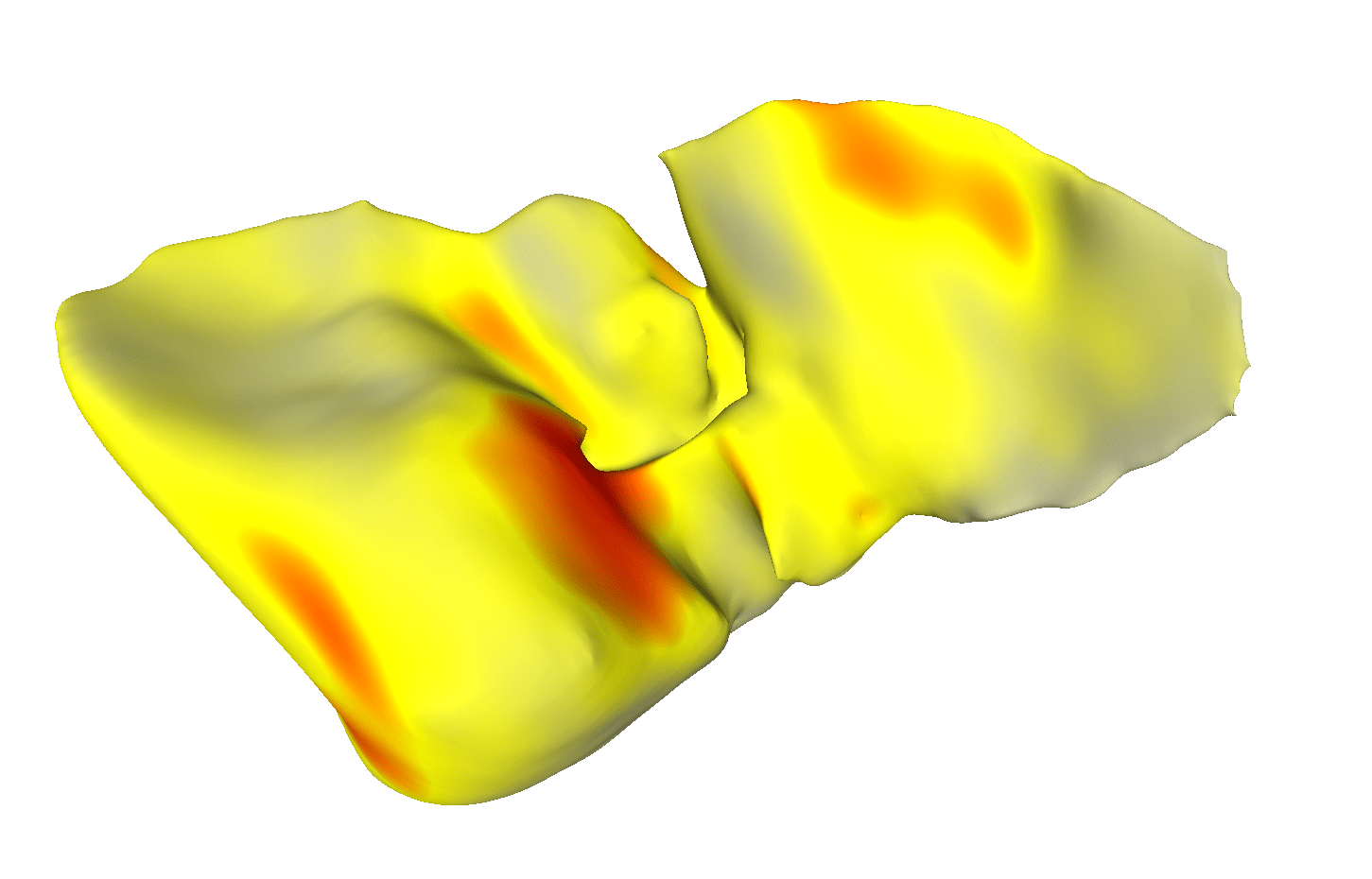}
                \vfil
                \includegraphics[width=1.11in]{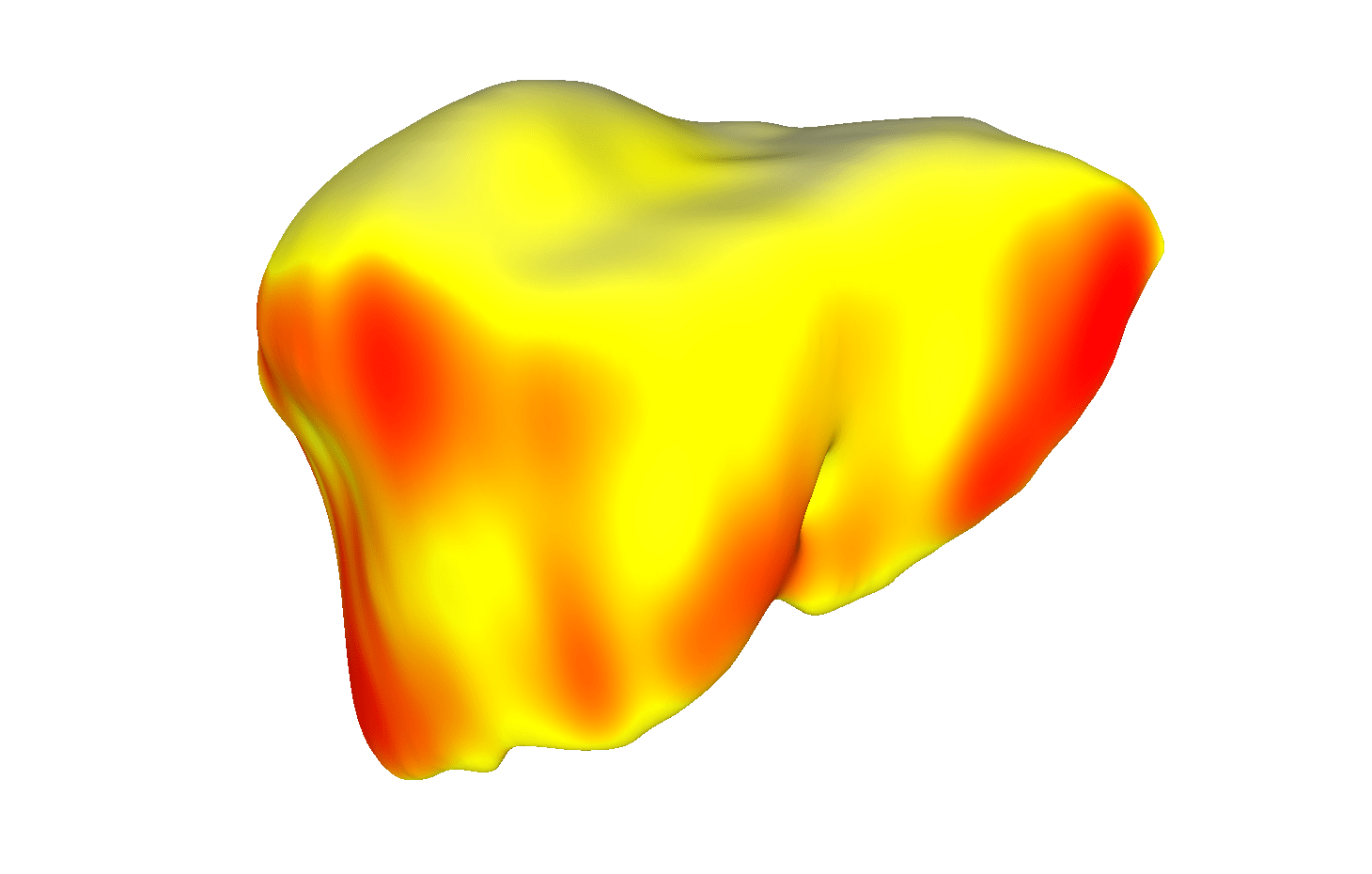}
                \includegraphics[width=1.11in]{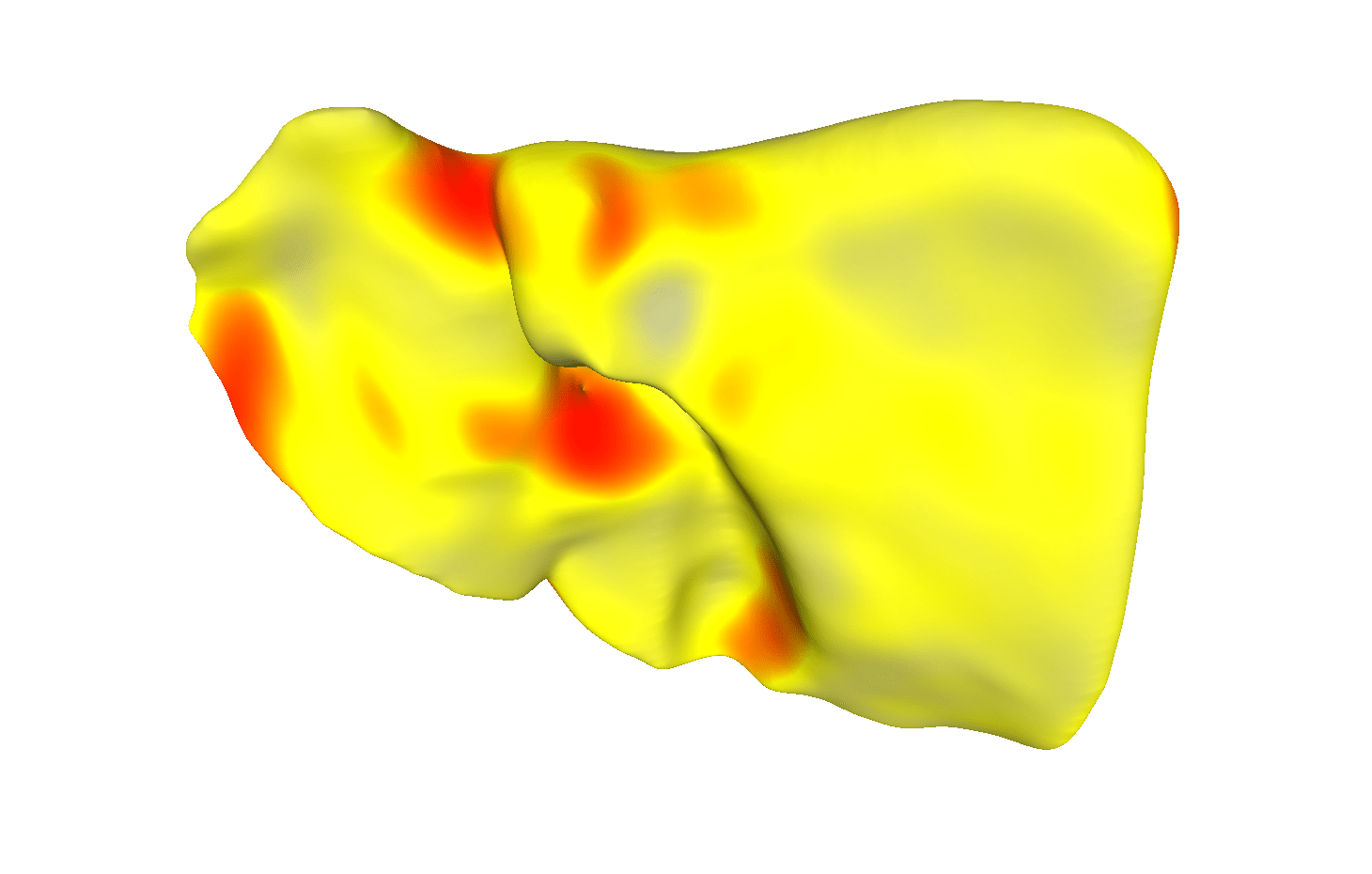}
                \includegraphics[width=1.11in]{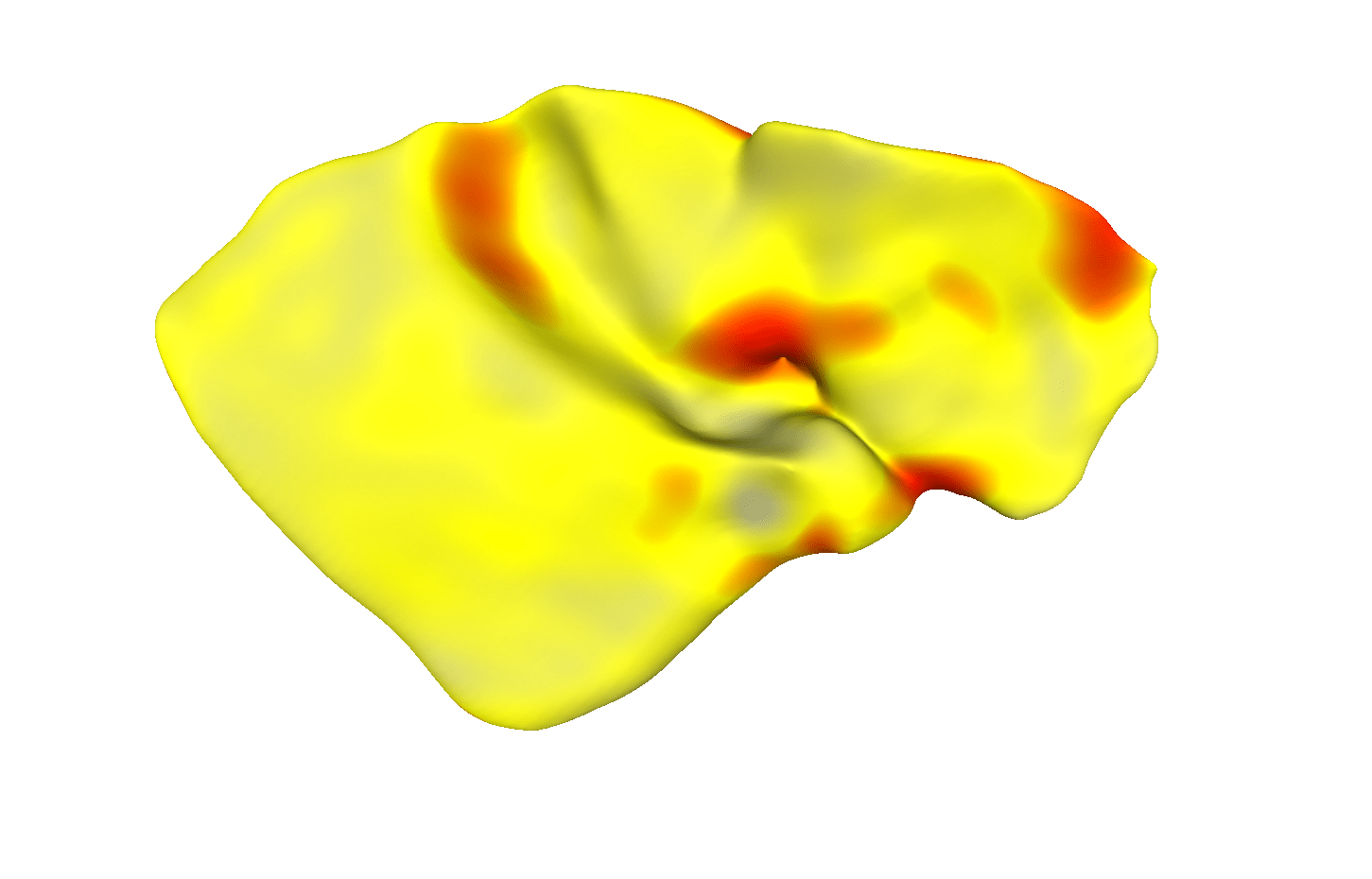}
                \vfil
                \includegraphics[width=1.11in]{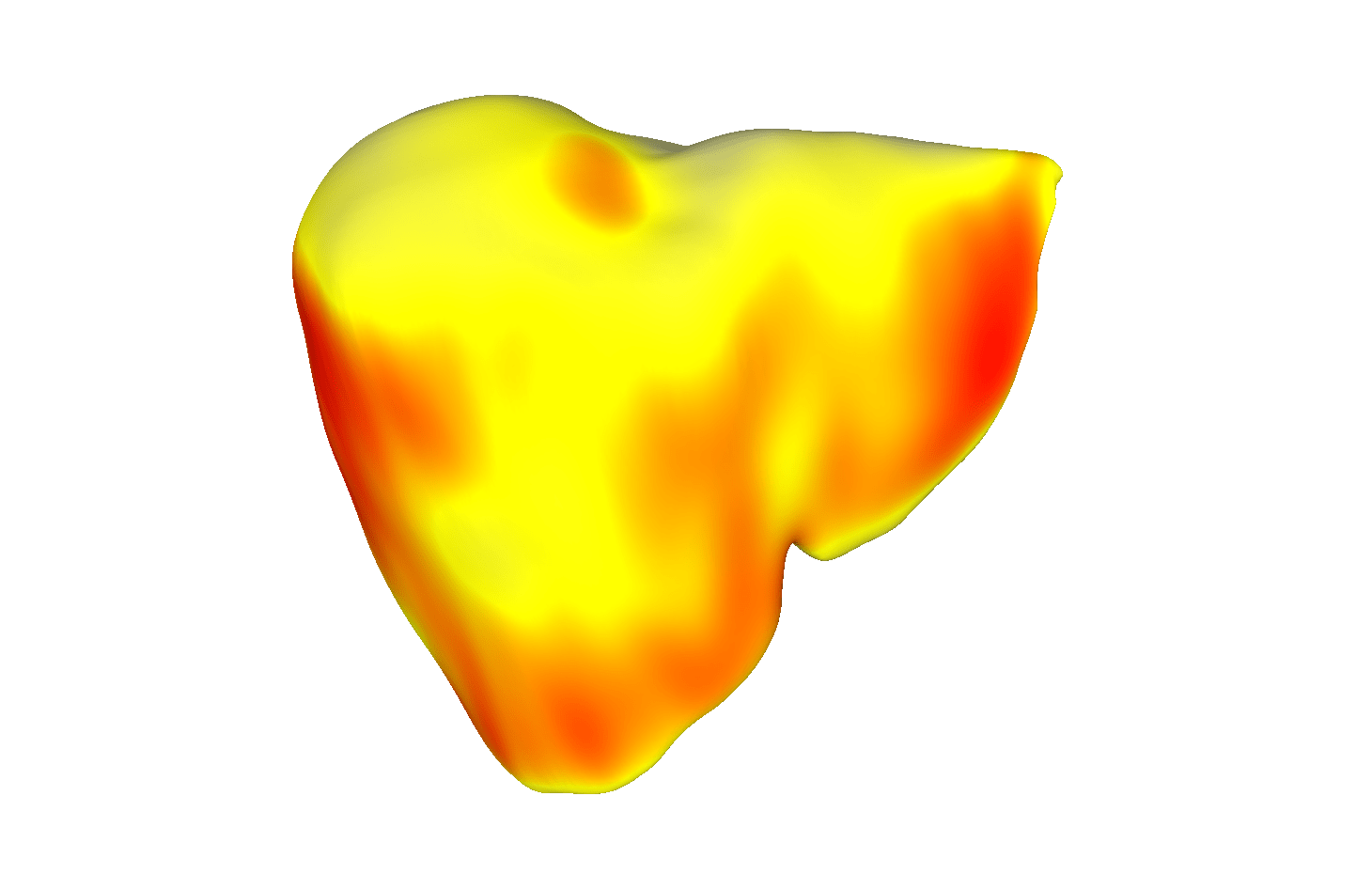}
                \includegraphics[width=1.11in]{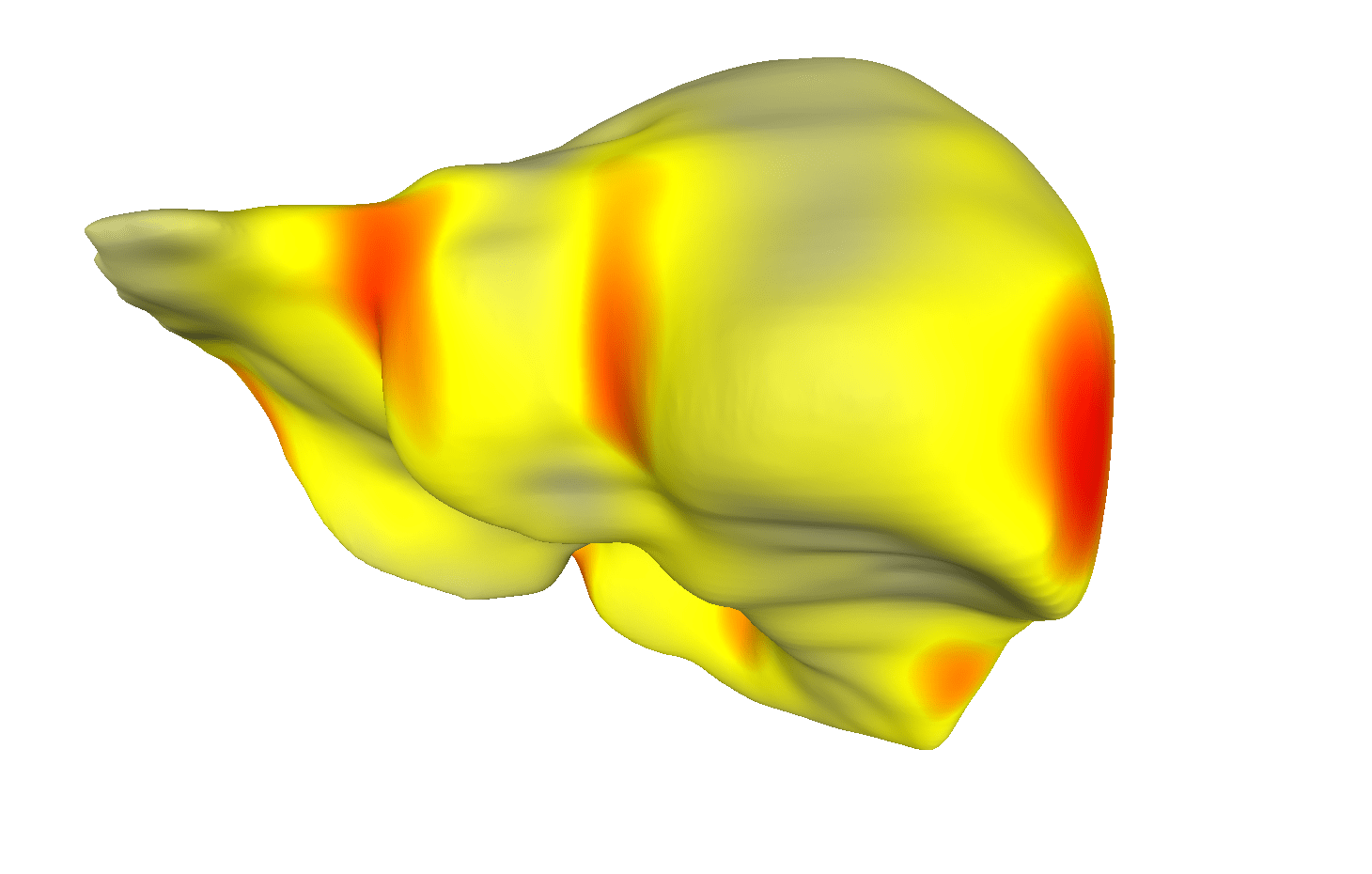}
                \includegraphics[width=1.11in]{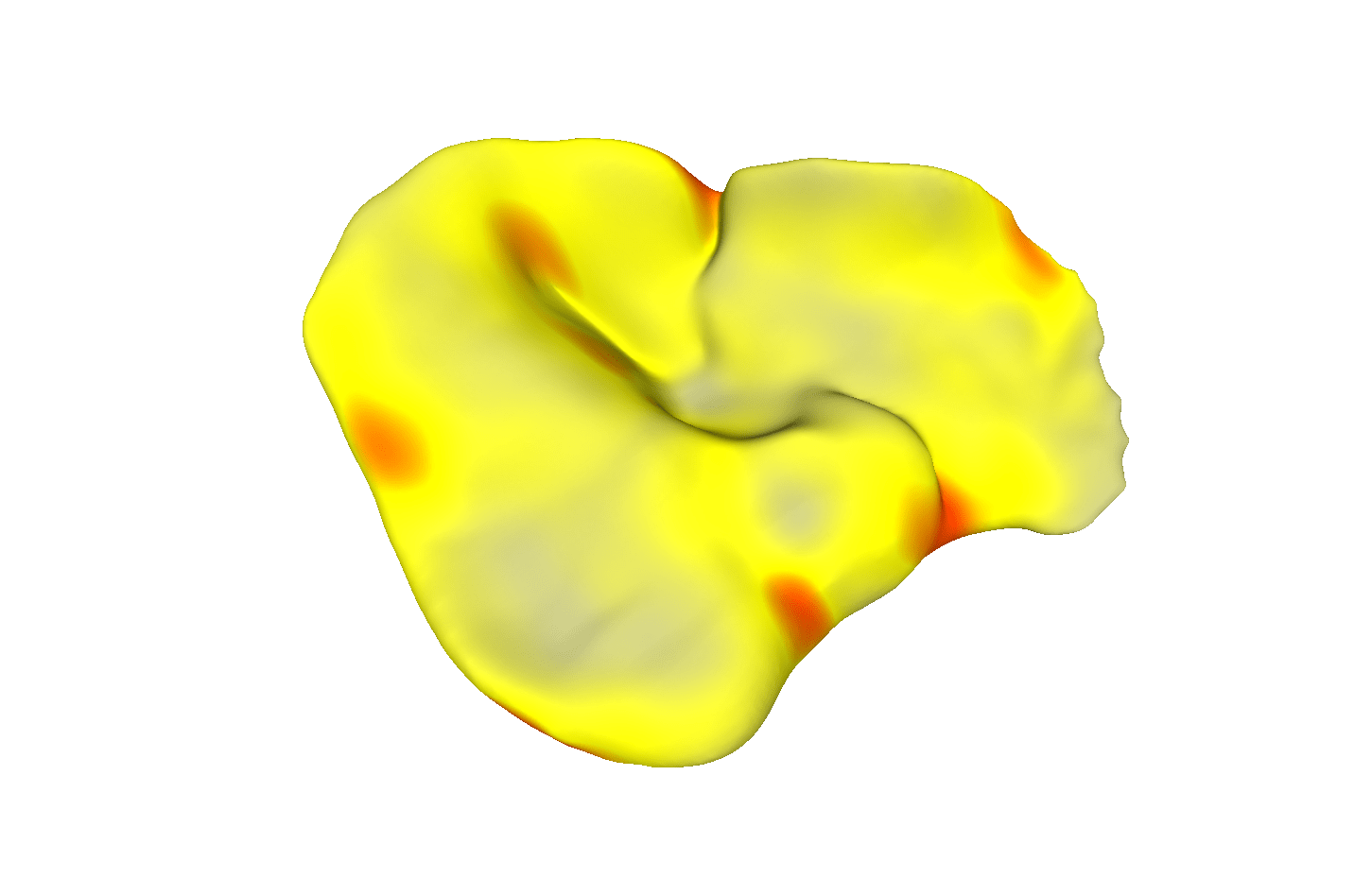}
                \vfil
                \includegraphics[width=1.11in]{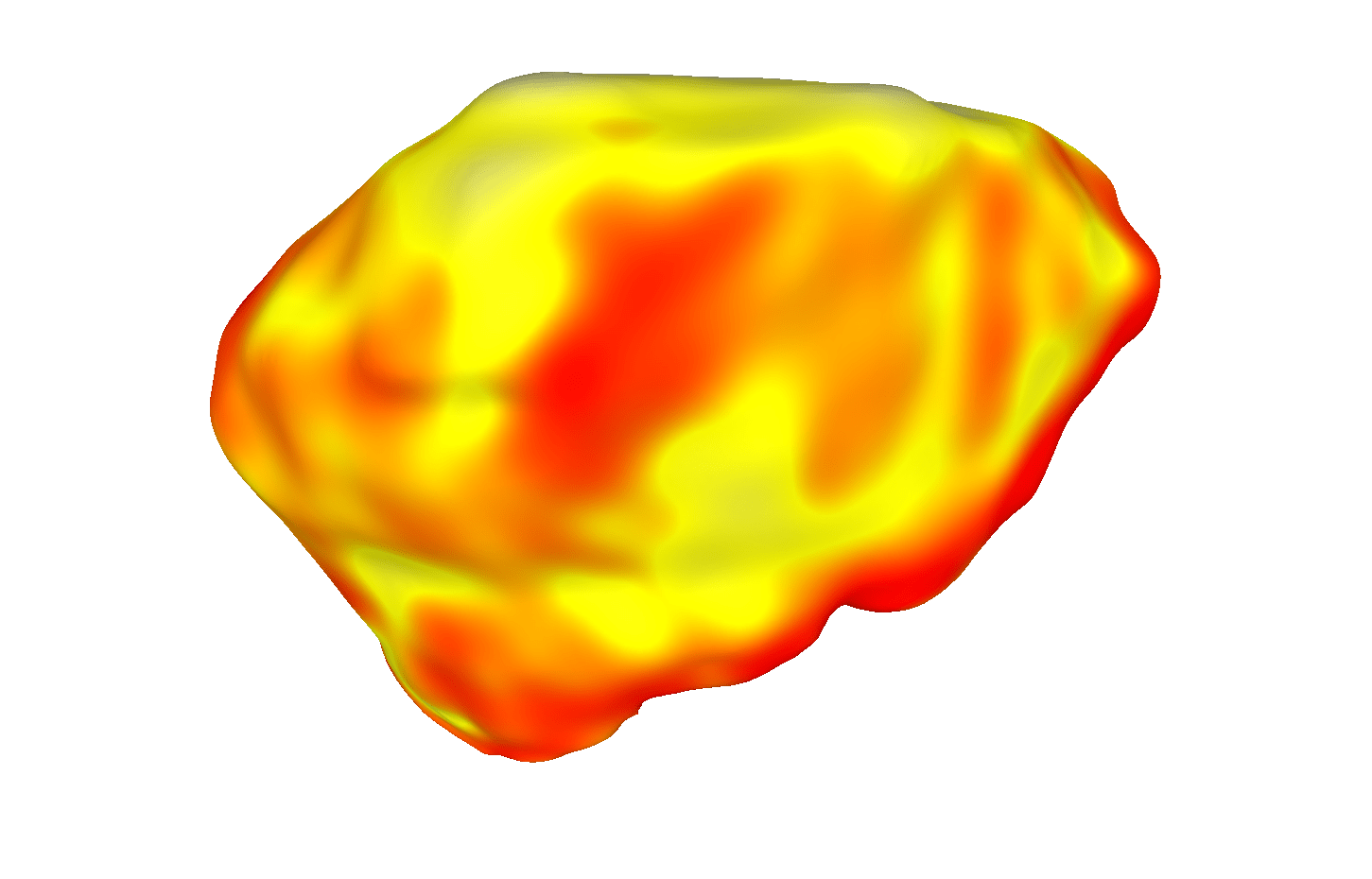}
                \includegraphics[width=1.11in]{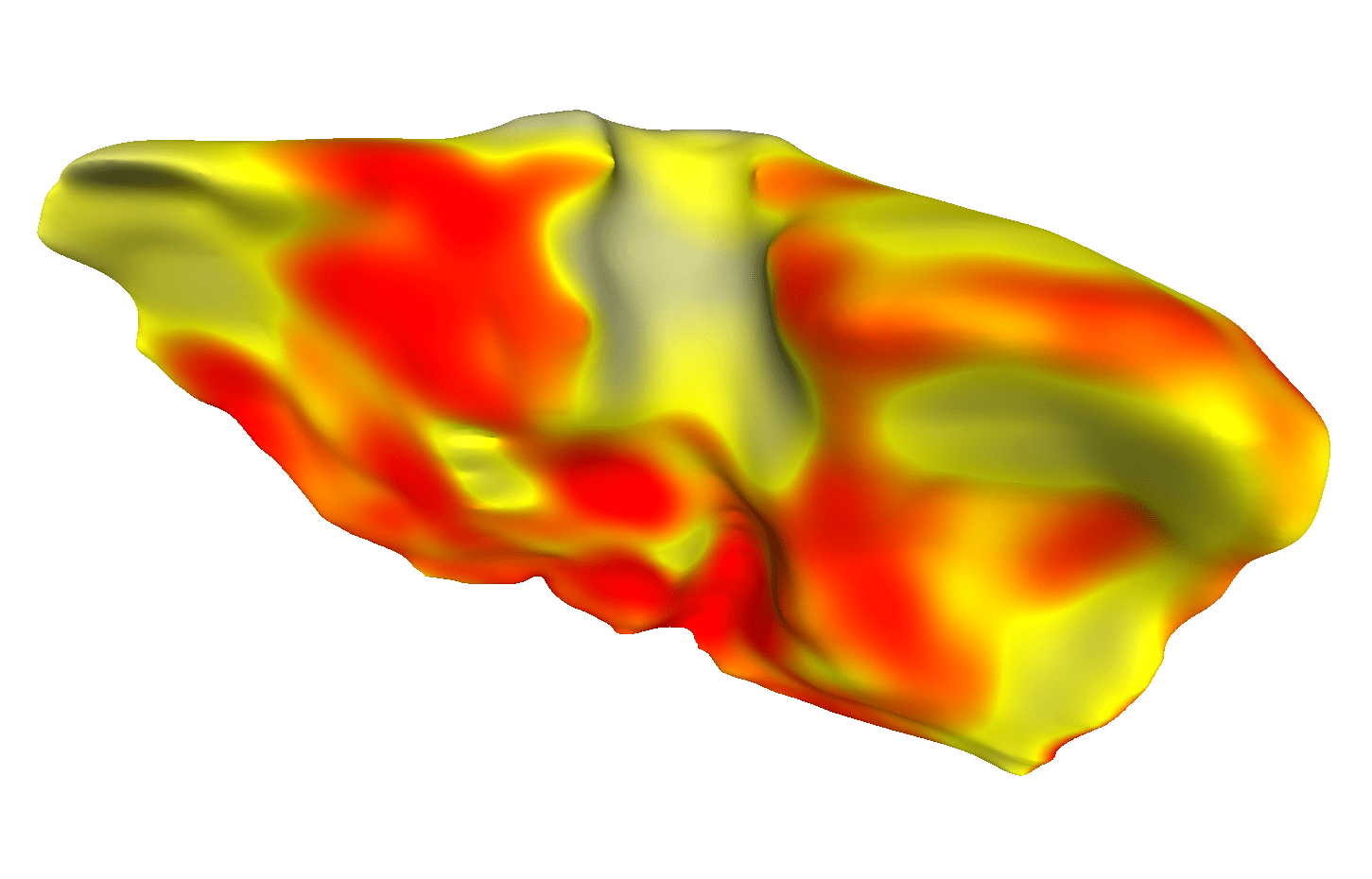}
                \includegraphics[width=1.11in]{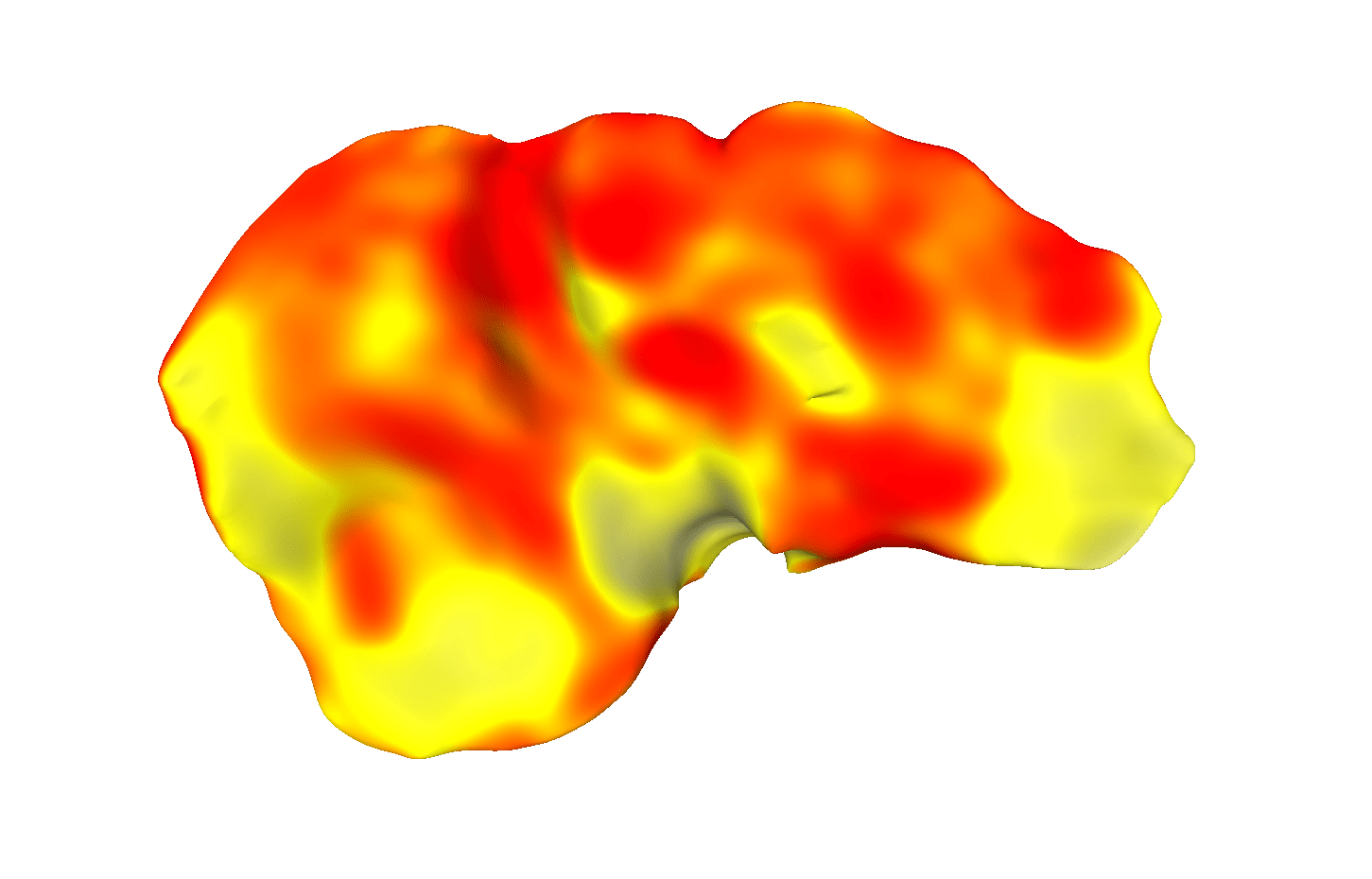}
            \captionof{subfigure}{Self-supervised contour feature maps.}
        \end{minipage}
        \vfil
       
        \includegraphics[width=4.5in]{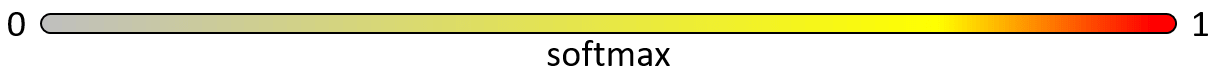}
        
    \caption{Contour feature (i.e., $F_c$) visualizations after full training: (a) without self-supervision and (b) with self-supervision (i.e., (\ref{eq:closs1})). The self-supervised contour feature map in (b) is sparser than that of the full-supervision and is later used as a strong contour features. The ground-truth surface is used for visualizing the distribution of the contour feature. The softmax value of $F_c$ is normalized into the range [0-1].}
    \label{fig:responses}
\end{figure*}

\section{EXPERIMENTS AND RESULTS}

In the proposed experiments, the learning curves and results of the proposed network were evaluated by comparing them with those of other FCN-based models. A DSN \cite{dou20173d}, VoxResNet \cite{chen2017voxresnet}, DenseVNet \cite{gibson2018automatic}, and the proposed network, CENet, were used for performance evaluation.

\subsection{Learning Curve}
A learning curve with the dice loss is plotted in Fig. \ref{fig:learningcurve}. All hyperparameters (such as learning rate and optimizer) were set as specified in the original studies. An eight-fold cross-validation was first designed for performance evaluation (i.e., 140 training images and 20 validation images). The plot in Fig. \ref{fig:learningcurve_8fold} indicates that our proposed network achieved the most successful training result. The other networks could not minimize the validation errors. The quantitative results are presented in Tables \ref{table:results} and \ref{table:results_stat}. A special experimental setting was have additionally designed with 10 training images and 150 validation images (Figs. \ref{fig:learningcurve_sfold_wcutout} and \ref{fig:learningcurve_sfold_wocutout}). This experimental setting approximately proxies the real-life deep learning problem and shows an extremely generalized regularization analysis. The overall validation errors increased in a special cross-validation with 10 training images (Fig. \ref{fig:learningcurve_sfold_wcutout}). Moreover, the proposed network did not over-fit (i.e., lowest generalization error) to the training images compared to other networks. Fig. \ref{fig:learningcurve_sfold_wocutout} shows the least accurate generalization curve without a cutout augmentation \cite{devries2017improved}, indicating that the cutout augmentation greatly aids the network training to be generalized. Comparing all training experiments, the proposed network made the fastest convergence, showed the lowest loss value, and resulted in the best generalization.

        
        
        
        
        
        

\begin{figure}[!tb]
    
    \centering
        \subfloat[\(F^0_s\).]{\includegraphics[width=1.05in]{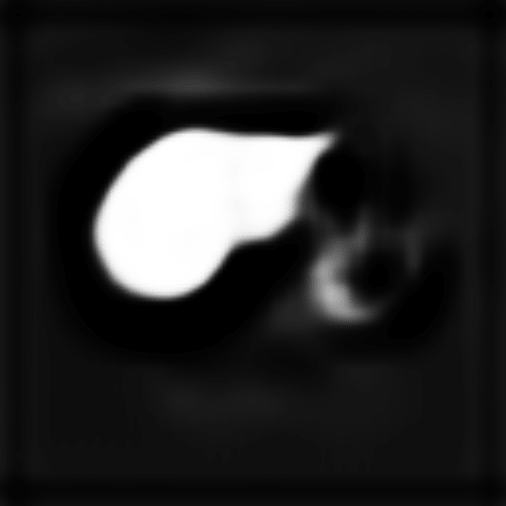}
        \label{fig:residual1}}\hfil
        \subfloat[\(F^1_s\).]{\includegraphics[width=1.05in]{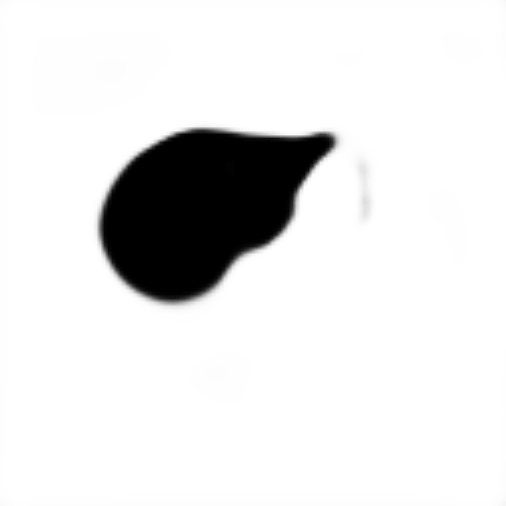}
        \label{fig:residual2}}\hfil
        \subfloat[\(F^0_s-F^1_s\).]{\includegraphics[width=1.05in]{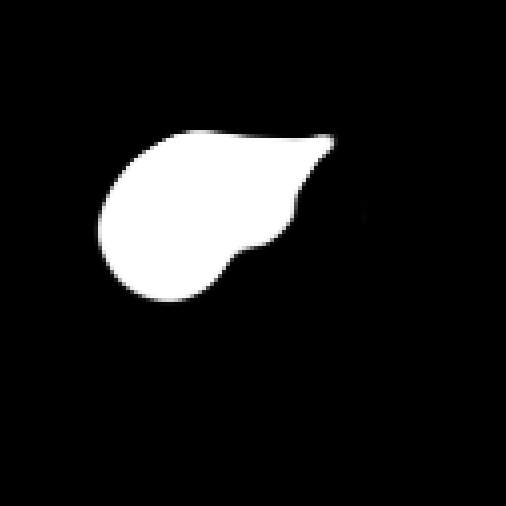}
        \label{fig:residual}}
    \caption{Visualization of the shape feature maps (i.e., \(F^i_s\)) after full training. (a) \(F^0_s\), (b) \(F^1_s\), and (c) shows the subtraction result of the \(F^i_s\) features.}
    \label{fig:residuals}
\end{figure}

\subsection{Contour and Shape Feature Layers}

The output feature map of the contour transition layer (i.e., $F_c$) is displayed in Fig. \ref{fig:responses}. The contour feature map of a fully supervised network (i.e., using ground-truth contour supervision without modification (\ref{eq:closs1})) was activated within all contour regions (Fig. \ref{fig:responses}a). Fig. \ref{fig:responses}a demonstrates that even with full training, the network failed to extract full contour features accurately (i.e., a part of the low softmax responses on the ground-truth contour region). Moreover, with a self-supervised network, the contour feature map was activated in the local contour regions that can further aid the accuracy of the segmentation (Fig. \ref{fig:responses}b). As shown in Fig. \ref{fig:responses}b, the contour transition layer successfully learned discriminative contours excluding ambiguous regions that can be better delineated by global shape prediction (i.e., $F^0_s-F^1_s$, presented in Fig. \ref{fig:residuals}). The quantitative evaluation between the two methods is presented in the following section.\par
The effects of the residuals in the shape transition layers are shown in Fig. \ref{fig:residuals}. Both shape transition layers learned complementary features (Figs. \ref{fig:residual1} and \ref{fig:residual2}) for accurate shape delineation by subtraction.


\begin{table*}[t]
\renewcommand{\arraystretch}{1.7}
\captionsetup{justification=centering, labelsep=newline}
\caption{QUANTITATIVE EVALUATION OF THE EIGHT-FOLD CROSS-VALIDATION WITH MEDIAN METRIC}
\label{table:results}
\begin{tabularx}{\textwidth}{X||>{\centering\arraybackslash}X|>{\centering\arraybackslash}X|>{\centering\arraybackslash}X|>{\centering\arraybackslash}X|>{\centering\arraybackslash}X}
Methods & DSC & HD [mm] & ASSD [mm] & Sensitivity & Precision\\
\hline
DSN \cite{dou20173d} & \(0.94\) & \(5.60\) & \(1.60\) & \(0.95\) & \(0.94\)\\
VoxResNet \cite{chen2017voxresnet} & \(0.93\) & \(6.71\) & \(1.78\) & \(0.92\) & \(0.96\)\\
DenseVNet \cite{gibson2018automatic} & \(0.92\) & \(10.13\) & \(2.27\) & \(\textbf{0.97}\) & \(0.88\)\\
CENet & \(\textbf{0.96}\) & \(\textbf{3.99}\) & \(\textbf{1.20}\) & \(\textbf{0.97}\) & \(\textbf{0.97}\)\\
CENet-A & \(\textbf{0.96}\) & \(4.97\) & \(1.23\) & \(\textbf{0.97}\) & \(0.96\)\\
CENet-C & \(\textbf{0.96}\) & \(4.56\) & \(\textbf{1.21}\) & \(\textbf{0.97}\) & \(0.96\)\\
CENet-S & \(\textbf{0.96}\) & \(4.21\) & \(\textbf{1.19}\) & \(0.96\) & \(\textbf{0.97}\)\\
CENet-R & \(\textbf{0.96}\) & \(5.25\) & \(1.27\) & \(0.96\) & \(0.96\)\\
\end{tabularx}
\end{table*}

\begin{table}[t]
\renewcommand{\arraystretch}{1.7}
\captionsetup{justification=centering, labelsep=newline}
\caption{MEAN AND STANDARD DEVIATION OF THE EIGHT-FOLD CROSS-VALIDATION}
\label{table:results_stat}
\begin{tabularx}{\linewidth}{c||X|X|X|X}
Metric & DSN \cite{dou20173d} & VoxResNet \cite{chen2017voxresnet} & DenseVNet \cite{gibson2018automatic} & CENet\\
\hline
DSC & \(0.94\pm0.02\) & \(0.93\pm0.02\) & \(0.92\pm0.02\) & \(\textbf{0.96}\pmb{\pm}\textbf{0.01}\)\\
HD & \(7.49\pm4.21\) & \(7.99\pm4.04\) & \(11.47\pm6.60\) & \(\textbf{5.25}\pmb{\pm}\textbf{2.70}\)\\
ASSD & \(1.77\pm0.51\) & \(1.93\pm0.48\) & \(2.53\pm0.59\) & \(\textbf{1.17}\pmb{\pm}\textbf{0.30}\)\\
S & \(0.93\pm0.04\) & \(0.91\pm0.04\) & \(\textbf{0.97}\pmb{\pm}\textbf{0.02}\) & \(0.96\pm 0.03\)\\
P & \(0.94\pm0.02\) & \(0.95\pm0.02\) & \(0.88\pm0.03\) & \(\textbf{0.96}\pmb{\pm}\textbf{0.02}\)\\
\end{tabularx}
\end{table}

\begin{table}[t]
\renewcommand{\arraystretch}{1.7}
\captionsetup{justification=centering, labelsep=newline}
\caption{PAIRED SAMPLE T-TEST RESULTS (P-VALUES) OF EIGHT-FOLD CROSS-VALIDATION WITH CENET}
\label{table:results_stat2}
\begin{tabularx}{\linewidth}{c||>{\centering\arraybackslash}X|>{\centering\arraybackslash}X|>{\centering\arraybackslash}X}
Metric & DSN \cite{dou20173d} & VoxResNet \cite{chen2017voxresnet} & DenseVNet \cite{gibson2018automatic}\\
\hline
DSC & 5.63e-06 & 5.15e-10 & 4.29e-10 \\
HD & 1.28e-02 & 3.81e-03 & 9.50e-05 \\
ASSD & 6.00e-06 & 7.36e-08 & 2.06e-09 \\
S & 1.77e-03 & 1.94e-09 & \textbf{4.05e-01} \\
P & 6.00e-06 & 3.55e-02 & 1.72e-15 \\
\end{tabularx}
\end{table}

\begin{figure}
    \centering
        \vfil
        \subfloat[DSC.]{\includegraphics[width=\linewidth]{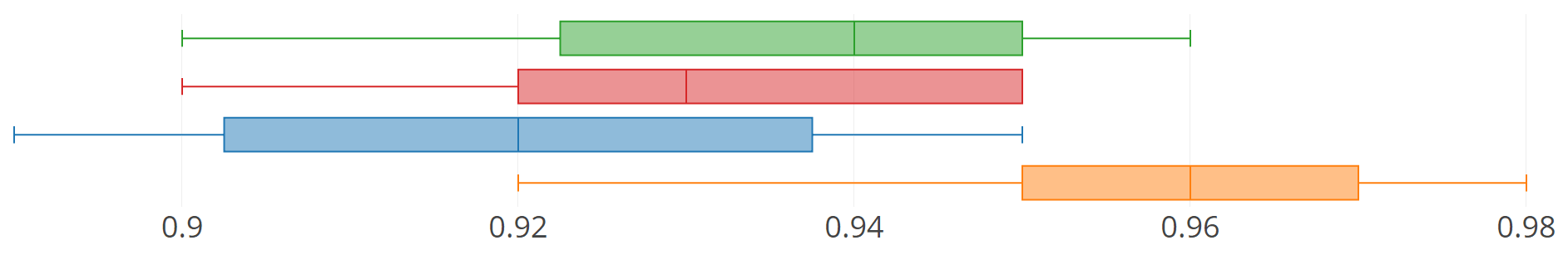}
        \label{fig:plot_dice}}
        \vfil
        \subfloat[95\% HD in mm.]{\includegraphics[width=\linewidth]{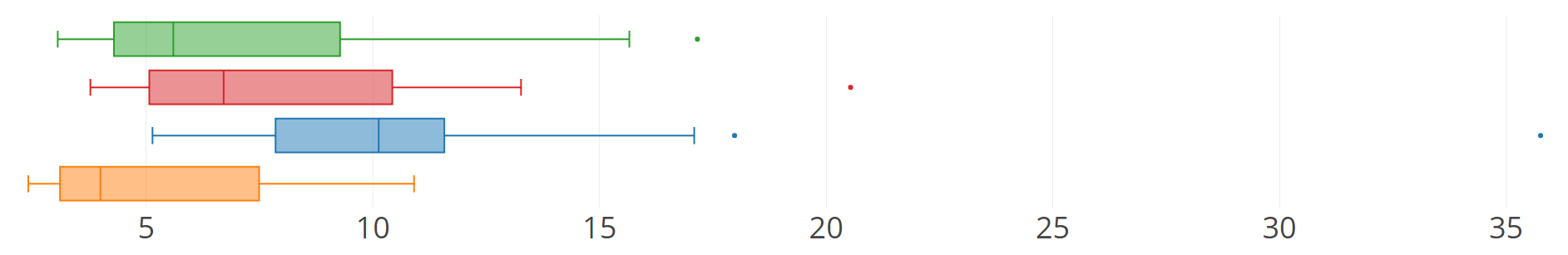}
        \label{fig:plot_hd95}}
        \vfil
        \subfloat[ASSD in mm.]{\includegraphics[width=\linewidth]{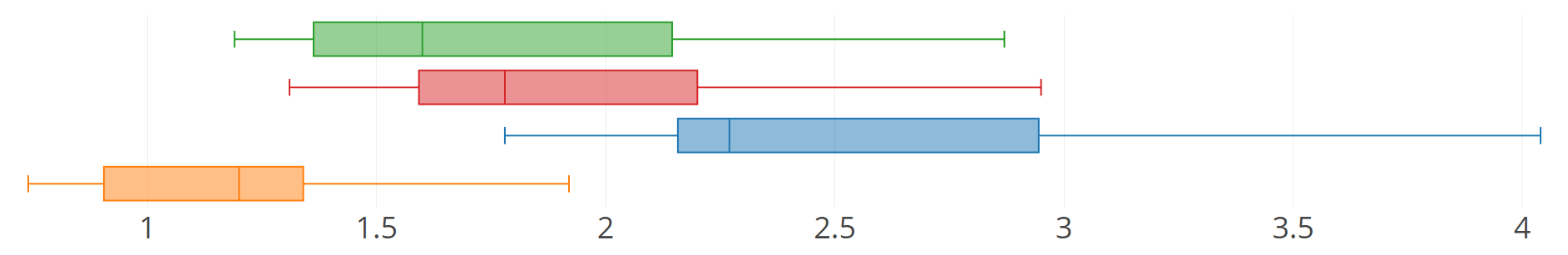}
        \label{fig:plot_mbd}}
        \vfil
        \subfloat[Sensitivity.]{\includegraphics[width=\linewidth]{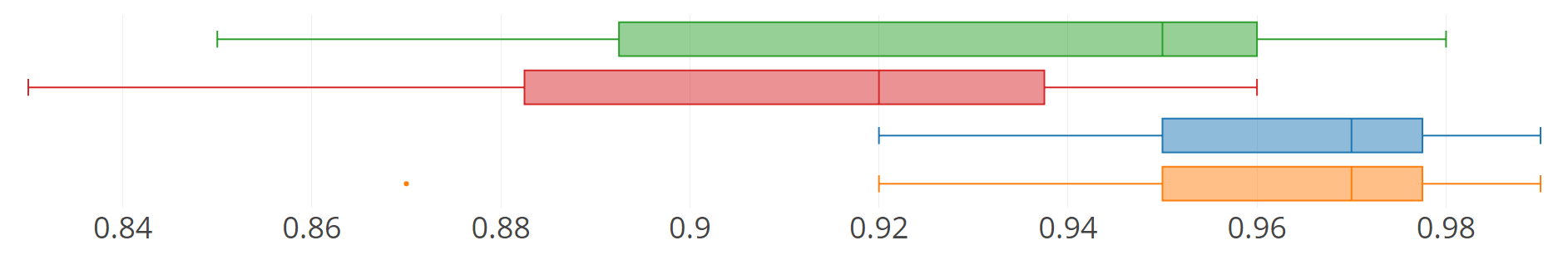}
        \label{fig:plot_s}}
        \vfil
        \subfloat[Precision.]{\includegraphics[width=\linewidth]{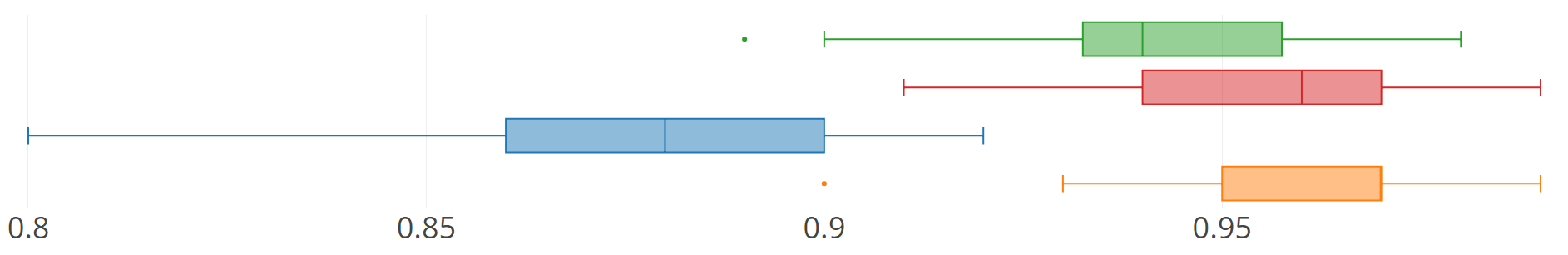}
        \label{fig:plot_p}}
        \vfil
        \subfloat{\includegraphics[width=2.5in]{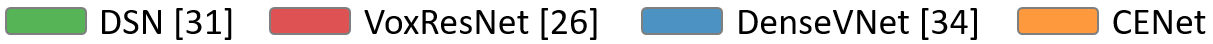}}
    \caption{Box plots of the segmentation metrics for the eight-fold performance evaluations.}
    \label{fig:plot}
\end{figure}

\subsection{Quantitative Evaluations}

\begin{figure}[tb]
    \centering
        \subfloat[Ground-truth.]{
        \includegraphics[width=1.12in]{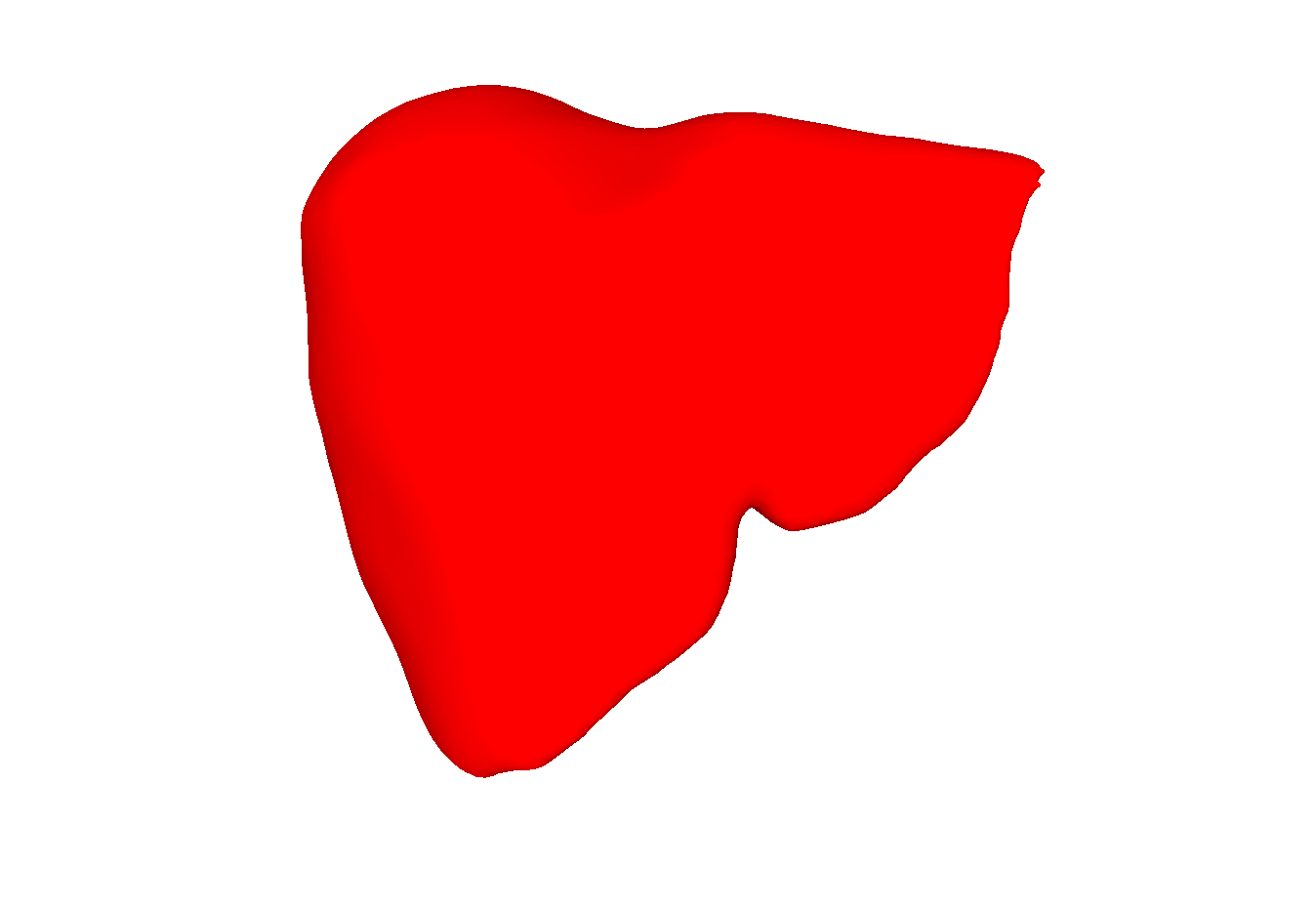}
        \includegraphics[width=1.12in]{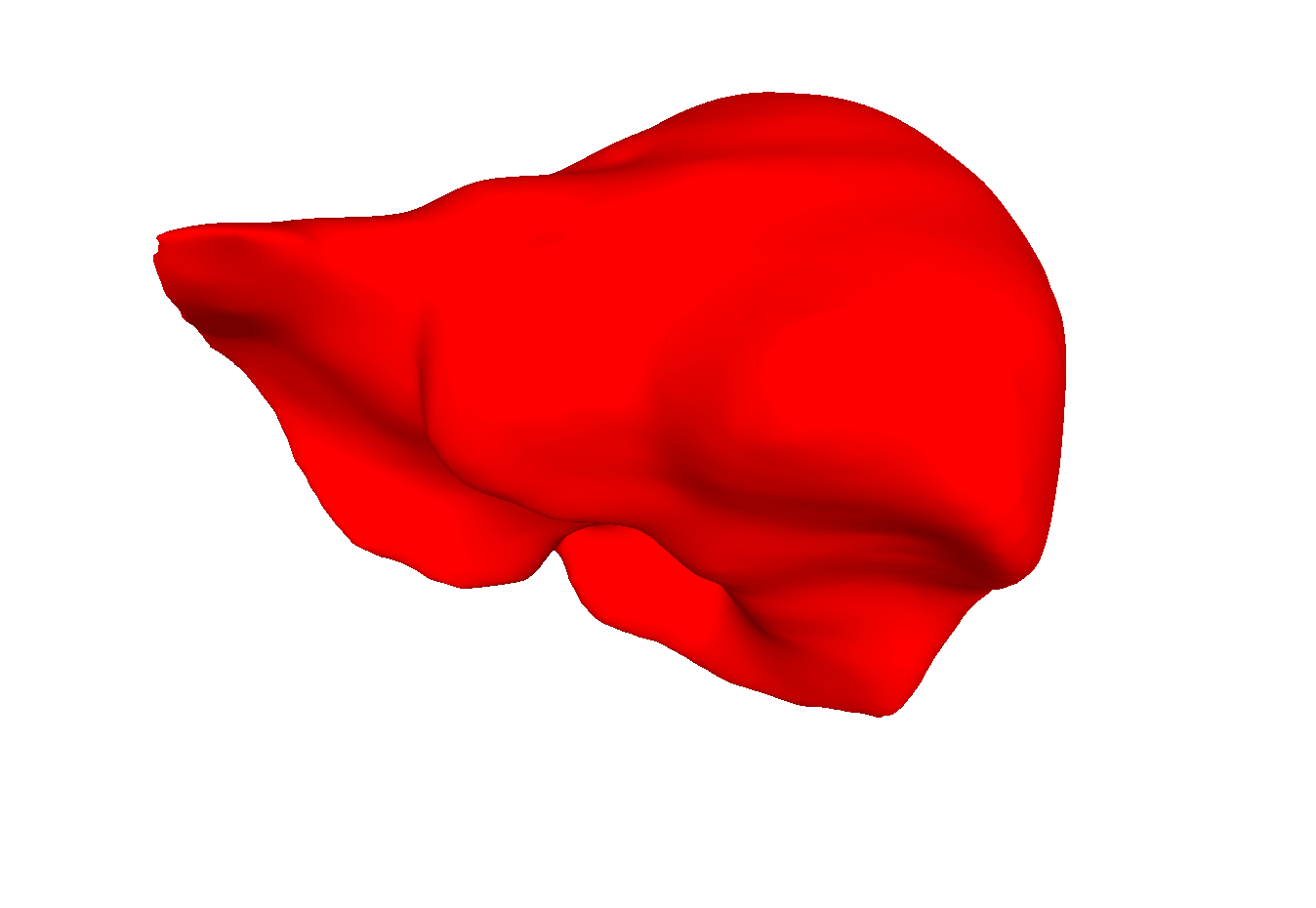}
        \includegraphics[width=1.12in]{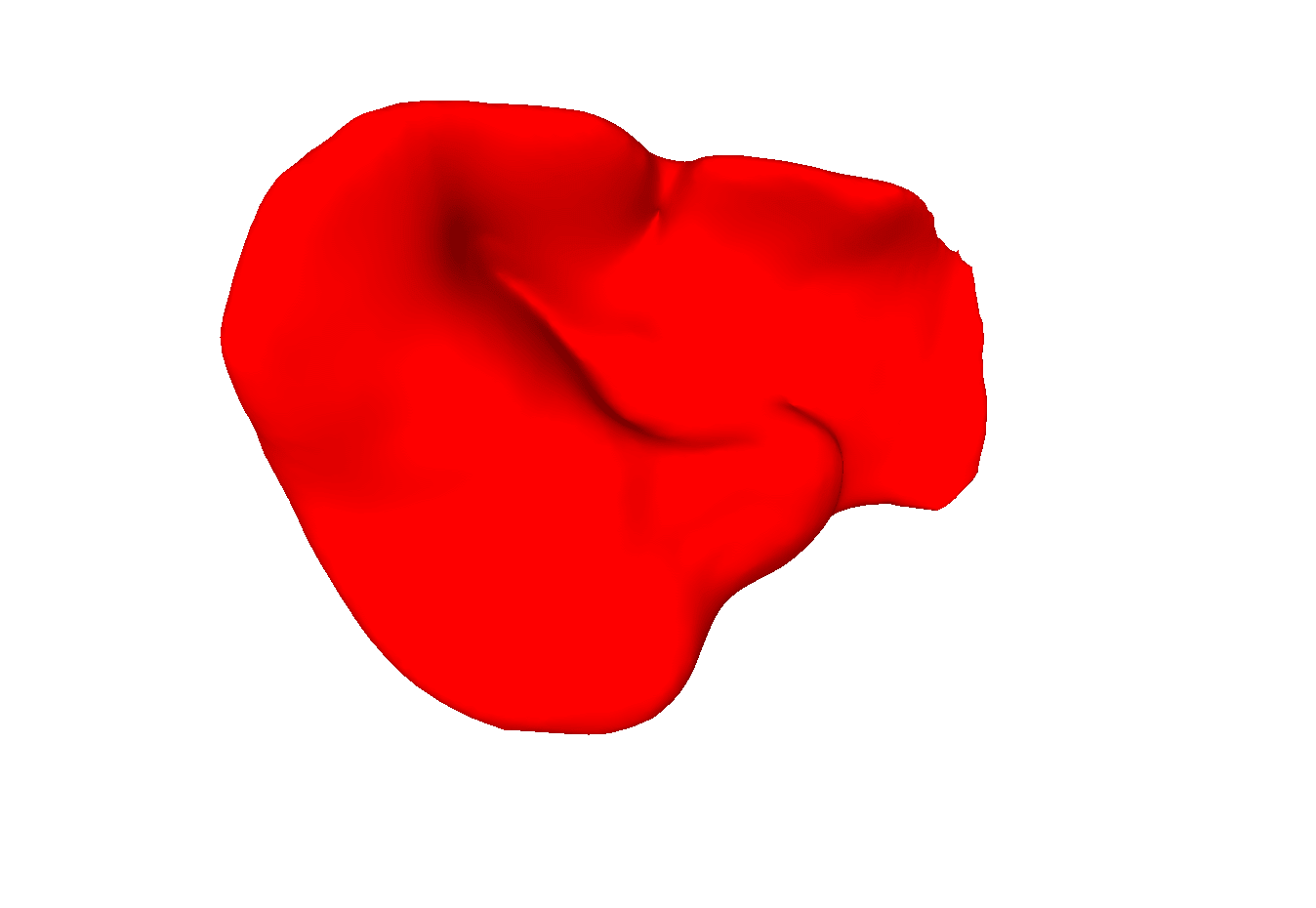}
        \label{fig:liver_gt}}
        \vfil
        
        \subfloat[DSN \cite{dou20173d}.]{
        \includegraphics[width=1.12in]{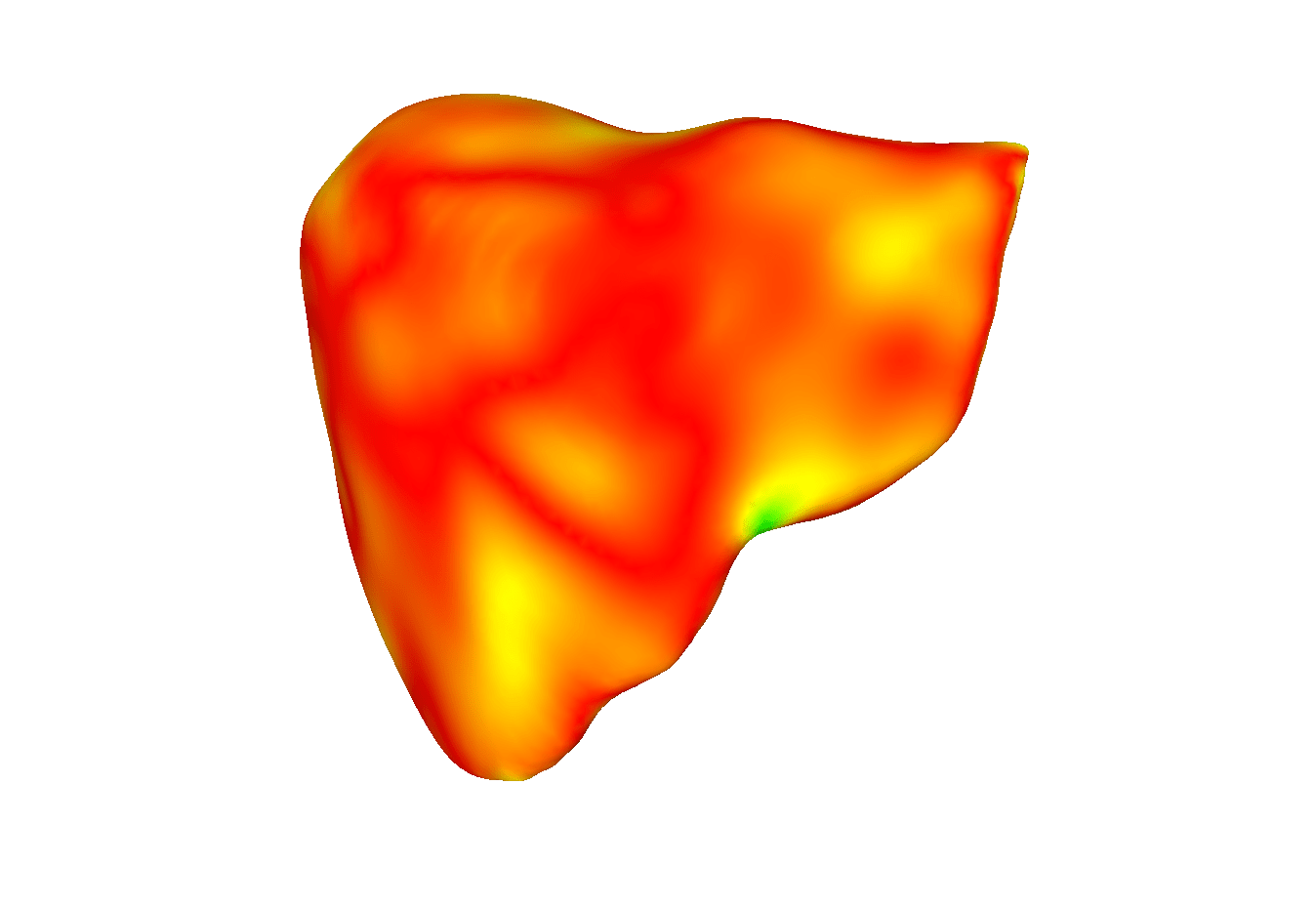}
        \includegraphics[width=1.12in]{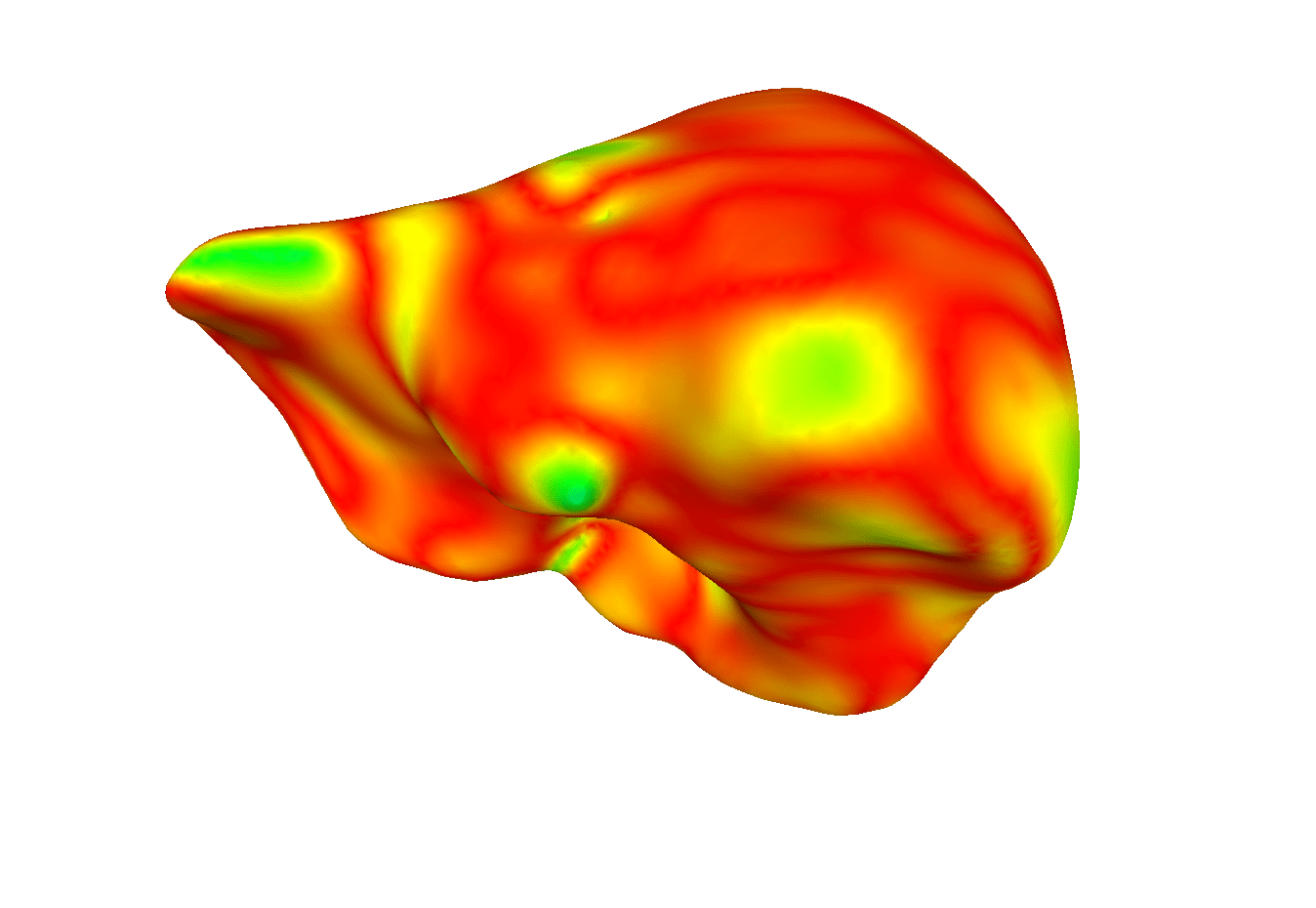}
        \includegraphics[width=1.12in]{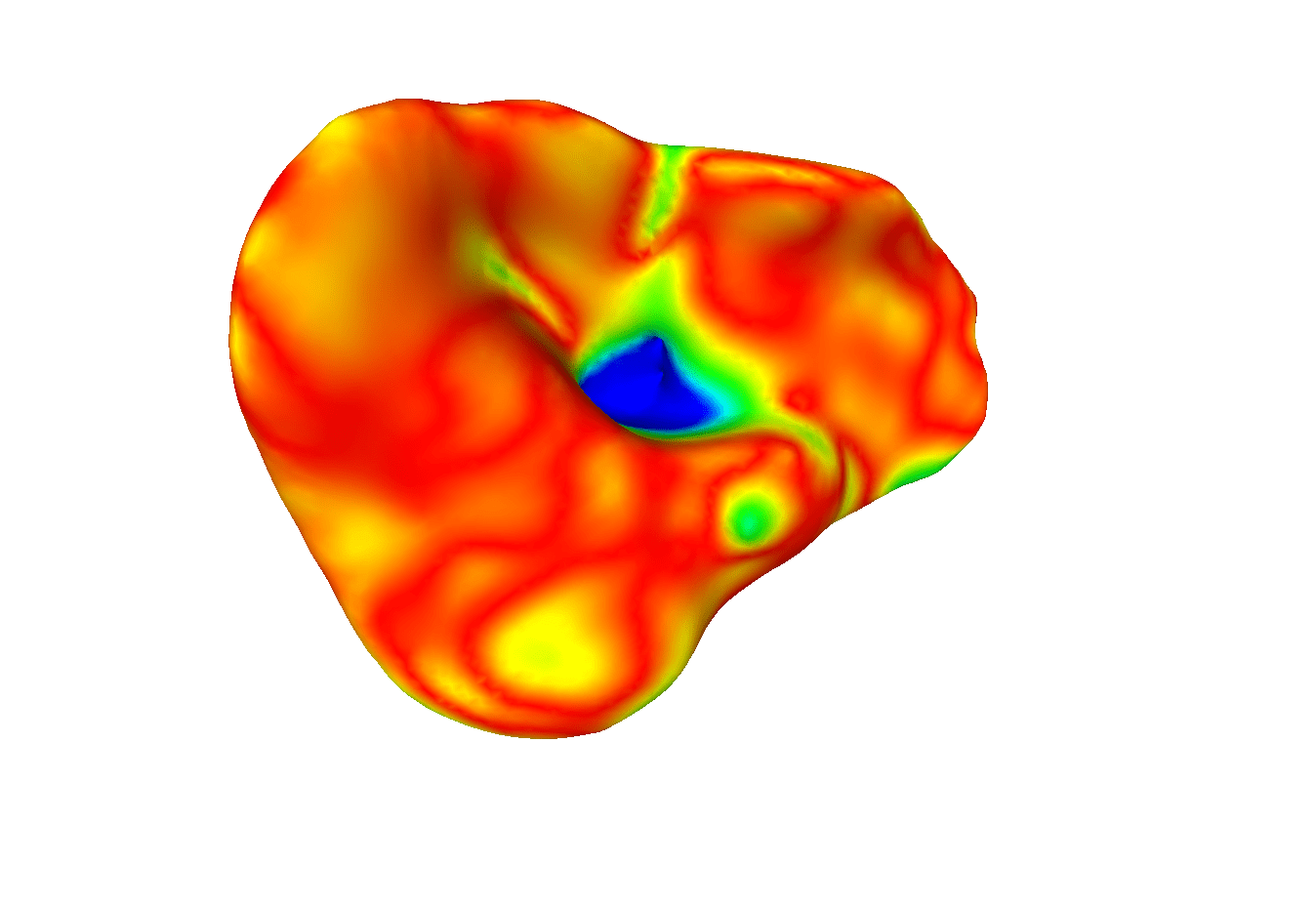}
        \label{fig:liver_dsn}}
        \vfil
        
        \subfloat[VoxResNet \cite{chen2017voxresnet}.]{
        \includegraphics[width=1.12in]{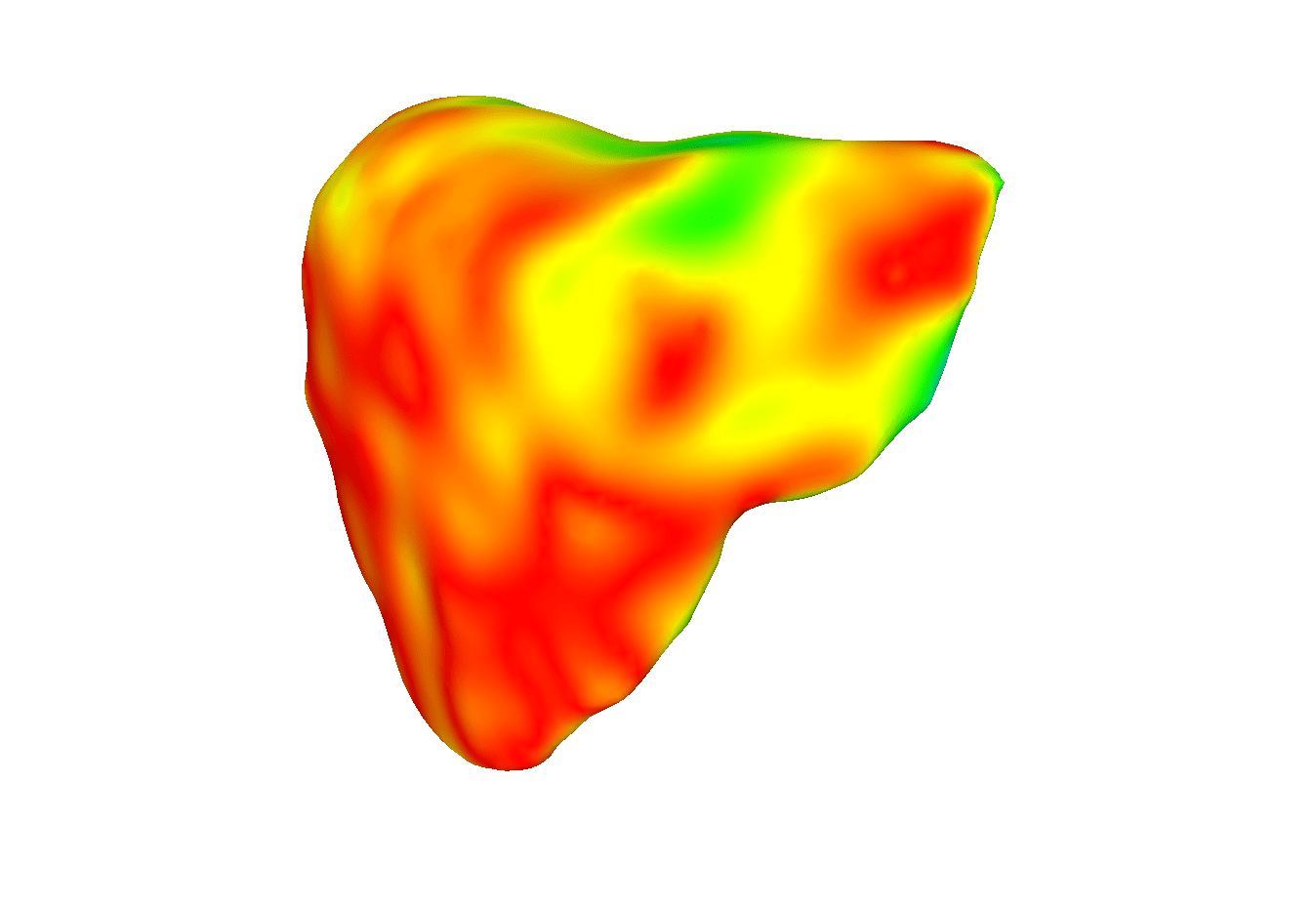}
        \includegraphics[width=1.12in]{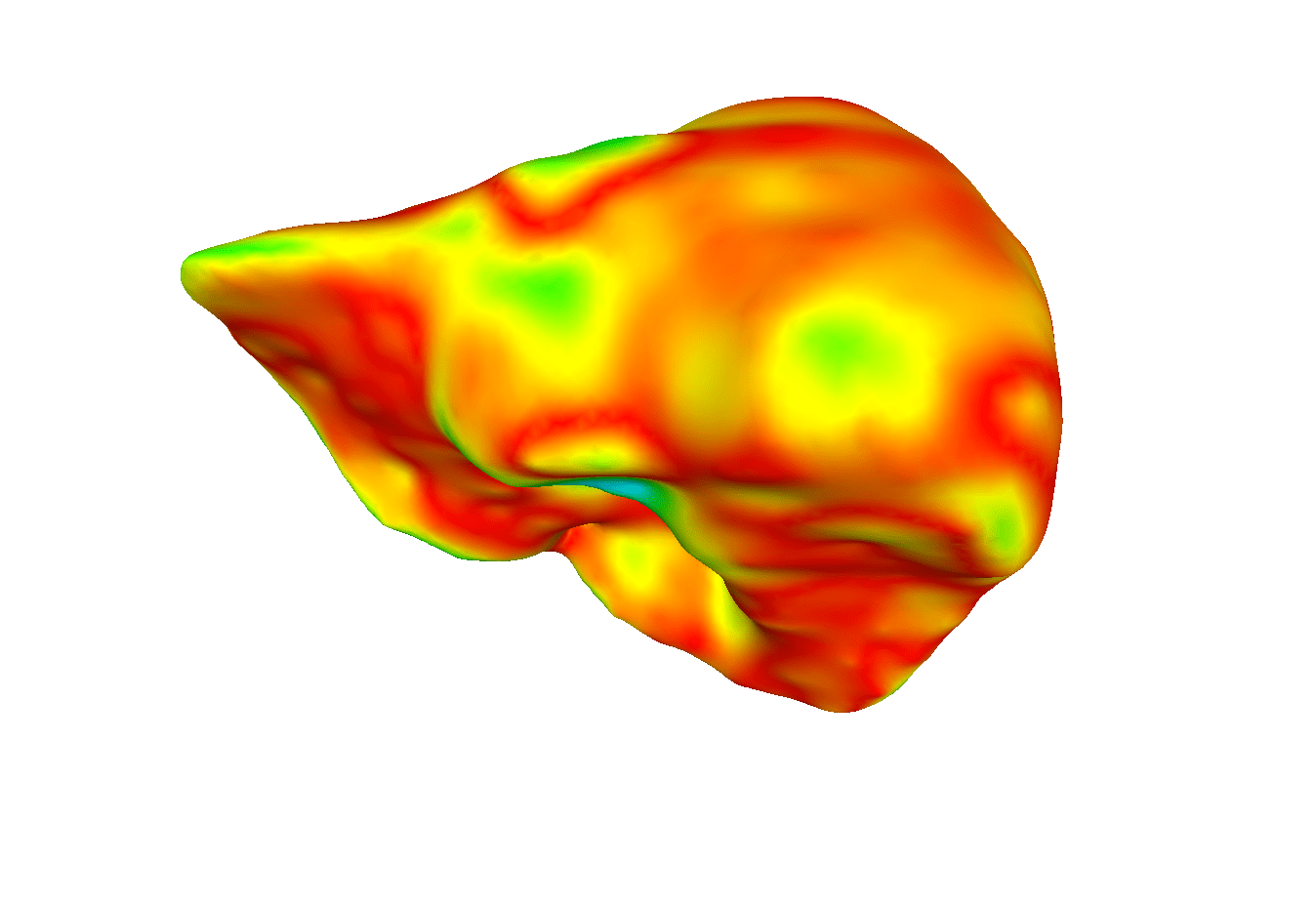}
        \includegraphics[width=1.12in]{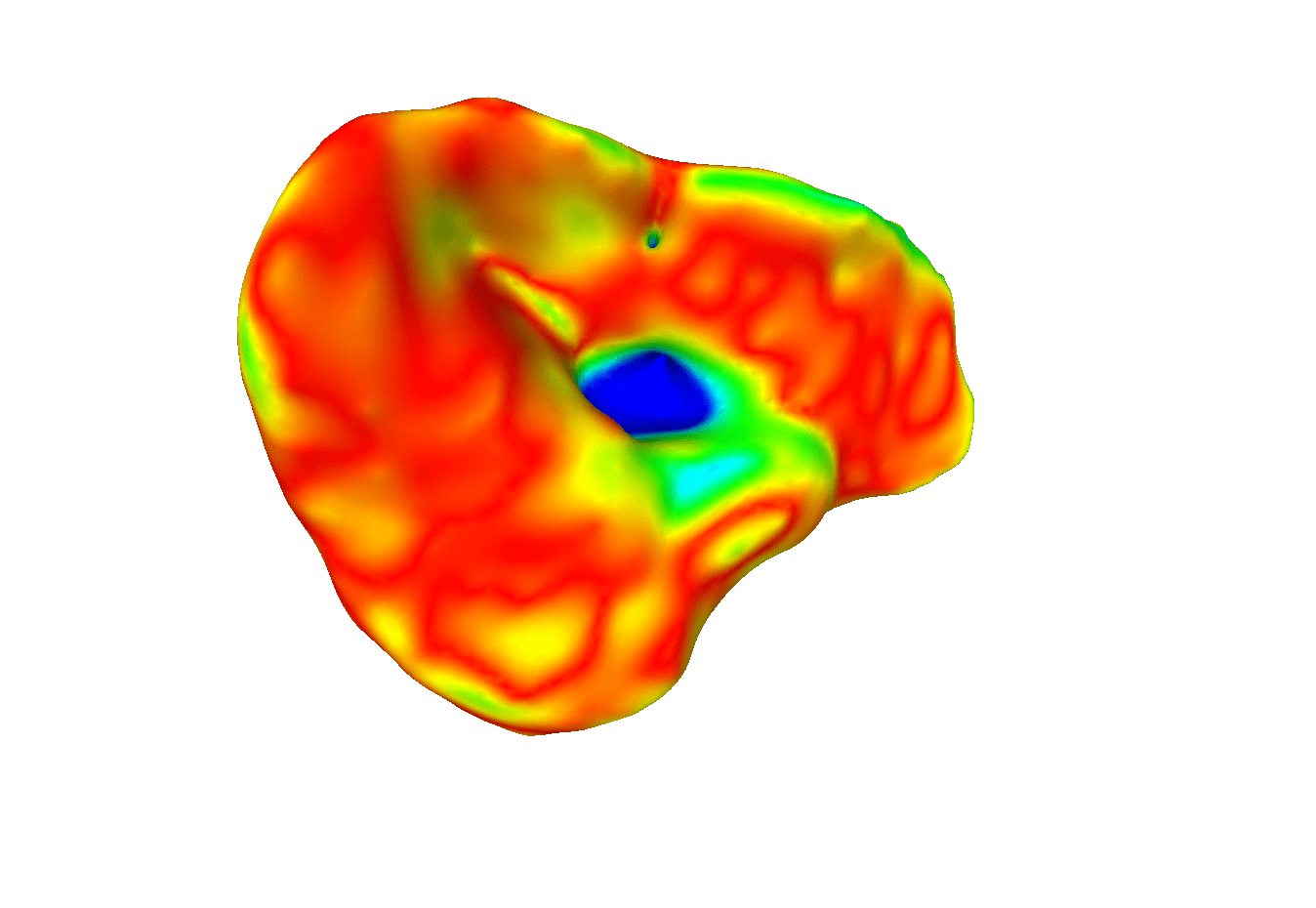}
        \label{fig:liver_voxresnet}}
        \vfil
        
        \subfloat[DenseVNet \cite{gibson2018automatic}.]{
        \includegraphics[width=1.12in]{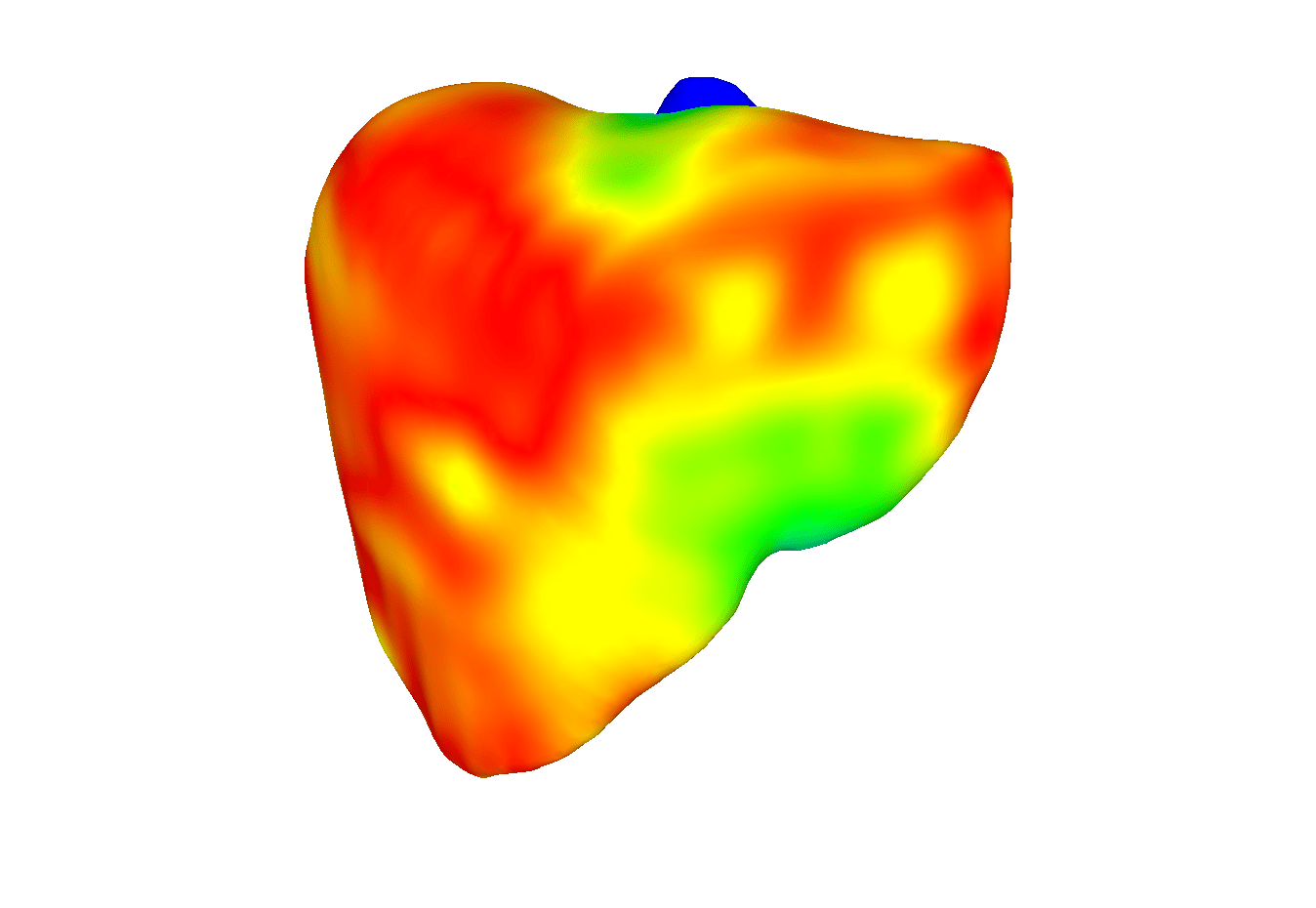}
        \includegraphics[width=1.12in]{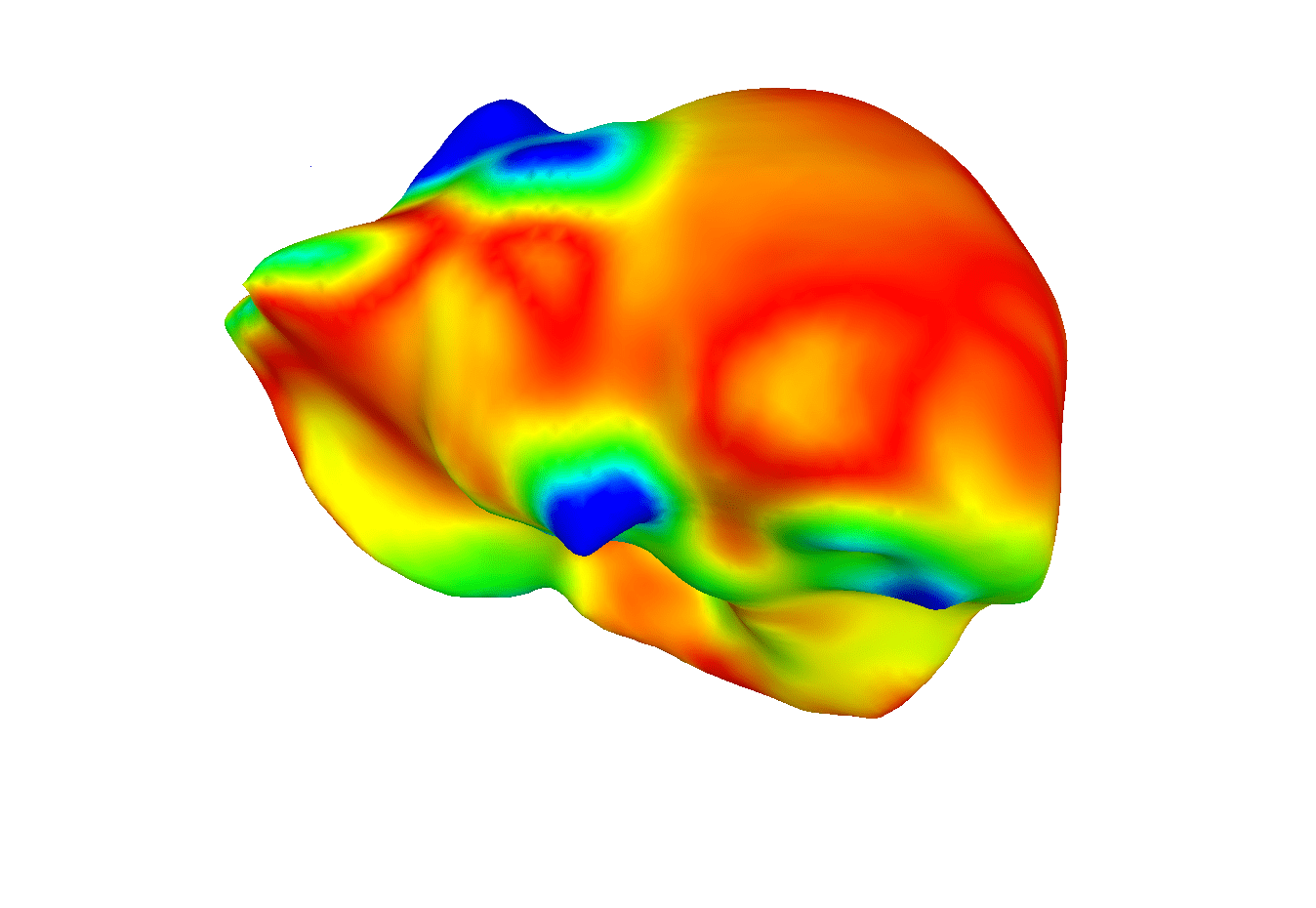}
        \includegraphics[width=1.12in]{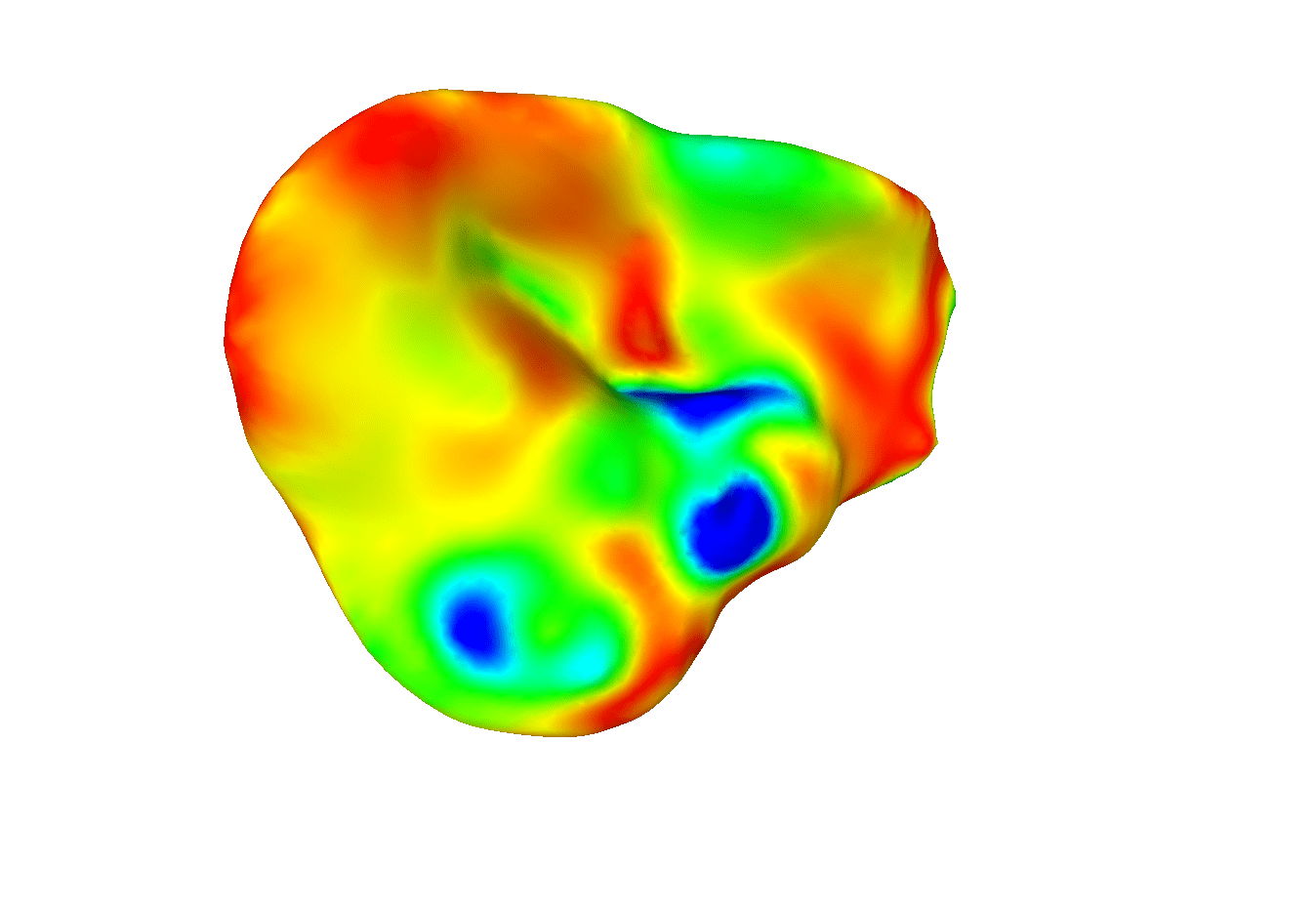}
        \label{fig:liver_densevnet}}
        \vfil
        
        \subfloat[CENet.]{
        \includegraphics[width=1.12in]{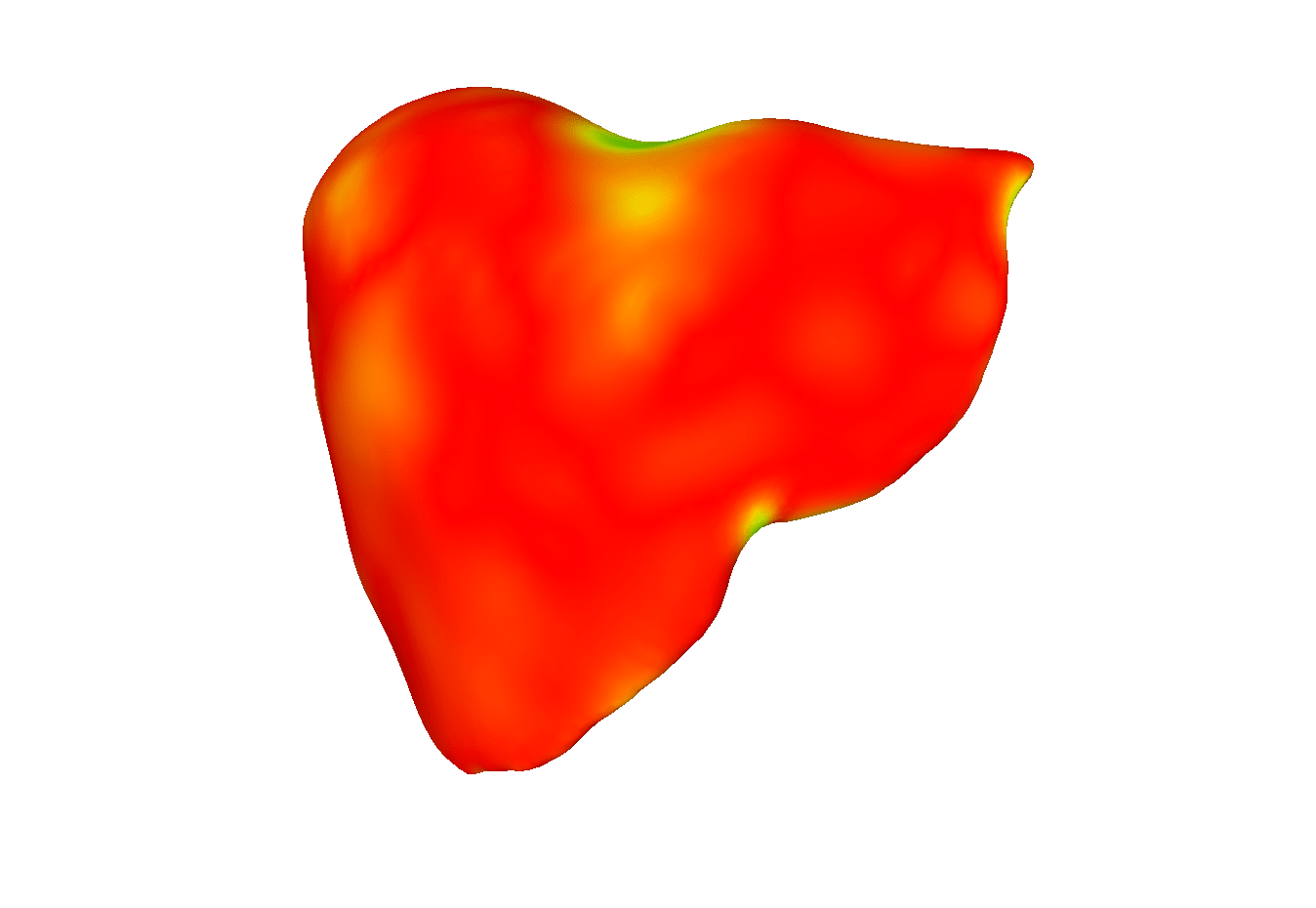}
        \includegraphics[width=1.12in]{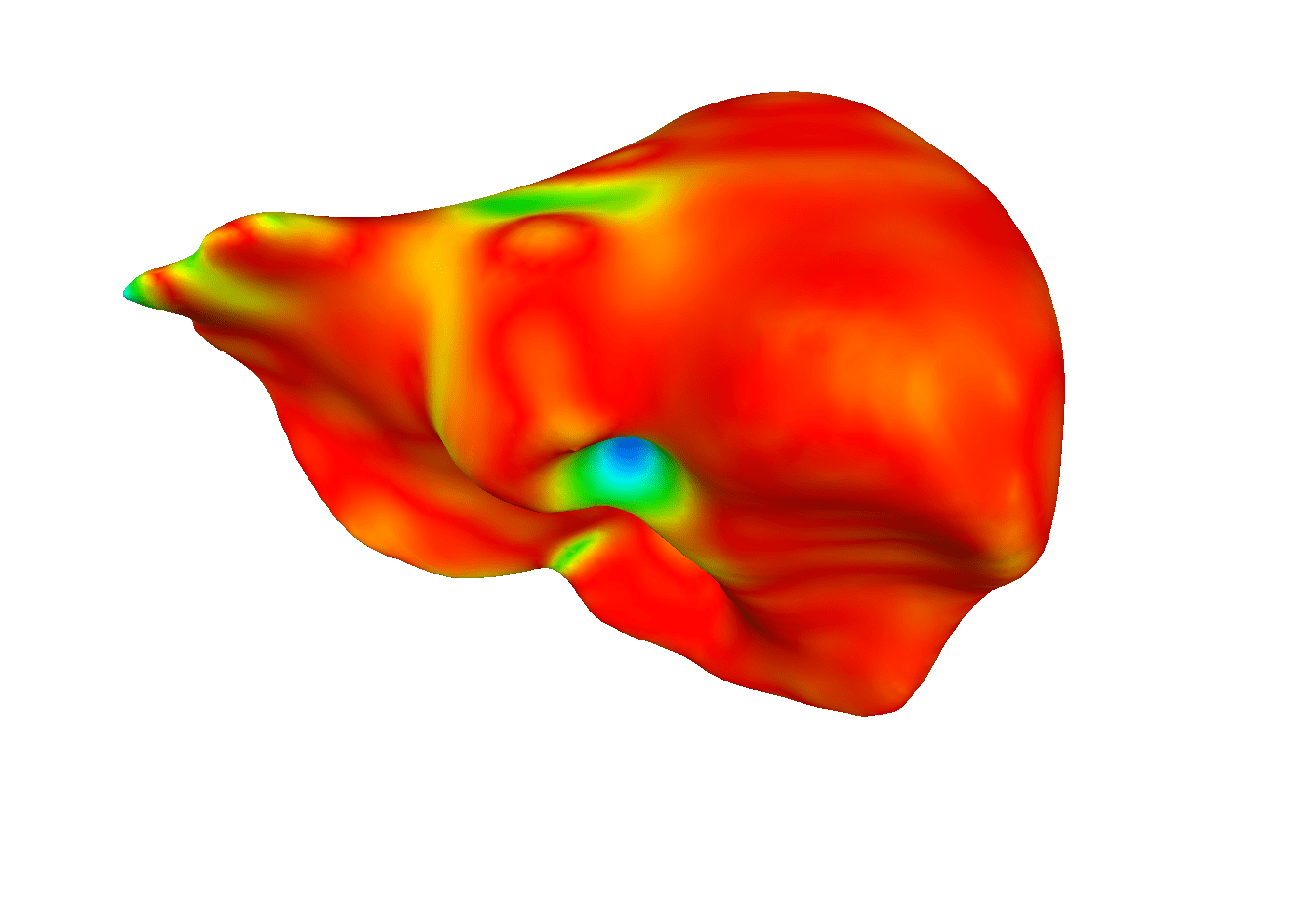}
        \includegraphics[width=1.12in]{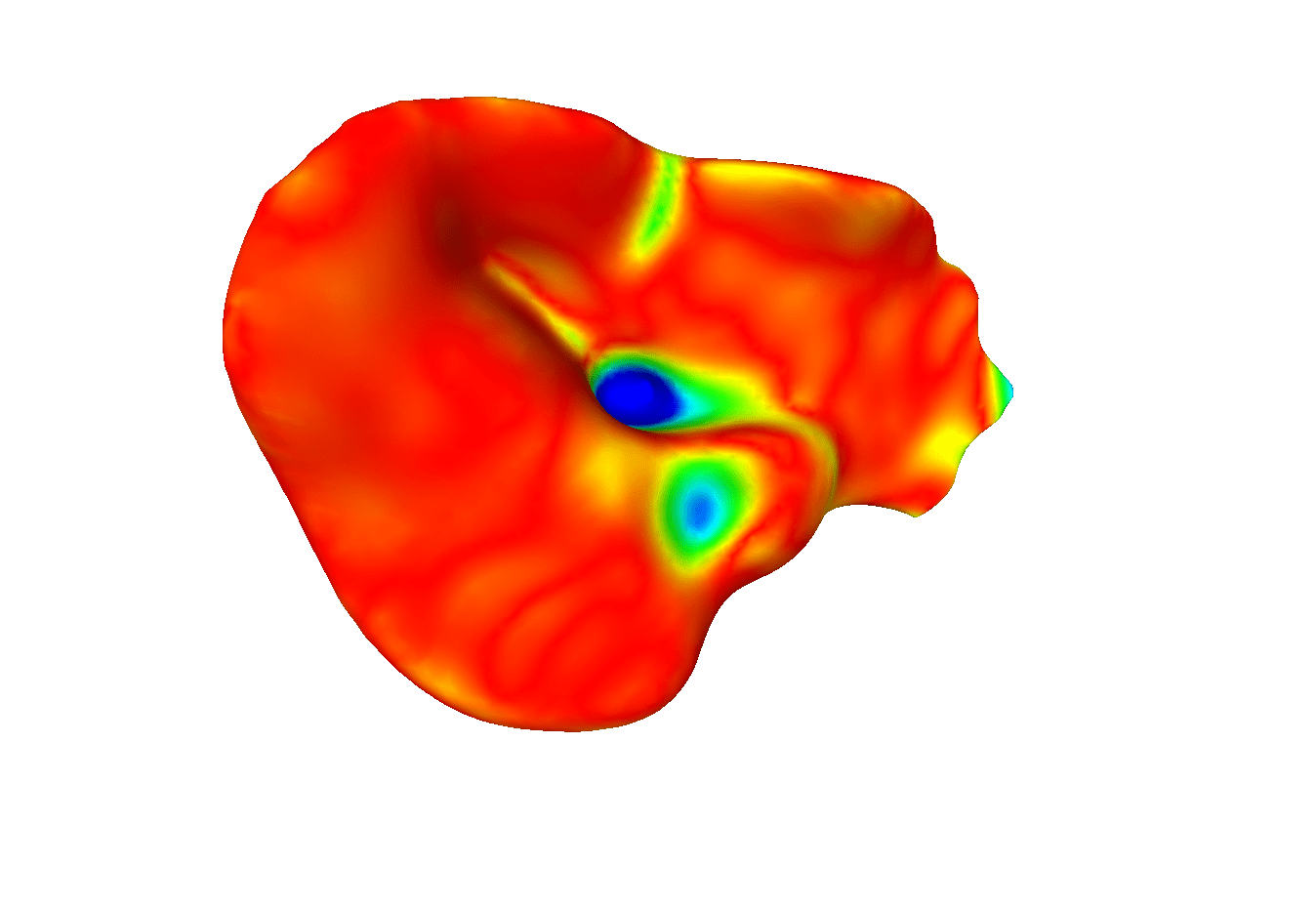}
        \label{fig:liver_CE}}
        \vfil
        
        \subfloat{
        \includegraphics[width=\linewidth]{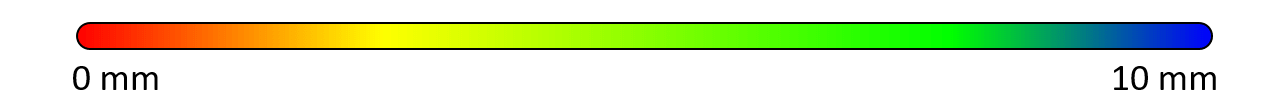}
        }
        
    \caption{Visualizations of the eight-fold segmentation results. The surface color is visualized with respect to the distance to the ground-truth surface. The visualized surfaces are smoothed via a curvature flow smoothing method \cite{smoothTriangMesh} at the original image resolution.}
    \label{fig:visualization}
\end{figure}

The segmentation results were evaluated using the dice similarity coefficient (DSC), 95\% Hausdorff distance (HD), average symmetric surface distance (ASSD), sensitivity (S), and precision (P). The DSC is defined as follows:
\begin{equation}
    DSC(X,Y)=\frac{2|X \cap Y|}{|X|+|Y|},
\end{equation}
\noindent
where \(|\cdot|\) is the cardinality of a set. \(\textbf{S}_X\) is defined as a set of surface voxels of a set \(X\), the shortest distance of an arbitrary voxel \(p\) is defined as follows \cite{heimann2009comparison}:
\begin{equation}
    d(p, \textbf{S}_X)=\min_{s_X \in \textbf{S}_X}{||p-s_X||}_2.
\label{eq:d}
\end{equation}
\noindent
Thus, HD is defined as follows \cite{heimann2009comparison}:
\begin{equation}
    HD(X,Y)=\max\{\max_{s_X \in \textbf{S}_X}{d(s_X, \textbf{S}_Y)}+\max_{s_Y \in \textbf{S}_Y}{d(s_Y, \textbf{S}_X)}\}.
\label{eq:hd}
\end{equation}
\noindent
Defining the distance function as
\begin{equation}
    D(\textbf{S}_X, \textbf{S}_Y)=\Sigma_{s_X \in \textbf{S}_X} d(s_X, \textbf{S}_Y),
\end{equation}
the ASSD can be defined as follows \cite{heimann2009comparison}:
\begin{equation}
    ASSD(X,Y)=\frac{1}{|\textbf{S}_X|+|\textbf{S}_Y|}(D(\textbf{S}_X, \textbf{S}_Y)+D(\textbf{S}_Y, \textbf{S}_X)).
\end{equation}
The sensitivity and precision are defined as follows:
\begin{equation}
    S=\frac{TP}{TP+FN},
\label{eq:sensitivity}
\end{equation}
\begin{equation}
    P=\frac{TP}{TP+FP}
\label{eq:precision}
\end{equation}
\noindent
where \(TP\), \(FN\), and \(FP\) are the numbers of true positive, false negative, and false positive voxels, respectively. In (\ref{eq:hd}), 95\% of the voxels in (\ref{eq:d}) were calculated to exclude 5\% of the outlying voxels. This allows to obtain a generalized evaluation of the distance without portal vein variations (Fig. \ref{fig:visualization}).\par

We used DSN \cite{dou20173d}, VoxResNet \cite{chen2017voxresnet}, DenseVNet \cite{gibson2018automatic}, and the proposed network, CENet, for the performance evaluation. In addition to comparing the state-of-the-arts, we extended the experiments with our network variants: the CENet without self-supervised contour learning (i.e., using the full ground-truth contour $\Gamma_c$ instead of the adaptively modified $\Tilde{\Gamma_c}$; CENet-A), without contour transition layer (i.e., removing the red box in Fig. \ref{fig:network}; CENet-C), without shape transition layer (i.e., removing the blue box in Fig. \ref{fig:network}; CENet-S), and without the residual shape estimation layer (i.e., removing the black box in Fig. \ref{fig:network}; CENet-R). In the case of the CENet-R, two shape transition layers were sequentially stacked for the shape estimation.\par

An eight-fold cross-validation is used to obtain the quantitative results in Tables \ref{table:results} and \ref{table:results_stat}. The visual box plot of Table \ref{table:results_stat} is presented in Fig. \ref{fig:plot}. The proposed CENet showed the best segmentation results within all evaluations. In particular, the DenseVNet failed to segment the liver accurately owing to two significant issues: 1) the network resolution is too low and 2) the shape prior has a weak representative power. Thus, for images with excessively coarse dimensions, the segmentation result suffers from the accurate delineation of an object in the original domain. Furthermore, the \(12^3\) resolution of the shape prior is too small and the training images must be accurately and manually cropped to fully utilize the learned shape prior. There is no specific metric presented in the previous research reported in reference \cite{gibson2018automatic} to crop testing images automatically. The Table \ref{table:results_stat2} shows the result of paired sample t-tests (i.e., p-values) with the proposed CENet by 95\% confidence interval. The values demonstrate that the statistics are significantly different from the CENet except for the sensitivity of DenseVNet \cite{gibson2018automatic}. DenseVNet showed comparable sensitivity which indicates the final predictions had no significant false negative responses (\ref{eq:sensitivity}).\par

\begin{figure}[t]
    \centering
    \includegraphics[width=\linewidth]{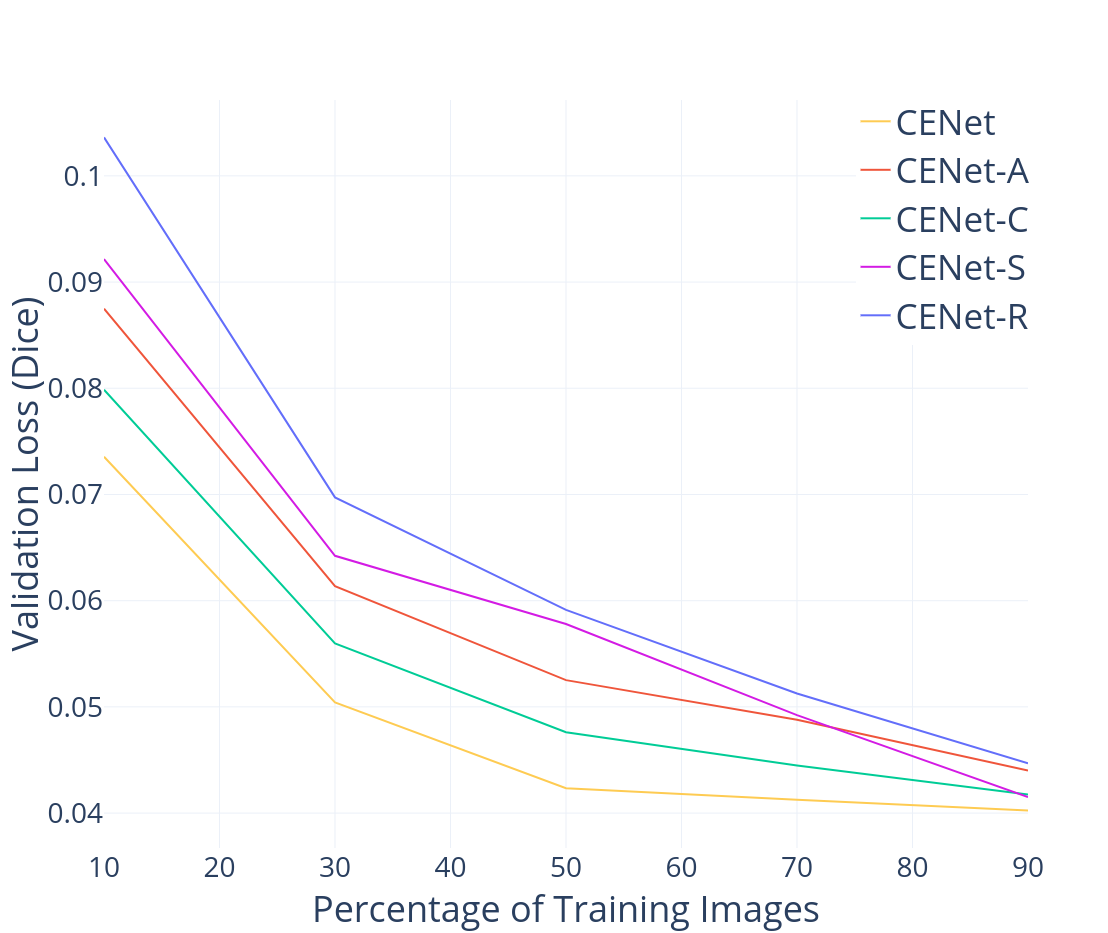}
    \caption{Multi-fold cross-validation results of the proposed variants. The dice loss for each validation set is plotted.}
    \label{fig:multi_fold}
\end{figure}

The accuracy of our network variants was slightly lower than that of the original CENet. The DSC, sensitivity, and precision scores of the variants were preserved while the distance errors (i.e., 95\% HD and ASSD) slightly increased. The CENet-S showed the lowest distance errors among the variants, while the CENet-R showed the highest distance errors. This indicates that the residual shape estimation process is critical for an accurate shape estimation. When using the CENet-R network, the \(F^0_s\) feature was similar to that shown in Fig. \ref{fig:residual1} which leads to an inaccurate output result. Without residuals, the design of more complex and deep transition layers is required for the shape estimation, which may lead to an over-fitting. The result of the CENet-C indicates that the contour transition part plays a key role in the accurate delineation of an object. However, the performance of the CENet-A was poorer than that of the CENet-C with respect to the HD and ASSD measurements, indicating that enforcing the network to learn the full ground-truth contour image has a negative effect on the performance. The overall performance of generalization is presented in Fig. \ref{fig:multi_fold} by employing multi-fold cross-validations. We have cross-validated the network variants by using 10\%, 30\%, 50\%, 70\%, and 90\% of images for training. The dice loss for each validation set is plotted. As shown in Fig. \ref{fig:multi_fold}, the ablations showed poorer performance compared to the original network. The CENet-C, which does not employ contour transition, showed a slight degradation in the dice score. The performance of CENet-A, which uses full contour, showed a lower dice score compared to CENet-C. Above all, CENet-S and CENet-R showed the worst dice scores in the fewer training images indicating that the deep supervision of shape with residual connection primarily aids the performance of generalization.\par

The visual result of an example liver subject is presented in Fig. \ref{fig:visualization}. As it is clearly visualized, the proposed CENet successfully segmented the liver with accurate guidelines of the contour and shape estimation. Segmenting the portal vein entry region accurately was difficult to achieve with all the networks, including the proposed one. However, our training database (i.e., clinically annotated ground-truth images) presented serious internal variations in the portal vein entry region. Several clinicians included the vessels but others excluded the major entry vessel region. A concurrent and integrated liver and vascular system segmentation framework could be built in the future to overcome the variability of annotations. In the case of the DenseVNet, the inaccurate shape prior seriously affected the final output, as shown in Fig. \ref{fig:liver_densevnet}.

\section{DISCUSSION}
The segmentation of organs in medical imaging is a challenging issue. The edge is unquestionably the most important feature for accurate object segmentation in the perspective of contour delineation. However, the full contour is hard to identify in various cases, such as unclear boundaries and false edges in contrast-enhanced vessels. Even with the strong capability of the neural network, it is difficult to classify ambiguous regions. Thus, the proposed network avoids learning the full contour features that are unnecessary in this study. The proposed method guided (i.e., self-supervised) the neural network to learn the sparse but essential contour that can be a great complementary feature to be later fused with the global shape estimation. Two major neural network branches were used: contour and shape estimations. This network may be seen as a multi-task learning framework. However, the network was not enforced to explicitly inference multiple tasks. The proposed network internally guides weights to represent the object contour features without supervising the entire contour image. The network was self-supervised with modified contour images for each iteration. The main underlying principle of the proposed network is to concentrate the contour delineation pass on the missing contour part of an object (i.e., fine details of an object that are easily misclassified using the end-to-end learning). There are two main reasons for using the proposed method: 1) even with a powerful deep neural network, unclear boundaries are challenging to be discriminated as a contour and 2) contour regions in unclear boundaries can be delineated by global shapes. Finally, we merged three strong discriminative features (i.e., shape, contour details, and deep features) to obtain accurate segmentation results. The proposed network can be intuitively interpreted as a robust contour guided shape estimation.\par

For the effective modification of the proposed network to other applications, the parameters of the dense block (Fig. \ref{fig:denseblock}) and \(p\) (i.e., the threshold value to determine the misclassified voxels) should be modified. The parameters \(k\) and \(n\) in the dense block adjust the complexity of the network and \(p\) adjusts the workload of the contour transition. The higher the value of \(p\), the larger the contour region required to be delineated in the contour estimation pass. Herein (i.e., liver segmentation), the parameter \(p\) was not sensitive to the presented results.\par


\section{CONCLUSION}
In this work, an FCN was designed for image segmentation with a self-supervised contour-guiding scheme. The proposed network combined the shape and contour features to accurately delineate the target object. The contour features were learned to delineate the complementary contour region in a self-supervising scheme. The network was divided into two big branches for shape and complementary contour estimations. The proposed network demonstrated that the critical and partial contour features, instead of the fully-supervised contour, could effectively improve the performance of the segmentation result. The quantitative experiments showed that our method performed 2.13\% more accurately than the state-of-the-art method with respect to the dice score. The deep contour self-supervision was automatically performed by the output of the network without any manual interactions. The building block of our network was a densely connected block with separable convolutions, which made the network more compact and representative. The proposed network successfully performed the liver segmentation without deepening or widening the neural network, unlike the state-of-the-art methods.



\ifCLASSOPTIONcaptionsoff
  \newpage
\fi

\bibliographystyle{IEEEtran}
\bibliography{myBiB}

\begin{thebibliography}{10}
\providecommand{\url}[1]{#1}
\csname url@samestyle\endcsname
\providecommand{\newblock}{\relax}
\providecommand{\bibinfo}[2]{#2}
\providecommand{\BIBentrySTDinterwordspacing}{\spaceskip=0pt\relax}
\providecommand{\BIBentryALTinterwordstretchfactor}{4}
\providecommand{\BIBentryALTinterwordspacing}{\spaceskip=\fontdimen2\font plus
\BIBentryALTinterwordstretchfactor\fontdimen3\font minus
  \fontdimen4\font\relax}
\providecommand{\BIBforeignlanguage}[2]{{%
\expandafter\ifx\csname l@#1\endcsname\relax
\typeout{** WARNING: IEEEtran.bst: No hyphenation pattern has been}%
\typeout{** loaded for the language `#1'. Using the pattern for}%
\typeout{** the default language instead.}%
\else
\language=\csname l@#1\endcsname
\fi
#2}}
\providecommand{\BIBdecl}{\relax}
\BIBdecl

\bibitem{lim2006automatic}
S.-J. Lim, Y.-Y. Jeong, and Y.-S. Ho, ``Automatic liver segmentation for volume
  measurement in ct images,'' \emph{Journal of Visual Communication and Image
  Representation}, vol.~17, no.~4, pp. 860--875, 2006.

\bibitem{rusko2007fully}
L.~Rusko, G.~Bekes, G.~Nemeth, and M.~Fidrich, ``Fully automatic liver
  segmentation for contrast-enhanced ct images,'' \emph{MICCAI Wshp. 3D
  Segmentation in the Clinic: A Grand Challenge}, vol.~2, no.~7, 2007.

\bibitem{suzuki2010computer}
K.~Suzuki, R.~Kohlbrenner, M.~L. Epstein, A.~M. Obajuluwa, J.~Xu, and M.~Hori,
  ``Computer-aided measurement of liver volumes in ct by means of geodesic
  active contour segmentation coupled with level-set algorithms,''
  \emph{Medical physics}, vol.~37, no.~5, pp. 2159--2166, 2010.

\bibitem{lee2007efficient}
J.~Lee, N.~Kim, H.~Lee, J.~B. Seo, H.~J. Won, Y.~M. Shin, Y.~G. Shin, and S.-H.
  Kim, ``Efficient liver segmentation using a level-set method with optimal
  detection of the initial liver boundary from level-set speed images,''
  \emph{Computer Methods and Programs iBiomedicine}, vol.~88, no.~1, pp.
  26--38, 2007.

\bibitem{zhang2010automatic}
X.~Zhang, J.~Tian, K.~Deng, Y.~Wu, and X.~Li, ``Automatic liver segmentation
  using a statistical shape model with optimal surface detection,'' \emph{IEEE
  Transactions on Biomedical Engineering}, vol.~57, no.~10, pp. 2622--2626,
  2010.

\bibitem{okada2007automated}
T.~Okada, R.~Shimada, Y.~Sato, M.~Hori, K.~Yokota, M.~Nakamoto, Y.-W. Chen,
  H.~Nakamura, and S.~Tamura, ``Automated segmentation of the liver from 3d ct
  images using probabilistic atlas and multi-level statistical shape model,''
  in \emph{International Conference on Medical Image Computing and
  Computer-Assisted Intervention}.\hskip 1em plus 0.5em minus 0.4em\relax
  Springer, 2007, pp. 86--93.

\bibitem{ling2008hierarchical}
H.~Ling, S.~K. Zhou, Y.~Zheng, B.~Georgescu, M.~Suehling, and D.~Comaniciu,
  ``Hierarchical, learning-based automatic liver segmentation,'' in
  \emph{Computer Vision and Pattern Recognition, 2008. CVPR 2008. IEEE
  Conference on}.\hskip 1em plus 0.5em minus 0.4em\relax IEEE, 2008, pp. 1--8.

\bibitem{heimann2009comparison}
T.~Heimann, B.~Van~Ginneken, M.~A. Styner, Y.~Arzhaeva, V.~Aurich, C.~Bauer,
  A.~Beck, C.~Becker, R.~Beichel, G.~Bekes \emph{et~al.}, ``Comparison and
  evaluation of methods for liver segmentation from ct datasets,'' \emph{IEEE
  transactions on medical imaging}, vol.~28, no.~8, pp. 1251--1265, 2009.

\bibitem{campadelli2009liver}
P.~Campadelli, E.~Casiraghi, and A.~Esposito, ``Liver segmentation from
  computed tomography scans: A survey and a new algorithm,'' \emph{Artificial
  intelligence in medicine}, vol.~45, no. 2-3, pp. 185--196, 2009.

\bibitem{van2007automatic}
E.~van Rikxoort, Y.~Arzhaeva, and B.~van Ginneken, ``Automatic segmentation of
  the liver in computed tomography scans with voxel classification and atlas
  matching,'' in \emph{Proceedings of the MICCAI Workshop}, vol.~3.\hskip 1em
  plus 0.5em minus 0.4em\relax Citeseer, 2007, pp. 101--108.

\bibitem{heimann2007statistical}
T.~Heimann, H.~Meinzer, and I.~Wolf, ``A statistical deformable model for the
  segmentation of liver ct volumes using extended training data,'' \emph{Proc.
  MICCAI Work}, pp. 161--166, 2007.

\bibitem{kainmuller2007shape}
D.~Kainm{\"u}ller, T.~Lange, and H.~Lamecker, ``Shape constrained automatic
  segmentation of the liver based on a heuristic intensity model,'' in
  \emph{Proc. MICCAI Workshop 3D Segmentation in the Clinic: A Grand
  Challenge}, 2007, pp. 109--116.

\bibitem{wimmer2009generic}
A.~Wimmer, G.~Soza, and J.~Hornegger, ``A generic probabilistic active shape
  model for organ segmentation,'' in \emph{International Conference on Medical
  Image Computing and Computer-Assisted Intervention}.\hskip 1em plus 0.5em
  minus 0.4em\relax Springer, 2009, pp. 26--33.

\bibitem{simonyan2014very}
K.~Simonyan and A.~Zisserman, ``Very deep convolutional networks for
  large-scale image recognition,'' \emph{arXiv preprint arXiv:1409.1556}, 2014.

\bibitem{he2016deep}
K.~He, X.~Zhang, S.~Ren, and J.~Sun, ``Deep residual learning for image
  recognition,'' in \emph{Proceedings of the IEEE conference on computer vision
  and pattern recognition}, 2016, pp. 770--778.

\bibitem{szegedy2017inception}
C.~Szegedy, S.~Ioffe, V.~Vanhoucke, and A.~A. Alemi, ``Inception-v4,
  inception-resnet and the impact of residual connections on learning.'' in
  \emph{AAAI}, vol.~4, 2017, p.~12.

\bibitem{long2015fully}
J.~Long, E.~Shelhamer, and T.~Darrell, ``Fully convolutional networks for
  semantic segmentation,'' in \emph{Proceedings of the IEEE conference on
  computer vision and pattern recognition}, 2015, pp. 3431--3440.

\bibitem{noh2015learning}
H.~Noh, S.~Hong, and B.~Han, ``Learning deconvolution network for semantic
  segmentation,'' in \emph{Proceedings of the IEEE International Conference on
  Computer Vision}, 2015, pp. 1520--1528.

\bibitem{badrinarayanan2017segnet}
V.~Badrinarayanan, A.~Kendall, and R.~Cipolla, ``Segnet: A deep convolutional
  encoder-decoder architecture for image segmentation,'' \emph{IEEE
  transactions on pattern analysis and machine intelligence}, vol.~39, no.~12,
  pp. 2481--2495, 2017.

\bibitem{fu2017stacked}
J.~Fu, J.~Liu, Y.~Wang, and H.~Lu, ``Stacked deconvolutional network for
  semantic segmentation,'' \emph{arXiv preprint arXiv:1708.04943}, 2017.

\bibitem{dong2016image}
C.~Dong, C.~C. Loy, K.~He, and X.~Tang, ``Image super-resolution using deep
  convolutional networks,'' \emph{IEEE transactions on pattern analysis and
  machine intelligence}, vol.~38, no.~2, pp. 295--307, 2016.

\bibitem{jegou2017one}
S.~J{\'e}gou, M.~Drozdzal, D.~Vazquez, A.~Romero, and Y.~Bengio, ``The one
  hundred layers tiramisu: Fully convolutional densenets for semantic
  segmentation,'' in \emph{Computer Vision and Pattern Recognition Workshops
  (CVPRW), 2017 IEEE Conference on}.\hskip 1em plus 0.5em minus 0.4em\relax
  IEEE, 2017, pp. 1175--1183.

\bibitem{burger2012image}
H.~C. Burger, C.~J. Schuler, and S.~Harmeling, ``Image denoising: Can plain
  neural networks compete with bm3d?'' in \emph{Computer Vision and Pattern
  Recognition (CVPR), 2012 IEEE Conference on}.\hskip 1em plus 0.5em minus
  0.4em\relax IEEE, 2012, pp. 2392--2399.

\bibitem{ronneberger2015u}
O.~Ronneberger, P.~Fischer, and T.~Brox, ``U-net: Convolutional networks for
  biomedical image segmentation,'' in \emph{International Conference on Medical
  image computing and computer-assisted intervention}.\hskip 1em plus 0.5em
  minus 0.4em\relax Springer, 2015, pp. 234--241.

\bibitem{cciccek20163d}
{\"O}.~{\c{C}}i{\c{c}}ek, A.~Abdulkadir, S.~S. Lienkamp, T.~Brox, and
  O.~Ronneberger, ``3d u-net: learning dense volumetric segmentation from
  sparse annotation,'' in \emph{International Conference on Medical Image
  Computing and Computer-Assisted Intervention}.\hskip 1em plus 0.5em minus
  0.4em\relax Springer, 2016, pp. 424--432.

\bibitem{chen2017voxresnet}
H.~Chen, Q.~Dou, L.~Yu, J.~Qin, and P.-A. Heng, ``Voxresnet: Deep voxelwise
  residual networks for brain segmentation from 3d mr images,''
  \emph{NeuroImage}, 2017.

\bibitem{milletari2016v}
F.~Milletari, N.~Navab, and S.-A. Ahmadi, ``V-net: Fully convolutional neural
  networks for volumetric medical image segmentation,'' in \emph{3D Vision
  (3DV), 2016 Fourth International Conference on}.\hskip 1em plus 0.5em minus
  0.4em\relax IEEE, 2016, pp. 565--571.

\bibitem{chen2017dcan}
H.~Chen, X.~Qi, L.~Yu, Q.~Dou, J.~Qin, and P.-A. Heng, ``Dcan: Deep
  contour-aware networks for object instance segmentation from histology
  images,'' \emph{Medical image analysis}, vol.~36, pp. 135--146, 2017.

\bibitem{kamnitsas2017efficient}
K.~Kamnitsas, C.~Ledig, V.~F. Newcombe, J.~P. Simpson, A.~D. Kane, D.~K. Menon,
  D.~Rueckert, and B.~Glocker, ``Efficient multi-scale 3d cnn with fully
  connected crf for accurate brain lesion segmentation,'' \emph{Medical image
  analysis}, vol.~36, pp. 61--78, 2017.

\bibitem{havaei2017brain}
M.~Havaei, A.~Davy, D.~Warde-Farley, A.~Biard, A.~Courville, Y.~Bengio, C.~Pal,
  P.-M. Jodoin, and H.~Larochelle, ``Brain tumor segmentation with deep neural
  networks,'' \emph{Medical image analysis}, vol.~35, pp. 18--31, 2017.

\bibitem{dou20173d}
Q.~Dou, L.~Yu, H.~Chen, Y.~Jin, X.~Yang, J.~Qin, and P.-A. Heng, ``3d deeply
  supervised network for automated segmentation of volumetric medical images,''
  \emph{Medical image analysis}, vol.~41, pp. 40--54, 2017.

\bibitem{chen2018deeplab}
L.-C. Chen, G.~Papandreou, I.~Kokkinos, K.~Murphy, and A.~L. Yuille, ``Deeplab:
  Semantic image segmentation with deep convolutional nets, atrous convolution,
  and fully connected crfs,'' \emph{IEEE transactions on pattern analysis and
  machine intelligence}, vol.~40, no.~4, pp. 834--848, 2018.

\bibitem{oktay2018anatomically}
O.~Oktay, E.~Ferrante, K.~Kamnitsas, M.~Heinrich, W.~Bai, J.~Caballero, S.~A.
  Cook, A.~de~Marvao, T.~Dawes, D.~P. O‘Regan \emph{et~al.}, ``Anatomically
  constrained neural networks (acnns): application to cardiac image enhancement
  and segmentation,'' \emph{IEEE transactions on medical imaging}, vol.~37,
  no.~2, pp. 384--395, 2018.

\bibitem{gibson2018automatic}
E.~Gibson, F.~Giganti, Y.~Hu, E.~Bonmati, S.~Bandula, K.~Gurusamy, B.~Davidson,
  S.~P. Pereira, M.~J. Clarkson, and D.~C. Barratt, ``Automatic multi-organ
  segmentation on abdominal ct with dense v-networks,'' \emph{IEEE Transactions
  on Medical Imaging}, 2018.

\bibitem{tu2010auto}
Z.~Tu and X.~Bai, ``Auto-context and its application to high-level vision tasks
  and 3d brain image segmentation,'' \emph{IEEE Transactions on Pattern
  Analysis and Machine Intelligence}, vol.~32, no.~10, pp. 1744--1757, 2010.

\bibitem{lee2015deeply}
C.-Y. Lee, S.~Xie, P.~Gallagher, Z.~Zhang, and Z.~Tu, ``Deeply-supervised
  nets,'' in \emph{Artificial Intelligence and Statistics}, 2015, pp. 562--570.

\bibitem{huang2017densely}
G.~Huang, Z.~Liu, K.~Q. Weinberger, and L.~van~der Maaten, ``Densely connected
  convolutional networks,'' in \emph{Proceedings of the IEEE conference on
  computer vision and pattern recognition}, vol.~1, no.~2, 2017, p.~3.

\bibitem{ioffe2015batch}
S.~Ioffe and C.~Szegedy, ``Batch normalization: Accelerating deep network
  training by reducing internal covariate shift,'' \emph{arXiv preprint
  arXiv:1502.03167}, 2015.

\bibitem{nair2010rectified}
V.~Nair and G.~E. Hinton, ``Rectified linear units improve restricted boltzmann
  machines,'' in \emph{Proceedings of the 27th international conference on
  machine learning (ICML-10)}, 2010, pp. 807--814.

\bibitem{chollet2017xception}
F.~Chollet, ``Xception: Deep learning with depthwise separable convolutions,''
  \emph{arXiv preprint}, pp. 1610--02\,357, 2017.

\bibitem{szegedy2016rethinking}
C.~Szegedy, V.~Vanhoucke, S.~Ioffe, J.~Shlens, and Z.~Wojna, ``Rethinking the
  inception architecture for computer vision,'' in \emph{Proceedings of the
  IEEE conference on computer vision and pattern recognition}, 2016, pp.
  2818--2826.

\bibitem{devries2017improved}
T.~DeVries and G.~W. Taylor, ``Improved regularization of convolutional neural
  networks with cutout,'' \emph{arXiv preprint arXiv:1708.04552}, 2017.

\bibitem{glorot2010understanding}
X.~Glorot and Y.~Bengio, ``Understanding the difficulty of training deep
  feedforward neural networks,'' in \emph{Proceedings of the thirteenth
  international conference on artificial intelligence and statistics}, 2010,
  pp. 249--256.

\bibitem{smoothTriangMesh}
\BIBentryALTinterwordspacing
D.-J. Kroon, ``Smooth triangulated mesh,'' File Exchange - MATLAB Central, 2010
  (accessed: Jun. 8, 2018). [Online]. Available:
  \url{http://mathworks.com/matlabcentral/fileexchange/26710}
\BIBentrySTDinterwordspacing

\end{thebibliography}

\end{document}